# Formalization and Parsing of Typed Unification–Based ID/LP Grammars

Frank Morawietz*

Sonderforschungsbereich 340
Universität Tübingen
Kleine Wilhelmstr. 113
72074 Tübingen

E-mail: `frank@sfs.nphil.uni-tuebingen.de`

February 13, 1995

*The research presented in this paper is a modified version of my Master's thesis and was partly sponsored by Teilprojekt B4 "Constraints on Grammar for Efficient Generation" of the Sonderforschungsbereich 340 "Sprachtheoretische Grundlagen für die Computerlinguistik" of the Deutsche Forschungsgemeinschaft. I want to thank Thilo Götz, John Griffith, Tilmann Höhle, Stephan Kepser, Paul King, Dieter Martini, Detmar Meurers, Guido Minnen, Frank Richter, Manfred Sailer, and last but not least my supervisors Prof. Dr. E. Hinrichs and Dr. D. Gerdemann for helpful discussion and advice. Needless to say, all errors and infelicities that remain are my own.

**Abstract**

This paper defines unification based ID/LP grammars based on typed feature structures as nonterminals and proposes a variant of Earley's algorithm to decide whether a given input sentence is a member of the language generated by a particular typed unification ID/LP grammar. A solution to the problem of the nonlocal flow of information in unification ID/LP grammars as mentioned in Seiffert (1991) is incorporated into the algorithm. At the same time, it tries to connect this technical work with linguistics by giving a motivation drawn from natural language and by drawing connections from this approach to HPSG (Pollard and Sag 1994) and to systems implementing it, especially the TROLL system (Gerdemann, Götz and Morawietz forthcoming).

# Contents

# 1 Introduction

In this paper we are going to define unification based ID/LP grammars based on typed feature structures as nonterminals and to propose a variant of Earley's algorithm to decide whether a given input sentence is a member of a language defined by a particular typed unification ID/LP grammar. At the same time, we try to connect this technical work with computational linguistics by drawing connections from our approach to a contemporary linguistic theory, namely Head Driven Phrase Structure Grammar (HPSG) (Pollard and Sag 1994) and to systems implementing it.

The immediate dominance – linear precedence distinction, henceforth ID/LP, was introduced into linguistic formalisms to easily encode word order generalizations by Gazdar, Klein, Pullum and Sag (1985) for Generalized Phrase Structure Grammar (GPSG). Ever since, it has been prominent with linguists and contemporary formalisms like HPSG want to express linearization facts with linear precedence (LP) rules. But HPSG does not provide succinct definitions how to incorporate the ID/LP format into the formalism. This thesis tries to show a way to handle the ID/LP distinction on another level. Instead of incorporating it into the theory, the information will be used during processing to determine validity of structures. We will use the formalization of the typed unification based ID/LP grammars as a basis for the presentation of the solution.

We do not try to give an extensive motivation and definition of the ID/LP formalism here, but rather try to provide some intuition about the distinctions involved. A complete introduction to the handling of ID/LP in GPSG can be found in Gazdar and Pullum (1982) and formal definitions for the purposes at hand will be given in section 5.

$$
\begin{array}{lcl}
S & \rightarrow & A\ B\ C \\
S & \rightarrow & A\ C\ B \\
S & \rightarrow & B\ A\ C \\
A & \rightarrow & a \\
B & \rightarrow & b \\
C & \rightarrow & c
\end{array}
$$

Figure 1: Some context free rules

In languages with relatively fixed word order, such as for example English, a context free phrase structure component seems to suffice to express the relation between dominance relations, as for example constituency, and the correlated precedence phenomena, i.e syntactic properties of the categories. Context free phrase structure rules express both kinds of information at the same time by specifying which category can be rewritten to which categories in which order. But research of free word order languages, as for example Makua (Stucky 1981), seems to indicate that this misses some important generalizations and leads to a somewhat unintuitive representation. Immediate dominance (ID) rules do convey only the information which category can be rewritten by what other categories, but the information in which order these categories have to be realized has to be specified additionally with LP rules. Although this may look more complicated at first, it allows for a more elegant and more efficient representation of data. Consider the very simple set of context free rules in figure 1.

They allow all the permutations for the terminals *a,b,c* apart from those where the *c* precedes the *a*. The ID/LP version of this grammar expresses the same language with fewer rules and the explicitly stated generalization that *C*s must not precede *A*s (see figure 2).

$$\begin{array}{lcl} S & \rightarrow & A,\ B,\ C \\ A & \rightarrow & a \\ B & \rightarrow & b \\ C & \rightarrow & c \end{array} \qquad\qquad A \prec C$$

Figure 2: Some ID/LP rules

These rules express the same language, but in a more concise and elegant way. Since linear precedence rules apply to all the siblings in all rules, it seems indeed worthwhile to separate the information on dominance and precedence contained in context free rules. Another intuition behind this is that free word order should be easier to express than fixed word order. This is reflected in the ID/LP formalism by the need to have more LP rules to forbid permutations compared to simply allowing all permutations. Naturally this assumption may be doubted. Although the ID/LP distinction may be doubted as well for reasons connected with learnability of negative information (Grimshaw and Pinker 1989) and computational and semantic reasons (Meurers and Morawietz ms), we nevertheless follow this paradigm since it provides a well established representation for word order variations and so far no other approach has been formalized properly.

As for the ID/LP format, we do not attempt to give a full motivation for the adequacy of HPSG for describing natural language, but rather try to provide the reader with a rough account of the background behind the approach. Therefore, the thesis assumes some familiarity with HPSG insofar as some concepts and specific properties of HPSG are relevant without being explained.

HPSG is well suited for computational linguistics for two reasons. On the one hand, HPSG provides a uniform representation for different levels of information such as syntax, semantics and phonology. On the other hand, HPSG is mathematically well founded by some logics, for example Kasper and Rounds (1986), King (1989) or Carpenter (1992)[1]. Clearly this qualifies HPSG as a framework for the formulation and implementation of linguistic theories. The formal development between the first formal definition of HPSG (Pollard and Sag 1987) and the recent version (Pollard and Sag 1994) is considerable, nevertheless this thesis does not refer to a particular version of the theory, but rather tries to define a new class of grammars which can be handled efficiently from a computational point of view in a way that HPSG grammars may be viewed as such by neglecting and simplifying some of HPSG's properties.

A concept of HPSG which is important for the thesis is the use of sorted and type resolved feature structures to model linguistic objects. HPSG uses a description language to express a theory which interprets the linguistic objects via those sorted and type resolved feature structures. As mentioned before, the formalization of the theory behind HPSG has been done by various kinds of feature logics (among others Kasper and Rounds (1986), King (1989) or Carpenter (1992)). The thesis is not concerned with this kind of approach toward HPSG, but nevertheless feature structures play a role as a means for conveying information. The partial feature structures defined later on in the thesis are different from those sorted, type resolved feature structures used by HPSG since they are used during processing of HPSG like grammars without a definition of their semantics and therefore without any claim towards modeling of linguistic objects.

An HPSG theory consists of two parts, namely the declaration of a type hierarchy with appropriateness restrictions which defines the domain of the grammar and the formulation of constraints on this domain. The resulting language is the collection of the licensed objects in this domain. It follows from that that HPSG is in essence a constraint based system, i.e. it does not employ any processing specific constructs as for example overt phrase structure rules. Linguistic structures are not generated or processed in a procedural way, but are admissible, i.e. satisfy all the constraints. The constraints are order independent and can therefore be processed in a monotonic fashion. And constraints can take the form of principles, for example the dominance principle. If a structure confirms to all constraints, it is a valid structure of the language. So HPSG allows for declarative specification of grammars, i.e. it is processing neutral towards generation/parsing.

Since HPSG does not employ overt phrase structure rules, the application of the ID/LP paradigm as known from GPSG is not immediately obvious. Perhaps the most natural way one may view the problem is to treat the phonology as representing the chronological order of the string, whereas the information contained in the daughters attribute is supposed to model the information on dominance. Following from that, it seems straightforward that any approach which is to be formulated in HPSG itself necessarily has to be very complex, since it has to relate those two concepts without being able to have a clear distinction of those constraints on phrase structure and word order from other

[1]Although it is an open issue whether any of the formalizations captures exactly the intentions behind HPSG.

constraints, but this is not the way taken in this thesis.

If one considers a natural language system which is supposed to implement HPSG or other unification based formalisms, two basic approaches are available. A rule based approach (cf. Shieber (1986), Gerdemann and Hinrichs (1988), Carpenter (1993) and Gerdemann et al. (forthcoming)) and a constraint based approach (cf. Aït-Kaci (1984), Franz (1990) and Zajac (1992)). The constraint based approach is more faithful to the theory, but so far implementations in this paradigm have been inefficient. If one chooses a rule based approach, which seems preferable considering efficiency, standard processing algorithms become applicable. This rule based approach allows one to view HPSG as a typed unification ID/LP grammar, though this does not meet HPSG's original intentions and does not reflect the more sophisticated properties, as for example the semantic ones of at least some of the systems.

This thesis presents an approach to a specialized algorithm to handle this rule based version of HPSG in one processing mode, namely parsing.

The thesis proceeds as follows. In section 2 some literature on approaches to ID/LP parsing is reviewed, a particular approach is chosen and the remaining problems are discussed. Section 3 gives some linguistic motivation behind the major problem. The next two sections define typed unification based ID/LP grammars where section 4 defines the nonterminals as feature graphs and section 5 contains the presentation of the actual definitions concerned with the language generated by typed unification ID/LP grammars. Following on that, section 6 contains the algorithm for parsing the previously defined class of grammars which solves the main remaining problem of the chosen approach. An implementation of the algorithm is the topic of section 7 to show the applicability of the algorithm to real natural language systems. The last section 8 sums up the results, discusses some open questions and presents some outlook on transfer possibilities of the approach.

# 2 ID/LP Parsing

This section presents some of the approaches toward ID/LP parsing found in the literature. At the same time, it tries to reason why the approach taken later on in the thesis seems desirable. Although the problem of ID/LP parsing in general has been shown to be $\mathcal{NP}$–complete by Barton, Berwick and Ristad (1987), this fact does not prevent having efficient algorithms for average applications.

Since ID/LP format is a development which gained importance for linguistics by the presentation of GPSG (Gazdar et al. 1985) most of the literature on ID/LP parsing is concerned with the parsing of GPSG grammars. Nevertheless (most of) the techniques can be extended to deal with more sophisticated grammars, as for example HPSG type grammars.

The simplest approach to parsing of ID/LP grammars is to fully expand the grammar into the underlying phrase structure grammar. In the case of GPSG, all the other principles were claimed to be similarly only of context free power so that a compilation to a context free grammar was an option. Thus parsing with this context free grammar with standard algorithms becomes possible. But the expansion creates a huge number of grammar rules[2] which dominates the parsing complexity and can therefore not be a basis for a reasonable implementation. This parsing with the object grammar is called *indirect parsing*. Since it is computationally not a suitable treatment of the ID/LP format, this approach will – in its pure form – not be pursued further.

On the other hand, there are approaches, called *direct parsing* which try to use ID and LP rules directly. The approach taken in this thesis is a result of augmenting this paradigm.

But before we present the concept of direct parsing, a couple of other approaches will be discussed briefly. Most of those are HPSG specific solutions to the problem.

## 2.1 Handling of LP Rules

Oliva (1992) is not directly concerned with parsing of ID/LP grammars, but rather tries to provide a possibility to express ordering constraints, i.e. linear precedence constraints, within the HPSG framework. This approach is supposed to take the place of the LP formalism as assumed for GPSG. He extends the type hierarchy for lists and collects the appropriate information in different types of lists, so that the domain for LP rules can become larger than local trees. By this non–sister constituents can be ordered relative to each other. So, the word order regularities have to be encoded in the type hierarchy on lists, rather than being used as a separate device (LP rules) to limit parsing. Since the author does not provide any indication that there is a way to create this type hierarchy from standard LP rules (at best automatically), the linguist would have to familiarize himself with this kind of formalism to be able to express word order limitations. This approach is to be taken as exemplary that there are ways to circumvent the need to encode the ID/LP formalism as such, i.e. to have some equivalent to ID and LP rules. But since HPSG is to be in ID/LP format, although the authors of the HPSG II book (Pollard and Sag 1994) do not provide sufficient information on this, it seems preferable to find a way to let the linguist express the grammar in ID/LP format. There are other approaches as well which encode word order generalizations in other ways than by LP rules, for example those by Reape (1990), Richter and Sailer (1995) or Meurers and Morawietz (ms). Naturally, it remains an empirical question what is needed to express the linguistic phenomena correctly and which formal apparatus is needed to achieve this. We do not try to resolve the issue here, but rather propose a way to deal with the ID/LP format in another way. Nevertheless we concede that those approaches are closer to the intentions of HPSG.

[2] For unification grammars the number of rules may even be infinite.

In the logical paradigm, the approach taken by Blache (1992) tries to incorporate the LP rules as active constraints. Since parsing can be seen as deduction (Pereira and Warren 1983), the author follows the analogy to treat the rules as clauses (or implications). The actual parsing algorithm is a variant of bottom–up filtering (Blache and Morin 1990). LP rules are then used to determine possible boundaries for phrases, possible immediate precedence relations and the notion of an initial element. LP acceptability is checked on each phrase. So the purpose of LP rules is twofold. They help to determine phrases and they check the validity of the phrases themselves. Though this is formulated only for basic ID/LP grammars, i.e. atomic nonterminals, Blache claims that it is no problem to extend this to feature based systems. Some problems remain open nevertheless. The rules are limited to contain each category at most once which is clearly inadequate if one considers linguistic needs, for example the case of prepositional phrases where undoubtedly more than one prepositional phrase can appear in a rule. It is not discussed when an LP rule actually applies to a structure. This is clearly important for feature based systems since identity does not suffice any more. No formal definition of an ID/LP grammar is given, so the properties of the LP rules which might lead to problems, as for example transitivity, are not discussed. The domain of the application of the LP rules is limited to phrases and therefore it is not clear whether it is possible to extend this to deal with phenomena which seem to require a larger domain (Uszkoreit 1986). And the problem of the nonlocal flow of information which will be explained below is not addressed at all.

The remaining two approaches try to follow HPSG's idea of representing information in principles and in the lexicon. Engelkamp, Erbach and Uszkoreit (1992) try to exploit HPSG's properties to cope with LP rules in a way that violation of an LP rule leads to unification failure. They limit themselves to binary branching ID rules, but this limitation can be avoided at the cost of more principles. Following Uszkoreit (1986) the domain for the application of LP rules is a so called head domain consisting of the head and its adjuncts and complements. LP rules are precompiled into the lexical entries in such a way that every entry contains information as to what must not appear to its left and right respectively. During parsing, this information is accumulated in an LP store so that the violation can be discovered locally. This accumulation is defined via argument – head relationships, so that each projection contains the information which categories appear in its head domain. Two new principles and some features are introduced to ensure this. The principles parallel in some way the ID schemata in the sense that they ensure the appropriate connections between the different LP informations and failure if the information is not consistent. Additionally, they choose to have the value for the SUBCAT feature to be a set rather than a list following (cf. Pollard (in press)) since they want to determine linear precedence entirely by LP statements. The linear order of the constituents is reflected in the phonology. Parameterized types (Dörre 1991) are used in the lexicon to ensure the right description of the LP relevant features.

The precompilation of the LP information is not demonstrated with an example and the theoretical explanations remain somewhat unclear. Since the lexical entries have to be initialized concerning the LP relevant information, the application of the LP rules to a particular entry is claimed to be done by subsumption. This may lead to problems if the lexical entry is further instantiated during processing because some LP rules may become applicable only in connection with other information. Following on that, they suffer from the problem of nonlocal feature passing as will be explained in later sections. Furthermore, they have to determine which entries can serve as a head because those have to have an LP store. If those entries contain LP relevant information as well, they have to contain information on what must not appear to its left and right and how the LP store is changed in case it is combined with some adjunct or complement. It is not clear how this might be determined since being a head is not a piece of information which is (standardly) encoded in the entry itself, but becomes only possible in relation to ID schemata.

Although the approach is very elegant since it does not necessitate any additional mechanisms to deal with word order and is neutral concerning processing mode (parsing/generation) and direction (top–down/bottom–up), the drawback is obvious. An enormous amount of information on each entry and all the intermediate stages of the processing has to be duplicated and carried around. Since feature structures tend to be large to begin with, this may be a problem for implementations in terms of the used storage space and the efficiency of the processing. The restriction of having binary branching structures only can be avoided easily by having additional principles to cope with the resulting variations on percolating the information relevant to linear precedence. But in case the grammar writer is really using ID schemata as opposed to ID rules, i.e. the number of the daughters is not fixed, it is not so clear anymore how to formulate the corresponding LP principles. Furthermore, the principles rely crucially on the order of the constituents, i.e. the position of the head in re-lation to the complement(s) or adjunct(s). This order does not exist in the formalism of HPSG. It is an artificial construct of the processing. The distinction between the word order which is supposed to be done in the formalism with the phonology and processing order which is reflected in the daughters feature gets blurred. With respect to this, the authors do not entirely support their claim to provide a mechanism for the ID/LP format in HPSG itself. Nevertheless it is an attractive approach which might be extended to conform to the claim if one clarifies the points discussed above.

Two concluding remarks seem necessary. The decision of having a subcat set would lead to problems with HPSG's binding theory. But this is not an issue here. Though the problem is mentioned, no treatment of conflicting LP statements is proposed. LP constraints are viewed as absolute constraints that lead to failure.

Morawietz (1993) takes the lexicalist approach even further. Starting from the simple fact that the subcat list pretty much reflects the word order for English, the LP rules are precompiled into the lexicon in such a way that the subcat list of the lexical entries reflects the word order. If several possibilities exist, this is represented by disjunctions, i.e. either several entries or a single entry with a disjunction of subcat lists if the formalism allows such a complex disjunction. This approach presupposes *signs* on the subcat list instead of the *synsem* objects assumed in HPSG since LP rules order elements on the right hand side of rules which are standardly assumed to be *signs*. In effect, each subcat list is taken and all LP acceptable permutations are generated. This is done using subsumption to determine if an LP rule applies. In cases where an LP rule might apply at some later stage, but the feature structure is not yet specific enough, the information is unified into the structure. Clearly this introduces a great deal of nondeterminism for the processing, but no further LP relevant calculations have to be done at runtime. The most efficient treatment of disjunctions possible as for example explained in Griffith (forthcoming) is essential. This approach towards the handling of LP rules is particularly efficient if the word order is fairly strict, since in this case less permutations are generated and less nondeterminism is introduced. Naturally the ID schemata have to be changed so that they order the complements according to the order of the elements on the subcat list. Since the binding theory relies on the obliqueness hierarchy reflected in the order of the elements on the subcat list, it is an issue of further research if those disjunctions exhibit the correct behavior concerning binding phenomena or whether one has to construct a special list to do word order. It would be particularly attractive and would provide support for this approach if one could show that binding and different versions of the subcat list correlate.

Open problems are the order of the head in relation to the complements and the adjuncts. But this seems to reflect the nature of the LP rules. Some LP rules are very strict and easily expressed in phrase structure rules, as for example the head filler schema. Some others are relative concerning specific instances of categories, as for example the fact that pronominal NPs (usually) precede non pronominal NPs. Whether this is indeed an empirical difference, is still open to research. Naturally the major drawback is the possible explosion of the number of lexical entries. If the language allows free word order, a lot of permutations have to be considered. In this sense, the approach is close to indirect parsing. Here, it is not the number of grammar rules that is too big, but rather the lexicon. If there is no mechanism for dealing with complex disjunctions, the lexicon may blow up. If an efficient approach to the handling of disjunctions is available the problem of too many entries reduces somewhat because the handling of the nondeterminacy is done by the mechanism that handles the disjunctions.

## 2.2 Direct ID/LP Parsing

Since HPSG does not provide clear definitions how to handle the ID/LP format and it seems somewhat unclear how to incorporate it in the formalism, the most promising approaches so far try to provide some extra mechanism outside the formalism to provide efficient handling of LP rules. Direct parsing is one of those possibilities to handle word order by directly applying the LP rules during processing.

The concept of direct parsing was first developed by Shieber (1984). He modifies Earley's algorithm (Earley 1970) to cope with ID/LP grammars.[3] Firstly (context–free) ID/LP grammars are defined by treating the formerly context–free rules as ID rules and by adding LP rules. The modification to the algorithm consists of two parts. The right hand side of a rule has no longer a fixed order, but is rather treated as a multiset.[4] Taking a multiset instead of simply a set becomes necessary to allow two or more occurrences of the same nonterminal on the right hand side of a rule. The predictor and the completer are limited to the LP acceptable structures by testing LP acceptability on the considered local trees, i.e. a nonterminal is extracted from the multiset and tested whether it may precede the remaining categories. In Earley's algorithm the prediction and completion were done for the leftmost nonterminal only. Since this approach deals only with atomic nonterminals, the question when an LP rule applies does not arise. This makes it possible to take the LP rules and to match them directly against proposed structures. Since this proposal obviously does not suffice to handle HPSG style grammars, it has to be augmented to feature based grammars. But firstly some optimizations to Shieber's algorithm are presented.

The approach proposed in Kilbury (1984b) and Kilbury (1984a) is essentially the same as Shieber's. But Kilbury notes that Shieber's predictor is fairly inefficient since the predictions are made without regard for the resulting structure. To achieve better performance the parser employs techniques commonly used with left corner parsers. Prediction is only done for the first legal daughters of a rule. Those legal daughters are determined in a precompilation step using the LP rules. To relate the rules and the first legal daughters to each other, there has to be a unique identifier for each rule so that an efficient representation of the relation becomes possible. This precompiled relation is called *First*. Dörre and Momma (1985) extend this even further by calculating the transitive closure of First, namely *First*$^{+}$. This allows the predictor in some sense to look even deeper into the structure to determine in which cases a category may indeed be a first legal daughter, or better may result in an LP acceptable first legal daughter. In some sense, the predictor is waiting until the structure it wants to propose has been completely analyzed. Clearly this strengthens the bottom up component of the algorithm. Again this is done for grammars with atomic nonterminals and it is not immediately clear that the First relation can indeed be precompiled for feature based grammars since the necessary information may not yet be present. If we are dealing with largely underspecified categories, structures may be deemed suitable as first legal daughters which are not allowed in any of their instantiations later on. This would reintroduce the problem of inefficiency of the predictor Kilbury wanted to avoid. The problem of underspecification of the categories is noted in another context by Weisweber (1987). He discusses the problem that information may be added by feature instantiation principles of GPSG and that this information may be necessary to determine the LP acceptability of a structure. His solution consists simply in the postponing of the test for LP acceptability until the whole local tree has been recognized and all the features have been instantiated. This excludes LP unacceptable structures until fairly late during processing and a lot of unacceptable partial structures are constructed and stored. Clearly this is not desirable.

[3] This section assumes some familiarity with Earley's algorithm. A more detailed introduction to Earley's algorithm can be found in section 6.3.

[4] A multiset (also called a *bag*) is just like a set, except that it may contain each element a finite number of times.

Another modification of the Earley/Shieber algorithm is proposed in Meknavin, Okumura and Tanaka (1992). Nothing changes concerning the actual principle of the algorithm or the definition of ID/LP grammars, but the internal representation is optimized to ensure the greatest possible efficiency. ID rules are compiled into generalized discrimination networks and LP rules into Hasse diagrams (cf. Meknavin et al. (1992) for details on these terms) using bit vectors. Though the implementation of this idea outperforms by far Shieber's parser, the drawbacks are the same as noted for Shieber. It does not deal with unification and feature structure based grammars, but just with atomic terminals and nonterminals. And therefore nothing deals with the nonlocal flow of information.

A slightly different view of LP rules is proposed by Evans (1987). For him, LP rules are a global property of a grammar, i.e. they are not directly connected with the ID rules. Otherwise LP rules would always be local to a rule, i.e. the ECPO property for the underlying context free grammar would not be necessary. He claims that LP acceptability has to be maintained on all partial parses. His parser works like a bottom up chart parser with an agenda. To reflect the globality of the LP rules, his parser works in some sense like a shift reduce parser on the edges. This means that as long as the structure is LP acceptable, categories can be added on some sort of stack and if they match the right hand side of some rule, a new edge is created which contains the left hand side of this particular rule. But nothing prevents the parser from further adding constituents to the old edges as long as the sequence is still LP acceptable. Since this is inefficient, he needs a new relation to determine the boundary of phrases. He proposes to precompile from the ID rules a *sisterhood* relation, i.e. which categories may appear together as sisters. Apart from complicating the grammar, nothing is achieved by this, since boundaries of phrases are naturally determined by the ID rules. But since he does not want to use this information, he has to introduce the special sisterhood relation. Unless one does think that globality of LP rules is an important linguistic feature, this approach does not seem very attractive.

To be able to handle HPSG style grammars, one major augmentation has to be done. Unification based ID/LP grammars have to be defined and the algorithm has to be changed accordingly. A step towards the needed formalism is taken in Seiffert (1987) and Seiffert (1991).

Unification based ID/LP grammars are defined straightforwardly by augmenting the domain of the nonterminals to feature structures and by defining when an LP rules applies. This application of LP rules is crucial. An LP rule applies to a local tree in case the LP elements subsume two categories contained in this local tree. A violation occurs if the category which is subsumed by the first LP element follows the category which is subsumed by the second LP element. The Earley/Shieber parser is then used almost unchanged. The only change is that nonterminals are not longer identical, but rather have to unify in case a rule might apply. But a problem comes up that makes a second step for the algorithm necessary. Before presenting this, two further, but minor, modifications and a further problem to the algorithm are presented briefly.

The test for LP acceptability has changed. Shieber tests whether a category may precede the categories still to be found. This may not be determined for unification grammars, since the information may not have been instantiated. So, LP acceptability has to be checked on the already recognized categories. This necessitates the construction of a parse. So far, the presentation was somewhat sloppy in the sense that the distinction between a parser and a recognizer was neglected. In the more formal section, this will be more exact. For now, the distinction is not relevant. But to be correct, only now the term parser is justified. Secondly, to ensure termination, a restrictor has to be incorporated in the predictor. For motivation of this and details of the processes involved, see Shieber (1985).

But the usage of the restrictor induces a problem. If one picks a restrictor, it may well be that exactly that part of the information which is necessary to determine LP acceptability is discarded in the predictor, although the information naturally is not lost, but will be added in the completer step. This can be easily overcome by changing the restrictor to a more appropriate one or checking LP acceptability on the whole feature structure and not on the restricted version of it. Furthermore, the restrictor has to be appropriate to the grammar in some sense, to provide the maximal guidance possible.

The other problem is far more serious and leads to the mentioned complication of the algorithm. Consider the grammar in figure 3 (taken from Seiffert (1987)).

$$
\begin{array}{lcl}
\text{Lexicon} & = & \left\{ \begin{array}{lcl}
[\textsc{cat}\ d] & \rightarrow & h \\
[\textsc{cat}\ e] & \rightarrow & i \\
\begin{bmatrix} \textsc{cat}\ f \\ \textsc{f1}\ \ one \end{bmatrix} & \rightarrow & j \\
\begin{bmatrix} \textsc{cat}\ g \\ \textsc{f2}\ \ two \end{bmatrix} & \rightarrow & k
\end{array} \right\} \\
\text{ID–Rules} & = & \left\{ \begin{array}{lcl}
[\textsc{cat}\ a] & \rightarrow & \begin{bmatrix} \textsc{cat}\ b \\ \textsc{f}\ \ \boxed{1} \end{bmatrix}, \begin{bmatrix} \textsc{cat}\ c \\ \textsc{f}\ \ \boxed{1} \end{bmatrix} \\
\begin{bmatrix} \textsc{cat}\ b \\ \textsc{f}\ \begin{bmatrix} \textsc{f1}\ \boxed{2} \\ \textsc{f2}\ \boxed{3} \end{bmatrix} \end{bmatrix} & \rightarrow & \begin{bmatrix} \textsc{cat}\ d \\ \textsc{f1}\ \ \boxed{2} \end{bmatrix}, \begin{bmatrix} \textsc{cat}\ e \\ \textsc{f2}\ \ \boxed{3} \end{bmatrix} \\
\begin{bmatrix} \textsc{cat}\ c \\ \textsc{f}\ \begin{bmatrix} \textsc{f1}\ \boxed{4} \\ \textsc{f2}\ \boxed{5} \end{bmatrix} \end{bmatrix} & \rightarrow & \begin{bmatrix} \textsc{cat}\ f \\ \textsc{f1}\ \ \boxed{4} \end{bmatrix}, \begin{bmatrix} \textsc{cat}\ g \\ \textsc{f2}\ \ \boxed{5} \end{bmatrix}
\end{array} \right\} \\
\text{LP–Rules} & = & \left\{ \begin{array}{lcl}
[\textsc{f1}\ one] & \prec & [\textsc{f2}\ two] \\
[\textsc{cat}\ b] & \prec & [\textsc{cat}\ c]
\end{array} \right\} \\
\text{Start Symbol} & = & [\textsc{cat}\ a] \\
\text{L(G)} & = & \{hijk\}
\end{array}
$$

Figure 3: Seiffert's example grammar

On input *ihjk* – which is not well formed – Seiffert's parser can not determine on the first pass whether the local tree $\begin{array}{c}[\textsc{cat}\ b]\\ \diagup\ \diagdown \\ [\textsc{cat}\ e]\ \ [\textsc{cat}\ d]\end{array}$ is well formed since the information that is relevant for the LP rule, namely that *one* has to precede *two*, is not available yet, since parsing is done left to right in this case. Only after the local trees $\begin{array}{c}[\textsc{cat}\ c]\\ \diagup\ \diagdown \\ [\textsc{cat}\ f]\ \ [\textsc{cat}\ g]\end{array}$ and $\begin{array}{c}[\textsc{cat}\ a]\\ \diagup\ \diagdown \\ [\textsc{cat}\ b]\ \ [\textsc{cat}\ c]\end{array}$ have been constructed, the information becomes accessible through structure sharing. The nonlocal flow can be seen in the parse tree given in figure 4.[5] The information *one* and *two* comes from the lexical entries for *f* and *g*, is passed to their mother *c* via the structure sharing indicated by the tags $\boxed{4}$ and $\boxed{5}$. The information on *c* is (partly) structure shared with the information on *b*, indicated by the tag $\boxed{1}$. This forces equality of the tags $\boxed{2}$ and $\boxed{4}$, and $\boxed{3}$ and $\boxed{5}$ respectively. Then the information is passed to the daughters of *b*, namely *e* and *d*, via the tags $\boxed{2}$ and $\boxed{3}$. Now the information that the value for F2

[5] For readability reasons the arrows are drawn just for one value, namely *two*.

at *e* is *two* and that the value for F1 at *d* is *one* is available and the local tree 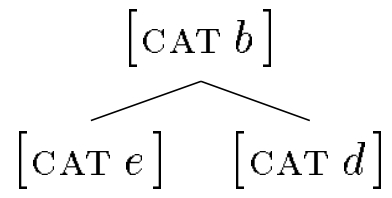

has to be ruled out since the LP rule *one* $\prec$ *two* is violated.

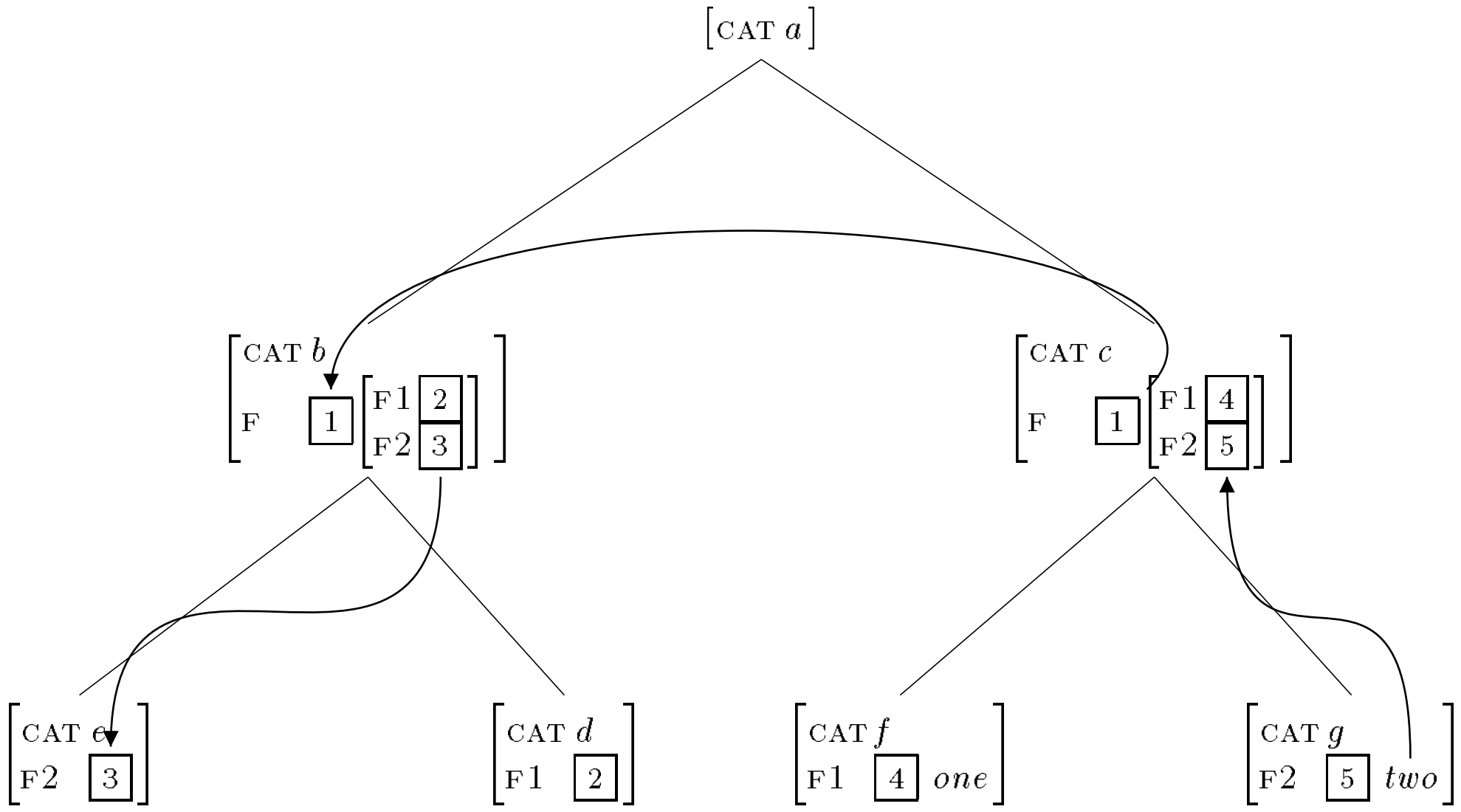

Figure 4: The parse tree for input **ihjk* and Seiffert's grammar (see figure 3)

This makes it necessary for Seiffert to take the completed parse tree and recheck it for any violations of LP rules after it has been completed in a second step. His entire algorithm consists of the modification of Shieber's algorithm and the second step of ruling out all the parse trees which still contain a local tree with a violation of LP rules. So he is checking LP acceptability twice, namely while constructing each local tree and after the parse has been completed. Clearly this is inefficient and not desired.

The other approaches discussed before are faced with the same problem, with the exception of the approaches by Oliva (1992) and Richter and Sailer (1995) which try to incorporate ID/LP into the HPSG formalism, since HPSG only demands a licensed structure, which can be achieved without regard to local processing problems. Implementations of those ideas may nevertheless be faced with the problem if they are in any way procedural. Naturally all the approaches which deal only with atomic nonterminals have to find a way to cope with the phenomenon if they are to be extended. The approach by Blache (1992) has the problem that the implications reflect only one rule, or better one category which might be extended in several ways, but there is no way that nonlocal information can be incorporated directly. Since the implementation of his approach is procedural, it may be possible to find a way to circumvent the problem along the lines proposed in this thesis. Even the approach by Engelkamp et al. (1992) suffers to some extend. Since they have to decide what must not appear to the left and to the right of lexical entries beforehand in a precompilation step, the relevant information might not be present and so the proper values for the features which contain what must not appear to the left and right of the feature structure in question may not be instantiated properly due to the determination of the application of the LP rules by subsumption. Maybe the problem is less likely to occur since the domain for the application of LP rules has been augmented to head domains instead of local trees. A possible solution to this problem in their approach might be similar to the way taken in this thesis. The approach by Morawietz (1993) avoids the problem by adding the necessary information by unification and thereby constructing all possible legal alternatives.

In the following, the thesis presents a slightly augmented class of grammars with a further modification of the Earley/Shieber algorithm which solves the presented problem of the nonlocal flow of information. But before that, we proceed by arguing to some extend whether the problem may indeed occur in natural language. It seems desireable to continue in the paradigm of direct parsing from a computational point of view since it is well understood and shows the advantage of clearly defining the problem. But nevertheless the solution to the problem may be adapted for other approaches as well.

# 3 Illustrating Nonlocal Feature Passing: A Linguistic Example

Although this thesis is mostly not concerned with linguistics, we nevertheless try to reason to some extent that the presented problem is indeed not solely a technical one, but rather turns up in natural language analyses as well. Nonlocality is a well known phenomenon in natural language, but so far without any influence on word order. This section tries to sketch the fact that in a reasonably large fragment for German such nonlocal word order phenomena do exist. Since this is not the main goal of the thesis, the presentation needs to be brief and can not be in any detail. The section can be understood without an exact knowledge of the linguistic theories involved if one accepts the claims made during the discussion, but for a complete understanding of the details involved, we assume some knowledge of the analysis of partial verb phrases in German by Hinrichs and Nakazawa (1993), on word order in German by Lenerz (1977) and the empirical analyses by Richter and Sailer (1995).[6]

Some data on word order in German sentences seem to suggest that there exists a connection between the main verb and the order of its complements. The example sentences in 1 do indicate the different behaviour of the verbs *geben* (to give) and *überlassen* (to leave).

**1 Example Sentences**

*(1)* a. Karl wird dem Kind das Geschenk geben wollen.
Karl will the child the present give want.
Karl will want to give the child the present.

*b.* *Karl wird das Geschenk dem Kind geben wollen.
$\begin{bmatrix}\text{RHEM} & plus \\ \text{CASE} & acc\end{bmatrix}$ $\begin{bmatrix}\text{RHEM} & minus \\ \text{CASE} & dat\end{bmatrix}$

*(2)* a. Karl wird das Geschenk dem Kind überlassen wollen.
Karl will the present the child leave want.
Karl will want to leave the child the present.

*b.* *Karl wird dem Kind das Geschenk überlassen wollen.
$\begin{bmatrix}\text{RHEM} & plus \\ \text{CASE} & dat\end{bmatrix}$ $\begin{bmatrix}\text{RHEM} & minus \\ \text{CASE} & acc\end{bmatrix}$

The *(a)* versions of the sentences do not exhibit any limitation on the order of the complements. The *(b)* versions reflect the fact that they do not allow the first NP to be rhematic in the sense of Lenerz (1977). With the verb *geben* this disallows the order of the accusative rhematic NP preceding the dative not rhematic NP, and in the case of *überlassen* the dative rhematic NP must not precede the accusative not rhematic NP. So the acceptance of the word order of the accusative and dative object seems to depend on the main verb and whether the first of those objects is to be rhematic. To express this in a grammar, one would have to ensure that there exists some connection from the main verb to the complements. We do not go into any detail how this connection could be formulated since it suffices to know that one has to exist. We do give a sketch for a parse tree for one of the example sentences in figure 5; $V_a$ stands for an auxiliary verb, $V_m$ for the main verb and $V_k$ for a verbal complex. This parse tree relies on the analysis of German verb phrases in Hinrichs and Nakazawa (1993).

[6] Particular thanks for this section are due to Frank Richter who listened very patiently to my ramblings about possible occurences of nonlocality concerning word–order in natural language and helped quite a lot getting my intuitions into the present form.

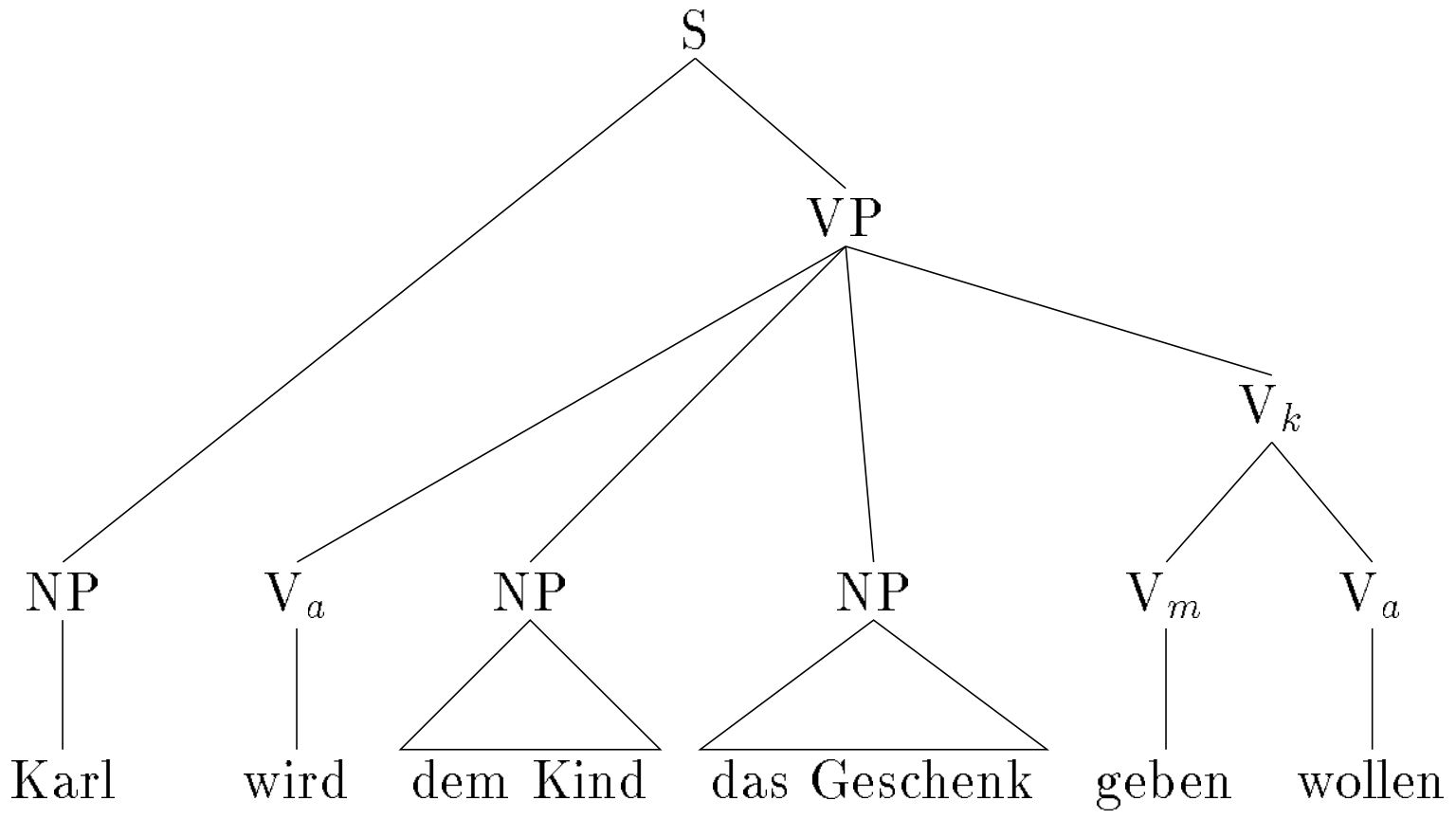

Figure 5: A sketched parse tree for the sentence *Karl wird dem Kind das Geschenk geben wollen.*

As one can see, the NP complements are not in one local tree with the main verb due to the necessary argument raising induced by the auxiliary in the verbal complex. But this is not sufficient to cause the problem presented in the previous section. If one considers direct parsing of such a sentence, the local tree labeled VP in the figure 5 which has to decide on the LP acceptability of the order of the two NP complements is not completed unless the local tree of the verbal complex is completed before. And by this, the necessary connection of the NP complements to the main verb can be validated so that the local tree containing the NP complements would not be completed and therefore the sentence would not be parsed in the non LP acceptable case.

A further complication is necessary to cause the problem – the topicalization of the verbal complex. The data to support the claim is not as sharp as for the non topicalized sentences, although it is plausible that topicalization does not alter the behaviour of the main verbs concerning the word orderof their complements. The examples are given below.

**2 Example Sentences**

*(3)* a. Geben wollen wird Karl dem Kind das Geschenk.

b. *Geben wollen wird Karl das Geschenk dem Kind.

$$\begin{bmatrix}\text{RHEM } \textit{plus}\\ \text{CASE } \textit{acc}\end{bmatrix} \quad \begin{bmatrix}\text{RHEM } \textit{minus}\\ \text{CASE } \textit{dat}\end{bmatrix}$$

*(4)* a. Überlassen wollen wird Karl das Geschenk dem Kind.

b. *Überlassen wollen wird Karl dem Kind das Geschenk.

$$\begin{bmatrix}\text{RHEM } \textit{plus}\\ \text{CASE } \textit{dat}\end{bmatrix} \quad \begin{bmatrix}\text{RHEM } \textit{minus}\\ \text{CASE } \textit{acc}\end{bmatrix}$$

As can be seen in the sketched parse tree in figure 6, again following Hinrichs and Nakazawa (1993), the verbal complex is completely independent of the local tree containing the NP complements. This creates the desired constellation. If one considers right to left traversal of the input string for parsing, as for example done in Ale (Carpenter 1993), the local tree whose mother is labeled VP has to be completed before the verbal complex is parsed. Therefore the information which main verb appears in the verbal complex is not (yet) known. The local tree would be LP acceptable although the whole parse tree may later contain information that the VP in question was not LP acceptable.

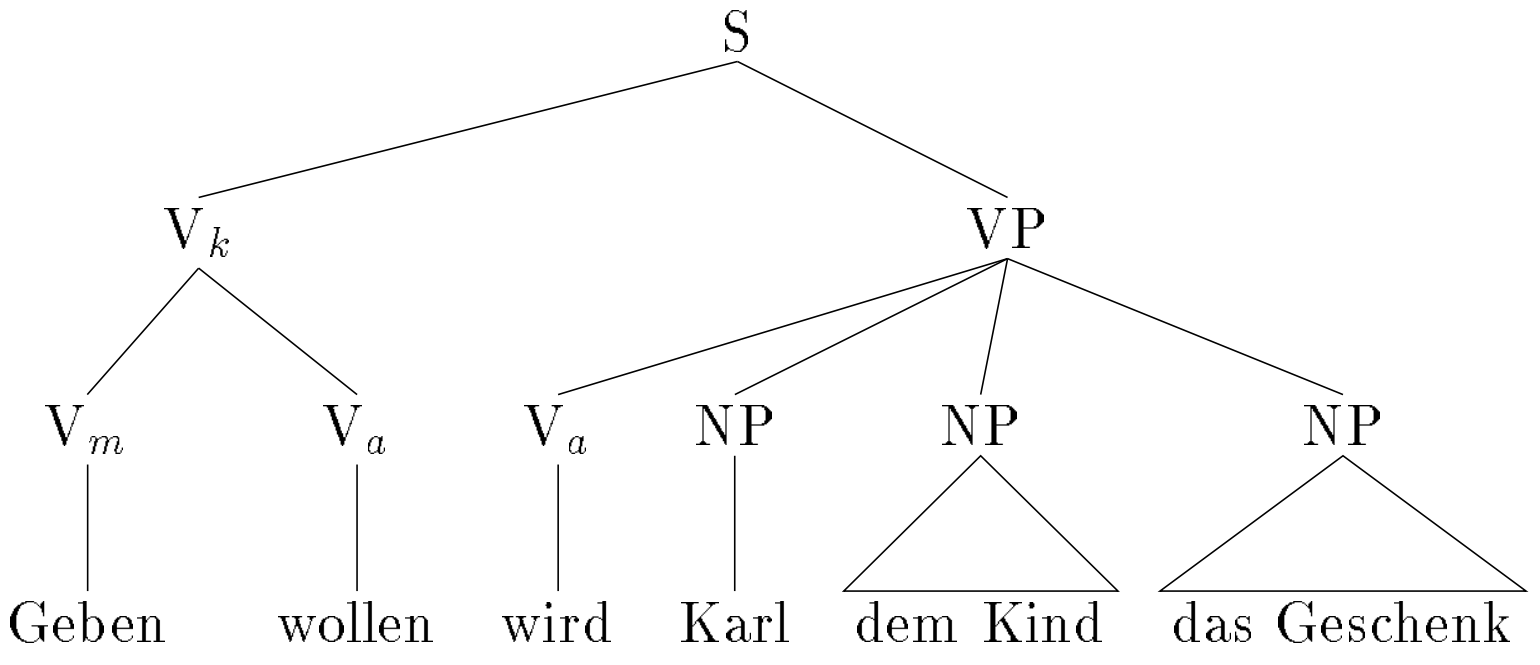

Figure 6: A sketched parse tree for the sentence *Geben wollen wird Karl dem Kind das Geschenk.*

Although the presentation is somewhat sketchy, we showed that in natural language there may well be cases where the technical problem noted by Seiffert does indeed occur. Since all the mechanisms involved are present in current HPSG analyses of German and their interaction is complex, one may conclude that such phenomena are not easily detected by the grammar writer and can not necessarily be avoided. Therefore each parser dealing with unification based ID/LP grammars has to take care of those nonlocal phenomena.

# 4 Feature Graphs

To be able to define typed unification grammars to cope with HPSG style grammars, the concepts of section 2 have to be altered. Seiffert does not provide any definitions of his feature structures. Assuming that he took a view that is close to PATR-II (Shieber 1986), the feature structures used as nonterminals were not typed, at least not as types are understood here. Apart from being demanded by HPSG, the extension to typed feature structures seems desirable for two reasons. Firstly, those typed feature structures allow linguistically relevant concepts like feature cooccurrence restrictions (Gerdemann and King 1993) to be encoded. And, as can be seen in section 6.2.1, the concept of possible LP violation developed there gains in efficiency. For those reasons, this thesis augments the domain of the nonterminals for unification grammars. The first step is to define those typed feature structures, henceforth feature graphs.[7]

Data may be classified according to the properties it possesses. This can be done using features and attributes for those features. Furthermore, one might wish to partition this data in some way by assigning types to classes of this data. This can be done recursively, i.e. the value for a feature may itself be complex. Informally, this descriptions of classifications can be pictured by some sort of labeled feature trees with constraints on subtrees so that distinct paths lead to identical subtrees. The arcs in the tree are labeled by features, the nodes by types. But trees are not suited to the task since the encoding of subtree identity is not supported. Moshier (1993) shows that therefore feature structures are created because they are optimally suited to the task. The particular version of feature structures, called feature graphs, used in this thesis, is defined in this following section.

Since the implementation has been done for the TROLL system (Gerdemann et al. forthcoming) the definitions given in this section follow to some extend those used in the literature on TROLL, in particular those by King (1989), Götz (1994) and King (1994a) who in parts refer to Carpenter (1992). Note that there are ontological differences between the approaches by Carpenter and King. Carpenter uses typed feature structures to model partial information about linguistic entities whereas King uses a description language to interpret the linguistic objects. Nevertheless, he shows in King (1994a) that the modeling level of feature structures may be added without causing any problems. Since this thesis is not concerned with feature logics, possible interpretations or satisfiability of feature graphs, it abstracts away from the definitions necessary for the semantics and presents only those dealing with the syntax, though those on the semantics are more important for any system implementing a logic like TROLL. Furthermore, even the given definitions do not represent the way feature structures are defined in TROLL, but rather present only the first step towards them because the definitions get too complex to be presented in a short section. Readers interested in all the details may look at the references given above. And from the given definitions, extensions in more than TROLL's way are possible. Nothing essential hinges on the use of typed feature structures. If one wants to use non typed feature structures, the definitions in the following sections simplify easily.

[7]The term feature graph has been chosen to avoid confusion with other versions of feature structures defined in the literature. Since the thesis does not deal with the semantics of feature structures, they are in essence graphs, therefore the name feature graphs.

As noted before, the application of this algorithm and the definitions to TROLL are only possible because of the phrase structure backbone TROLL provides and additionally does not meet TROLL's underlying semantic properties, i.e. feature structure interpretations.

## 4.1 The Type Hierarchy

The types used in the feature graphs are ordered in a type hierarchy. This hierarchy expresses the relation of subsumption on the types, i.e. which types are more specific than others.

**3 Definition (type hierarchy)**

*A type hierarchy is a pair* $\langle \mathsf{Type}, \sqsubseteq \rangle$ *such that:*

$\langle \mathsf{Type}, \sqsubseteq \rangle$ *is a finite bounded complete partial order.*[8]

*If* $t, t' \in \mathsf{Type}$ *and* $t \sqsubseteq t'$, *then we say that* $t$ *subsumes* $t'$.

This definition is in effect the same as to define the type hierarchy to be a finite meet semi–lattice as is often done in the literature.[9] For a discussion of the motivations and details of this definition see Carpenter (1992).

$\bot$ (called *bottom*) is the most general type, presented at the bottom of the type hierarchy. Types without further types that are more specific are called *varieties*.

To express the feature cooccurrence restrictions – which seem desirable from

[8] A partial order is a binary relation which is transitive and in addition either reflexive and anti–symmetric or irreflexive and asymmetric. Since we need to refer to properties of relations again later on, we note the standard definitions of mathematical terms used for convenience ($\sim$ an arbitrary symbol used to denote the relation $\mathcal{R}$ on a set $\mathcal{S}$):

transitivity: $x \sim y \in \mathcal{R}$ & $y \sim z \in \mathcal{R} \Rightarrow x \sim z \in \mathcal{R}$
symmetry: $x \sim y \in \mathcal{R} \Rightarrow y \sim x \in \mathcal{R}$
asymmetry: $x \sim y \in \mathcal{R} \Rightarrow \neg(y \sim x) \in \mathcal{R}$ This is anti–symmetry and irreflexivity.
anti–symmetry: $x \sim y \in \mathcal{R}$ & $y \sim x \in \mathcal{R} \Rightarrow x = y$
reflexivity: $\forall x \in \mathcal{S} : (x \sim x) \in \mathcal{R}$
irreflexivity: $\neg \exists x \in \mathcal{S} : (x \sim x) \in \mathcal{R}$
non–reflexivity: $\neg$ reflexive & $\neg$ irreflexive.

An order is bounded complete iff for every set of elements (which may be empty or infinite) that has an upper bound, there exists a join. This kind of order always has a least element.

The join or least upper bound $\bigsqcup \mathcal{S}$ of a set $\mathcal{S}$ is defined such that $\forall y \in \mathcal{S}\ y \sqsubseteq \bigsqcup \mathcal{S}$ and $\forall z$ (such that $y \sqsubseteq z$ for all $y \in \mathcal{S}$) $\bigsqcup \mathcal{S} \sqsubseteq z$. Intuitively the first condition ensures that $\bigsqcup \mathcal{S}$ is an upper bound and the second one that $\bigsqcup \mathcal{S}$ is the least upper bound.

[9] The necessary mathematical definitions for finite meet semi–lattice can be found in chapter 11 of Partee, ter Meulen and Wall (1990).

a linguistic viewpoint – it is necessary to define which types deserve which features and in turn which types those features deserve, though this thesis will not go in more detail how this can be done.

**4 Definition (appropriateness specification)**

*If* $\langle \mathsf{Type}, \sqsubseteq \rangle$ *is a type hierarchy, then an appropriateness specification over* $\langle \mathsf{Type}, \sqsubseteq \rangle$ *and a finite set* $\mathsf{Feat}$ *of feature names is a partial function* $Approp : \mathsf{Type} \times \mathsf{Feat} \rightharpoonup \mathsf{Type}$ *from type-feature pairs to types such that:*

*if* $Approp(t, \text{F})$ *is defined and* $t \sqsubseteq t'$
*then* $Approp(t', \text{F})$ *is defined and* $Approp(t, \text{F}) \sqsubseteq Approp(t', \text{F})$

These concepts are further developed for TROLL for semantic reasons, but those changes do not need to concern the progress in this thesis since they do not alter the way the types are defined, but rather define ways to give them the appropriate meanings.

Type hierarchies will be presented graphically together with their appropriateness specifications by indicating subsumption by arrows and appropriateness specifications in brackets. The most general type is displayed at the bottom of the type hierarchy, the most specific types at the top. To avoid confusion, we will not talk about supertypes or subtypes since these terms are used in too many different setups in the literature. For example, the set $\{\bot, a, b, c, d\}$ with the following subsumption relation $\{\langle \bot, a\rangle, \langle \bot, b\rangle, \langle \bot, c\rangle, \langle \bot, d\rangle, \langle a, b\rangle, \langle a, c\rangle\}$ and appropriateness specification $\{\langle b, \text{F}, \bot\rangle, \langle c, \text{F}, d\rangle\}$ is represented as in figure 7.

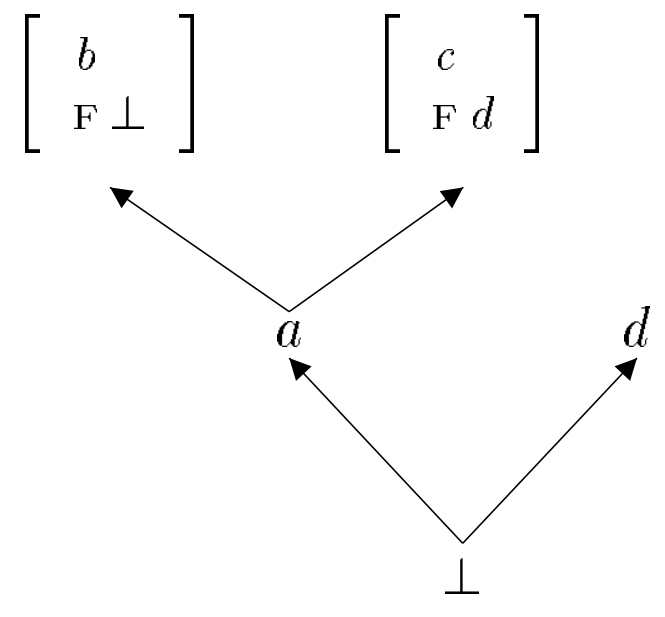

Figure 7: An example for the graphic representation of a type hierarchy

## 4.2 The Syntax of Feature Graphs

In this section the syntax of feature graphs is given as finite state automata following Carpenter (1992) based on definitions by Moshier (1988). Note that again only the preliminary (easy) definition is given which is changed for semantic and formal reasons in TROLL. For an exhaustive representation of the development from the definition given here to the “real” TROLL feature struc-tures see Götz (1994). The definition allows to picture feature graphs as finite, rooted, connected and directed graphs with types as labels for the nodes and features as labels for the arcs.

**5 Definition (feature graph)**

*Let* $\langle \mathsf{Type}, \sqsubseteq \rangle$ *be a type hierarchy,* Feat *a set of features, then a feature graph is a quadruple* $\langle Q, \overline{q}, \theta, \delta \rangle$ *such that*

> $Q$ *is a finite set of states,*
>
> $\overline{q} \in Q$ *is a distinguished start state,*
>
> $\theta : Q \rightarrow \mathsf{Type}$ *is a type assignment function,*
>
> $\delta : Q \times \mathsf{Feat} \rightharpoonup Q$ *is a transition function, and*
>
> *if* $q \in Q$ *then for some* $n \in \mathbb{N}$*, some* $q_0, \ldots, q_n \in Q$*, and some* $\mathrm{F}_1, \ldots, \mathrm{F}_n \in \mathsf{Feat}$
>
> > $q_0 = \overline{q}$,
> >
> > $q_n = q$, *and*
> >
> > *for each* $i < n$, $\delta(q_i, \mathrm{F}_{i+1})$ *is defined and* $\delta(q_i, \mathrm{F}_{i+1}) = q_{i+1}$.

The last condition ensures that if a state, i.e. a node, is defined, it has to be reachable by a path of features from the root node. This covers the intuition that the underlying feature tree has to be connected.

This definition allows feature graphs to be different, if they differ in their state set, but have exactly the same type and feature labelings. Therefore an equivalence relation ($\sim$) which groups those feature graphs together which differ only in their state sets can be defined, see for example Moshier (1993). Following on that, feature graphs are regarded *modulo* this equivalence relation.[10]

To get the most benefit from using typed feature structures as will become clear in section 6, we demand additionally that feature graphs conform to the appropriateness specification, i.e. that they are well–typed. The concept of well–typing has been developed by Carpenter (1992). This expresses that whenever a feature is present, the value has to be of an appropriate type, i.e. it has to be at least as specific as the most general appropriate type. And every feature has to be defined for a type for which it is appropriate. This does not induce any conditions on features that are not present. In the following, feature graphs are always assumed to be well–typed unless stated otherwise.

**6 Definition (well–typed)**

*Let* $F = \langle Q, \bar{q}, \theta, \delta \rangle$ *be a feature graph.*

> $F$ *is well–typed iff for every* $q \in Q$,
> *if* $\delta(q, \mathrm{F})$ *is defined then* $Approp(\theta(q), \mathrm{F})$ *is defined and*
> $Approp(\theta(q), \mathrm{F}) \sqsubseteq \theta(\delta(q, \mathrm{F}))$

---

[10] If $\bowtie$ is an equivalence relation, the notation for the equivalence class over the set $\mathcal{S}$ is:

$$[s]_{\bowtie} = \{r \in \mathcal{S} | r \bowtie s\} \text{ and } \mathcal{S}/_{\bowtie} = \{[s]_{\bowtie} | s \in \mathcal{S}\}$$

denotes the quotient set of $\mathcal{S}$ *modulo* $\bowtie$. An equivalence relation is transitive, reflexive and symmetric.

Since even the equivalence classes of automatons as representation of feature graphs is sometimes cumbersome, attribute value matrices (AVMs) are used later on in this thesis to picture feature graphs. In the AVMs as used in this thesis, the feature graphs are presented in brackets. The types are in the left upper corner, the features are written in small capital letters and reentrant paths are indicated with boxed integers where the value is written behind all boxes, i.e. by variables. An example for all three ways of presentation is given in figure 8.

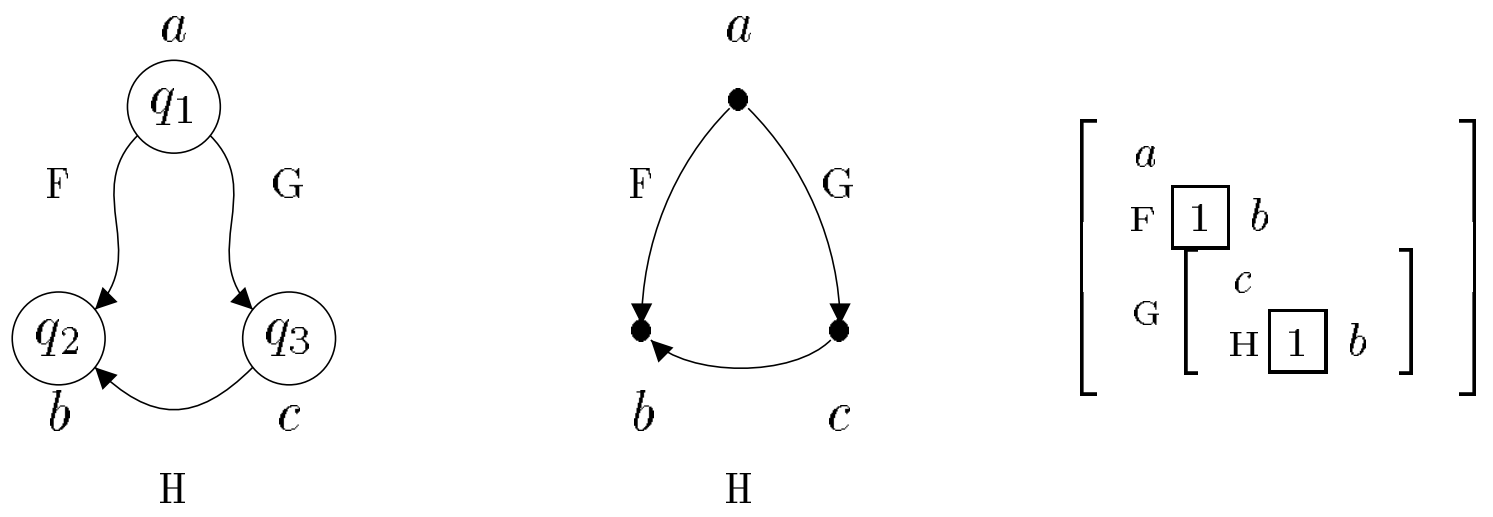

Figure 8: An example for the graphic representation of a feature graph as an automaton, a graph and an AVM

## 4.3 Subsumption

Since the goal of this thesis is the presentation of a chart parser which is able to parse grammars of the ID/LP format, and LP acceptability is determined via subsumption, the definition of subsumption is crucial. Naturally subsumption may be important for a chart parser for reasons not connected with LP acceptability as well (see section 6), but the main motivation here is the need to deal with the LP rules.

Since feature graphs are not total, they are used to represent partial information, i.e. we have to be able to tell in which cases different feature graphs can be related. Subsumption is a relation on the feature graphs that holds if one is in some sense more specific than another. This presupposes that all the types of the general one subsume the types in the specific one as specified in the type hierarchy, and all the features of the general one have to be present in the specific one. If one were to take a semantic denotational approach, then the denotation of the more general one is a superset of the denotation of the more specific feature graph (Götz 1994).

This definition is taken from Carpenter (1992) and presupposes a closed world assumption.[11] This closed world assumption means that the varieties partition the entire domain of objects, i.e. there can not be an object that is not of one of the types specified and no object can be of two varieties.

[11]Carpenter has a further limitation on his type hierarchies, namely the feature introduction condition which is of no relevance here since it can be eliminated without loss of any of the desired properties (King and Götz 1993).

**7 Definition (subsumption ($\sqsubseteq$))**

*Let* $\langle \mathsf{Type}, \sqsubseteq \rangle$ *be a type hierarchy,* $\mathsf{Feat}$ *a (finite) set of features, then* $FS_1 = \langle Q_1, \bar{q}_1, \theta_1, \delta_1 \rangle$ *subsumes* $FS_2 = \langle Q_2, \bar{q}_2, \theta_2, \delta_2 \rangle$*,* $FS_1 \sqsubseteq FS_2$*, iff there is a total function* $h : Q_1 \rightarrow Q_2$ *such that*

$h(\bar{q}_1) = \bar{q}_2$

*for every* $q_1 \in Q_1$, $\theta_1(q_1) \sqsubseteq \theta_2(h(q_1))$

*for every* $q_1 \in Q_1$ *and every* $\text{F} \in \mathsf{Feat}$

*if* $\delta_1(q_1, \text{F})$ *is defined then* $\delta_2(h(q_1), \text{F})$ *is defined and* $h(\delta_1(q_1, \text{F})) = \delta_2(h(q_1), \text{F})$

Note that the same symbol ($\sqsubseteq$) is used for type subsumption and feature graph subsumption, but it should not be a problem to determine from the context which one is meant.

Since the construction of an interesting example would take some space to define the necessary type hierarchy and appropriateness restrictions and we assume the reader to be somewhat familiar with subsumption and feature structures, we just give a trivial example in figure 9 using the type hierarchy from figure 7.

$$[a] \quad \sqsubseteq \quad \begin{bmatrix} c \\ \text{F}\ d \end{bmatrix}$$

Figure 9: A simple example for subsumption

## 4.4 Unification

In a unification based parser it is essential to tell when two nonterminals match, i.e. when a rule may be applied. This is done by unification. This definition is taken from Carpenter (1992), but was modified by Paul King. Again, we do not present TROLL's way of handling unification because it would necessitate explaining the whole theory behind TROLL which does not seem necessary to provide the reader with an idea what is meant.

Intuitively, unification represents a way of conjoining compatible information about (two) feature graphs. The result is supposed to be unique (modulo alphabetic variants), not to contain any additional information, i.e. information that was not present in at least one of the feature graphs involved must not appear, and to contain as least as much information as was contained in the original feature graphs. Inconsistent information leads to failure. Intuitively, if unification is viewed procedurally, one starts with the respective root types, unifies them and recursively traverses the feature graphs and unifies the substructures that are reachable by identical features. The results are the labeling for the resulting feature graph. Those features that appear only on one feature graph have to be added to the result as well. This is done until no new features and types can be added. The unification of types is done by calculating the join of the types in question by using the type hierarchy.

**8 Definition (unification ($\sqcup$))**

*Suppose $FS_1$ and $FS_2$ are feature graphs such that*

$FS_1 = \langle Q_1, \bar{q}_1, \theta_1, \delta_1 \rangle$,

$FS_2 = \langle Q_2, \bar{q}_2, \theta_2, \delta_2 \rangle$, *and*

$Q_1 \cap Q_2 = \emptyset$.

*Let $\bowtie$ be the smallest equivalence relation such that*

$\bar{q}_1 \bowtie \bar{q}_2$, *and*

*if* $q_1 \bowtie q_2$, $(\delta_1 \cup \delta_2)(q_1, \mathrm{F}) = \tilde{q}_1$ *and* $(\delta_1 \cup \delta_2)(q_2, \mathrm{F}) = \tilde{q}_2$ *then* $\tilde{q}_1 \bowtie \tilde{q}_2$.

$FS_1 \sqcup FS_2$ *is defined iff for each* $x \in (Q_1 \cup Q_2)/_{\bowtie}$, $\{(\theta_1 \cup \theta_2)(q) \mid q \in x\}$ *has an upper bound.*[12]

*If* $FS_1 \sqcup FS_2$ *is defined then* $FS_1 \sqcup FS_2 = \langle (Q_1 \cup Q_2)/_{\bowtie}, [\bar{q}_1]_{\bowtie}, \theta^{\bowtie}, \delta^{\bowtie} \rangle$, *where*

$\theta^{\bowtie}(x) = \bigsqcup \{(\theta_1 \cup \theta_2)(q) \mid q \in x\}$,

$\delta^{\bowtie}(x, \mathrm{F})$ *is defined iff for some* $q \in x$, $(\delta_1 \cup \delta_2)(q, \mathrm{F})$ *is defined,*

*if* $\delta^{\bowtie}(x, \mathrm{F})$ *is defined then* $\delta^{\bowtie}(x, \mathrm{F}) = \bigcup \{[(\delta_1 \cup \delta_2)(q, \mathrm{F})]_{\bowtie} \mid q \in x\}$.

Note that we overload the symbol for unification ($\sqcup$). It will be used later in the thesis with and without a result, i.e. $FS_1 \sqcup FS_2 = FS_3$ meaning that $FS_1$ and $FS_2$ unify with result $FS_3$ and $FS_1 \sqcup FS_2$ meaning that the unification of the two feature graphs succeeds.

Again, we are just presenting a simple example assuming the type hierarchy from figure 7, see figure 10.

$$\begin{bmatrix} a \\ \mathrm{F}\; d \end{bmatrix} \sqcup \begin{bmatrix} c \\ \mathrm{F}\; \bot \end{bmatrix} = \begin{bmatrix} c \\ \mathrm{F}\; d \end{bmatrix}$$

Figure 10: A simple example for unification

Carpenter shows that the result is again what is here called a feature graph. But it is important to note that it can not be concluded that the resulting feature graph is well–typed, i.e. all features and types are appropriate[13], since the definition does not contain anything which deals with the appropriateness specifications. But since it is desired for the result to be well–typed, Carpenter explicitly gives a partial function (*TypInf*) which ensures the well–typing of the result. In TROLL the problem does not arise since the disjunctive resolved feature structures defined there are closed under unification (see also section 7.1), therefore we do not discuss Carpenter's *TypInf* in any detail, but note that the well–typed feature graphs can be closed under unification by some further inferencing. Carpenter proves as well that the result is the most general feature graph that contains the information of both feature graphs.

[12] This information is drawn from the type hierarchy.

[13] For an example see Carpenter (1992).

# 5 Typed Unification ID/LP grammars

In this section we extend the concept of context free grammars to typed unification grammars and those to typed unification ID/LP grammars. The major change is that the nonterminals, and maybe even the terminals as explained below, are no longer atomic but rather feature graphs. The feature graphs represent the linguistic categories. Clearly the nonterminals become a nonfinite domain by this. The extension utilizes the operation of unification from section 4.4 on them, to be able to tell when the nonterminals match in some sense, i.e. when a rule may be applied. Similarly we need the concept of subsumption (see section 4.3) to tell when an LP rule applies. We give a derivation type of definition for both types of grammars along the lines standardly given for context free grammars following Shieber (1984) and Seiffert (1991). A different approach in defining unification grammars is taken in Carpenter (1992) more along the lines taken in formal language theory, in Gerdemann (1991) in terms of admissibility conditions and in Sikkel (1993) in terms of decorated trees or constraint sets.

Recall that to define a context free grammar, one can proceed by describing the language generated by the grammar (Aho and Ullman 1972). This can be done for context free grammars by constructing a derivation according to inference rules from the start symbol (or axiom) to the terminal string (or theorem). The language is defined as all strings derivable from the start symbol.

## 5.1 Typed Unification Grammars

Before starting with the important definitions, one auxiliary concept needs to be defined. Intuitively the concatenation of the elements of a set form a sequence. We use this definition to simplify the following definitions in such a way that we can talk about a sequence of nonterminals thereby abstracting from the need for an exact formalization of the realization of said sequences in terms of multiply rooted feature graphs or some system inherent types and features. Note that the definition as given here does not allow structure sharing between the elements (feature graphs) forming the sequence. We come back to this later.

**9 Definition (sequence)**

*A sequence* $\sigma = \sigma_1 \ldots \sigma_n$ *from a (multi) set* $\rho$, $|\rho| = n$ *is the concatenation of the elements of* $\rho$ *in an arbitrary ordering.*
$\rho^*$ *denotes the set of sequences of all elements of* $\wp_m(\rho)$.[14]

Now the definition of a typed unification grammar can be given. It is very similar to the original definition for a context free grammar, only that we need more information to deal with the types. A context free grammar is standardly defined as consisting of a set of terminals, a set of nonterminals, a set of context free rules and a distinct nonterminal, the start symbol.

[14] $\wp_m(X)$ denotes the multiset counterpart of the power set of $X$.

**10 Definition (typed unification grammar (TUG))**

*A typed unification grammar is a sextuple* $G = \langle T, NT, TH, AS, UR, SFS \rangle$ *over sets* Type *and* Feat *where*

> $T$ *is the set of terminals*
>
> $NT$ *is the set of nonterminals such that each element is a feature graph conforming to* $TH$ *and* $AS$ *and* $T \cap NT = \emptyset$
>
> $TH$ *is a type hierarchy,* $\langle \mathsf{Type}, \sqsubseteq \rangle$
>
> $AS$ *is an appropriateness specification over* $\langle \mathsf{Type}, \sqsubseteq \rangle$ *and* Feat
>
> $UR$ *is the set of unification based phrase structure rules, i.e. ordered pairs* $\langle \alpha, \beta \rangle$ *with* $\alpha \in NT$ *and* $\beta$ *a sequence from an element from* $\wp_m(NT \cup T)$*. For notational convenience we will write them as* $FS_0 \rightarrow FS_1 \ldots FS_n$ *where* $FS_0 \in NT$ *and* $\forall i \; i > 0 \; FS_i \in (NT \cup T)$
>
> $SFS$ *is an arbitrary feature graph from* $NT$*, the start feature graph.*

The differences to a context free grammar are immediately obvious. Since a context free grammar is a quadruple, we have here two additional components. Firstly the type hierarchy and secondly the appropriateness specification as explained in section 4.1. The grammar depends on both for the encoding of the properties of the feature descriptions and for unification and subsumption. Furthermore, the nonterminals are not longer atomic, but complex feature graphs. This leads to a nonfinite domain of nonterminals. Nevertheless those nonterminals that appear in the grammar can be specified. But it is no longer possible to specify all nonterminals that may appear in the derivations. Instead they have to unify with those appearing in the grammar.

It seems clear that in parsing one wants as input a sentence of terminals that consist of atomic words. In the case of HPSG, one could view those atomic terminals as representing in some sense the phonology. On the other hand, they could be feature graphs themselves as long as they are distinct from the nonterminals. We do not force the terminals to be atomic here since it is not really necessary and it is desired to give the most general definition possible. The theoretic construct of feature structures can naturally be used for all purposes. Their interpretation is the crucial question for systems concerned with feature logics. We do not interpret them in any way, but use them just as a syntactic means for the definition of the nonterminals and terminals. Therefore this thesis does not make the attempt to resolve any of the ontological questions involved, but tries to cover a broad range of possible views. HPSG takes the view, as already discussed in section 1, that sort resolved and totally well–typed feature structures model the linguistic objects. And those feature structures are themselves described using AVMs. From section 4 on the feature graphs it is clear that we do not pursue any alternative way to deal with these descriptions, but rather treat them as a notational variant of the feature graphs. Apart from this limitation, the definition given here allows the other views described above. Atomic words as terminals could stand for the actual object words which are modeled by feature graphs and therefore somehow connected to them, whereas structured terminals may be seen as a way of treating the feature structures as being closer to the linguistic objects themselves. We do not make a difference between terminals and preterminals. But if one wants, it is easy to treat what has been called structured terminals above as preterminals which stand in a special relation to the "real" terminals or the phonology respectively. The parser and all the processes described later would in this case operate on the preterminals, indirectly influencing the terminals, instead of on the terminals themselves. For a more extensive discussion of ontological questions see Meurers (1994).

The empty word in unification or HPSG based systems is represented by either a feature graph with empty phonology or an atomic terminal attached to a feature graph, both representing the empty string, depending on the view taken concerning the discussion above. But apart from the empty phonology, there may well be several empty categories with differences between them, so that for example NP and VP traces can be distinguished. They have to be included among the terminals since they do not make sense on the left hand side of rules.

The unification rules allow both terminals and nonterminals to occur on the right hand sides with structure sharing between them (if they are feature graphs). The definition here is somewhat imprecise because it uses the auxiliary concept of a sequence to describe the unification rules instead of presenting a rigorous formalization. Since structure sharing is a local operation on feature graphs, it is not as simple as in the given definition for unification rules. To do it properly, one can pursue two alternatives. For both approaches it is necessary that the whole rule has to be treated as one feature graph and then structure sharing is allowed between parts of this feature graph. The first alternative is to encode the rule with special types and features. For example, by having a type *rule* with features *lhs* and *rhs*, *lhs* demanding a feature graph as value and *rhs* a list of terminals and nonterminals, i.e. feature graphs. This makes it necessary to have some grammar independent types, features and appropriateness specifications which one might or might not like. Gerdemann (1991) shows how the definitions have to be for the class of unification grammars. The concept can easily be augmented for the typed unification grammars discussed here. Or alternatively, as proposed in Sikkel (1993), one can define a special kind of feature graph, namely multiply rooted feature graphs which are allowed for the rules only. So, the categories of the right hand side of a rule are all standard feature graphs which are rooted in this kind of composite feature graph. The definitions given here abstract away from both approaches since it is not a problem to change the definitions accordingly if one wishes to do so, but the gain in clarity and readability is obvious. The notation used is close to the approach using multiply rooted feature graphs.

In the following, we will not present the nonterminals in detail any more since they are not really necessary. Instead, we will assume that any feature graph can be a nonterminal, apart from the ones which serve as terminals or empty categories.

We wish to define the language generated by the grammar in such a way that we let the language be the set of all sentences given by the derivation from the start feature graph to all possible sequences of terminals. Therefore it is necessary to define a single step in the derivation, here called derives directly. The idea is simply that a feature graph may be replaced by the right hand side of a rule, if the left hand side of the rule unifies with the feature graph. In context free grammars this happened if the nonterminal on the left hand side of the rule was the same as the one in the sequence. The subscripts are used to identify the feature graphs positionally.

**11 Definition (derives directly ($\Rightarrow$))**

*Let* $G = \langle T, NT, TH, AS, UR, SFS\rangle$ *be TUG and all* $FS_k, FR_j$ *are feature graphs or terminals, then*

$$\begin{array}{llll} & FS_1 \ldots FS_{i-1} & FS_i & FS_{i+1} \ldots FS_n \\ \Rightarrow & FS'_1 \ldots FS'_{i-1} & FR'_1 \ldots FR'_m & FS'_{i+1} \ldots FS'_n \end{array}$$

*iff* $FR_0 \rightarrow FR_1 \ldots FR_m$ *is in* $UR$ *and*
$FS_i \sqcup FR_0$ *thereby instantiating all other* $FS_i, FR_j$ *further to* $FS'_i, FR'_j$ *via structure sharing.*

Since there is no special map between terminals and nonterminals defined, the definition of derives directly does not demand a special treatment of the terminals. This generality is due to the fact that all terminals have to occur explicitly in the rules. If one wants to define this map between terminals and nonterminals, it can be mimicked in this formalism by having unary rules linking a nonterminal to a terminal. Naturally no terminal must occur on the left hand side of a rule so that no rewriting of a terminal can occur. Note that terminals neither unify with each other nor with feature graphs.

By unifying the left hand side of the rule with a feature graph from the sequence, all the other elements of the sequence may change as well via structure sharing. This is well defined since we are dealing in reality with either a single feature graph or a multiply rooted feature graph. This is indicated by having dashed versions of all the feature graphs after the '$\Rightarrow$'. We assume that in all further definitions, this is implictly understood.

The extension of this concept to a derivation of arbitrary length is done by iterating the previously defined one step derivation.

**12 Definition (derives ($\Rightarrow^*$))**

*Let* $G = \langle T, NT, TH, AS, UR, SFS\rangle$ *be a TUG ,* $B$ *is a feature graph or a terminal,* $\alpha, \beta, \gamma, \alpha', \gamma', \mu_i$ *are (possibly empty) sequences of feature graphs or terminals, then*

$\alpha B \gamma \Rightarrow^* \alpha' \beta \gamma'$

*iff* $\alpha B \gamma = \alpha' \beta \gamma'$ *or*
$\alpha B \gamma \Rightarrow \mu_1 \Rightarrow \ldots \Rightarrow \mu_n \Rightarrow \alpha' \beta \gamma'$.

As can be seen in the definition, a derivation can consist of any number of steps, including zero.

Now all auxiliary concepts to define the language of the grammar have been introduced.

**13 Definition (Language of the Grammar ($L(G)$))**

*Let $G = \langle T, NT, TH, AS, UR, SFS \rangle$ be a TUG, then the language generated by $G$ is defined as*

$$L(G) = \left\{ w = w_1 \ldots w_n \;\middle|\; \begin{array}{l} SFS \Rightarrow^* w \text{ and} \\ \forall i \; 1 \leq i \leq n \; w_i \in T \end{array} \right\}$$

Since the definition in section 4.4 of the operation of unification ensures monotonicity, the result will not change if we do apply the derivation steps in a different order, i.e the derivation is independent of the order of the single derivation steps. The possibility of different derivations by a different order of the derivation steps should not be confused with real ambiguity, i.e. different parses because of the grammar. If we construct equivalence classes of the derivations under this order independence, then a parse tree is a way of displaying one of those equivalence classes.

A simple example for such a TUG with the language generated is given in figure 11.

The language of this grammar is the set containing only the sentence *she walks*. For example, the sentence **I walks* is not in the language. A parse tree for *she walks* looks as displayed in figure 12. Naturally this is only a toy grammar which does not represent any linguistic facts. Nevertheless it shows to some extent how phenomena like agreement could be handled by structure sharing.

As can be seen, the given definitions lead to derivations where the root of the parse tree does no longer need to be identical to the start symbol as it was the case with context free grammars, but rather has to be subsumed by it. The same is true for all the nonterminals which appear in the rules and the respective ones in the parse tree.

## 5.2 Typed Unification ID/LP Grammars

In this section we extend the typed unification grammars by splitting up the information of linear precedence and immediate dominance contained in the unification rules to separate sets of ID and LP rules, thereby allowing the desired linguistic generalizations.

To keep the definitions as parallel as possible to the ones in the previous section, some complications arise later on. In Seiffert (1987) another way to define the language generated by the grammar has been taken which seems more elegant in terms of having less definitions. It utilizes a new class of grammars where the set of grammar rules corresponds to all possible permutations of the rules in the ID/LP version. Nevertheless, the way of defining the language

Let $G = \langle T, NT, TH, AS, UR, SFS\rangle$ be TUG with

$$T = \{I, she, walks\}$$

$TH + AS$ =

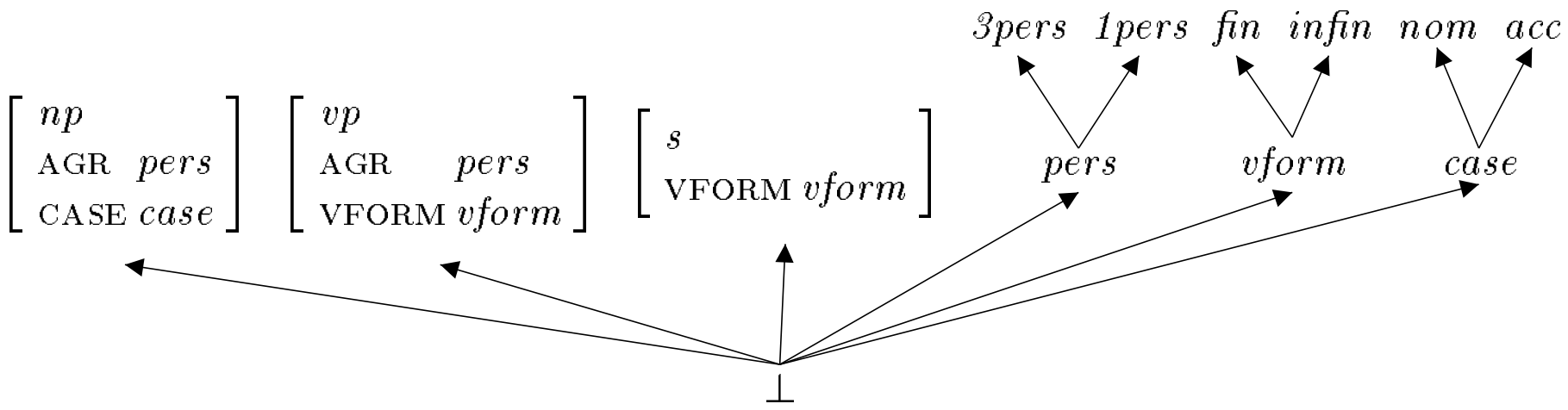

$$UR = \left\{ \begin{array}{lcl} \begin{bmatrix} s \\ \text{VFORM } \boxed{1}\ vform \end{bmatrix} & \rightarrow & \begin{bmatrix} np \\ \text{AGR } \boxed{2}\ pers \\ \text{CASE } nom \end{bmatrix} \begin{bmatrix} vp \\ \text{AGR } \boxed{2}\ pers \\ \text{VFORM } \boxed{1}\ vform \end{bmatrix} \\ \begin{bmatrix} np \\ \text{AGR } 1pers \\ \text{CASE } nom \end{bmatrix} & \rightarrow & \text{I} \\ \begin{bmatrix} np \\ \text{AGR } 3pers \\ \text{CASE } nom \end{bmatrix} & \rightarrow & \text{she} \\ \begin{bmatrix} vp \\ \text{AGR } 3pers \\ \text{VFORM } fin \end{bmatrix} & \rightarrow & \text{walks} \end{array} \right\}$$

$$SFS = \begin{bmatrix} s \\ \text{VFORM } vform \end{bmatrix}$$

$$L(G) = \{she\ walks\}$$

Figure 11: An example for a TUG

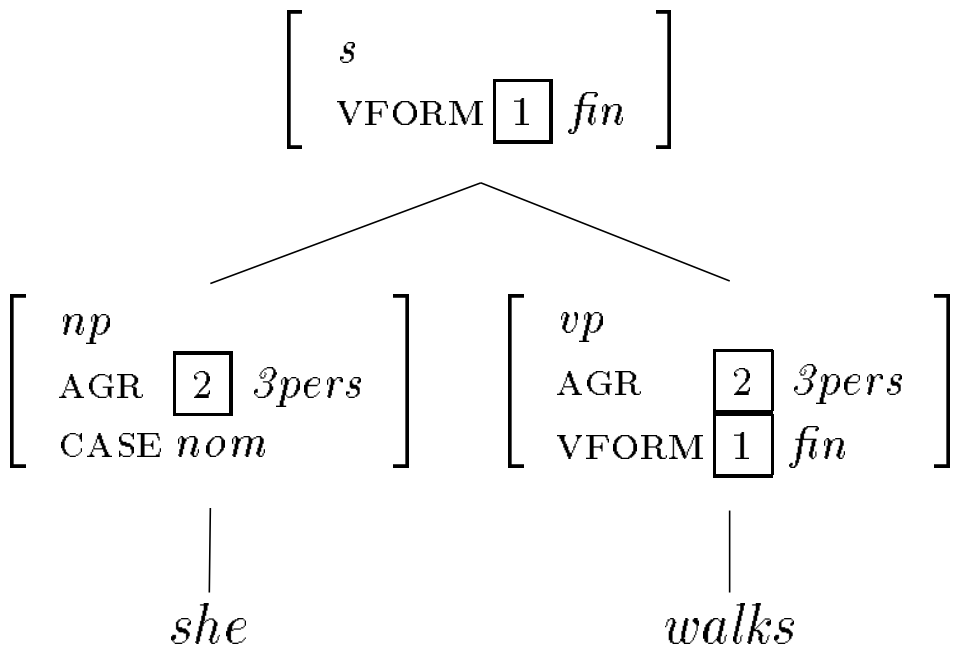

Figure 12: A parse tree for *she walks*

of a typed unification ID/LP grammar taken here seems preferable since the problem with the nonlocal flow of information can be seen more clearly.

ID/LP format is standardly defined on local trees, but it seems more precise to think of LP acceptability on sequences of daughters since the mother and the information on immediate dominance connected with it do not play a role for determining linear precedence. To avoid confusion, the next definition relates the concepts of local trees and sequences, though it is not strictly necessary for any of the following definitions. Additionally, the notion of a sequence for a set was defined in the previous section. It should be clear from the context which one is meant.

**14 Definition (sequence)**

*A sequence $\sigma = d_1 \ldots d_n$ generated from a local tree $\tau =$*

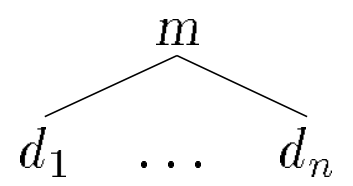

*is the concatenation of all the daughters of $m$ preserving their respective orderings.*

Parallel to the definition in the previous section runs the development of the necessary formalism to define typed unification ID/LP grammars and the language generated by a given typed unification ID/LP grammar.

First we define the components of a typed unification ID/LP grammar.

**15 Definition (typed unification ID/LP grammar (TU ID/LP G))**

*A septuple $G = \langle T, NT, TH, AS, IDR, LPR, SFS \rangle$ defines a typed unification grammar over sets* Type *and* Feat *where*

> *$T$ is the set of terminals*
>
> *$NT$ is the set of nonterminals such that each element is a feature graph conforming to $TH$ and $AS$ and $T \cap NT = \emptyset$*
>
> *$TH$ is a type hierarchy, $\langle \mathsf{Type}, \sqsubseteq \rangle$*
>
> *$AS$ is an appropriateness specification over $\langle \mathsf{Type}, \sqsubseteq \rangle$ and* Feat
>
> *$IDR$ is the set of unification ID rules, i.e. ordered pairs $\langle \alpha, \beta \rangle$ with $\alpha \in NT$ and $\beta$ an element from $\wp_m(NT \cup T)$. For notational convenience we will write the unification rules as $FS_0 \rightarrow FS_1, \ldots, FS_n$ where $FS_0 \in NT$ and $\forall i\ 1 \leq i \leq n\ FS_i \in (NT \cup T)$*
>
> *$LPR$ is the set of unification linear precedence rules where each rule has the form $FS_1 \prec FS_2$ with $FS_1, FS_2 \in (NT \cup T)$. Take $LP$ to be the set of all the $FS_i$ occurring in $LPR$, then $\langle LP, \prec \rangle$ is a transitive, asymmetric, irreflexive relation.*
>
> *$SFS$ is an arbitrary feature graph from $NT$, the start feature graph.*

The components apart from the unification rules remain unchanged. The set of the unification rules is now split into two parts – the immediate dominance rules and the rules on linear precedence as desired. The right hand side of a rule is not longer a sequence, but a multiset. Note also the change in the notation: the elements of the right hand sides of the ID rules are now separated by commata meaning that their order is not fixed, i.e. they do not directly constitute a sequence. According to the view in Gazdar et al. (1985) the LP rules are defined in such a way that they constitute an asymmetric, irreflexive and transitive relation. For a more detailed discussion of those conditions see (partly) section 6.2.2 or Meurers and Morawietz (ms). Note that it has not been proposed by any linguist to have an LP rule which has a structure sharing between the two LP elements. If this possibility would have to be incorporated, analog to the comment on the rules, it would necessitate for each LP rule to be a single feature graph to formalize this properly.

Again, we are going to neglect the nonterminals in the following with the same reservations as in the previous section.

Additionally to the concept of a one step derivation and a derivation, it is necessary to clarify when a sequence is acceptable under the LP rules. To determine this, we need to define when an LP rule actually applies to a sequence. This is done in the following definition. The idea is that an LP rule applies to a sequence, if the information in the sequence is as least as specific as in the LP rule. This means that in some sense two feature graphs from a sequence match the feature graphs in the LP rule.

**16 Definition (applies)**

*An LP rule* $FS_1 \prec FS_2$ *applies to a sequence* $\sigma = \sigma_1 \ldots \sigma_n$, $\sigma_k \in (NT \cup T)$ *iff there exist* $i, j$, $i \neq j$, $1 \leq i \leq n$, $1 \leq j \leq n$ *such that*

$FS_1 \sqsubseteq \sigma_i$ *and*
$FS_2 \sqsubseteq \sigma_j$.

Naturally the empty sequence and each sequence consisting of just one element are LP acceptable since no LP rule applies to them. A problem concerning this definition are the atomic terminals. If the terminals are not feature graphs (as they usually are not) or we are dealing with preterminals rather than with the terminals directly, the subsumption is not defined, unless the terminals are included in the type hierarchy. And they have to occur explicitly in the LP rules as well, since otherwise they would be ordered freely because no LP rule would apply. Or alternatively, one would have to define an extra subsumption relation for the terminals. We do not force a choice here, but it seems easier to demand the inclusion of the terminals in the type hierarchy. All this is not an issue for practical purposes, since implementations like Troll usually do not allow terminals to occur in the rules.

The idea in the definition of LP acceptability is simply that no LP rule is violated, i.e. applies to the sequence in the specific setup with the second element preceding the first one.

**17 Definition (LP acceptable)**

*A sequence* $\sigma$ *is LP acceptable iff no LP rule applies to it with* $j < i$.

For the definition of derives directly it is necessary to define which sequences are possible permutations from the right hand side of a rule. This is done in two steps. Firstly all permutations are allowed and then they are limited to the LP acceptable ones.

The function *permute* constructs all permutations from a set of feature graphs by nondeterministically choosing one element from the set followed by the permutation of the remaining set. The input is a multiset of feature graphs, the output a set of the permutations of all sequences of the feature graphs from the input multiset.

**18 Definition (permute)**

*Let* $G = \langle T, NT, TH, AS, IDR, LPR, SFS\rangle$ *be a typed unification ID/LP grammar,* $\alpha$ *a multiset of feature graphs or terminals,* $\beta$ *a sequence of feature graphs or terminals and* $A$ *a feature graph or terminal, then the function* $permute : \wp_m(NT \cup T) \rightarrow \wp((NT \cup T)^*)$ *is defined as*

$$permute(\emptyset) = \emptyset$$
$$\forall \alpha \in \wp_m(NT \cup T)$$
$$permute(\alpha) = \left\{ A\beta \;\middle|\; \begin{array}{l} A \in \alpha \text{ and} \\ \beta \in permute(\alpha \backslash A) \end{array} \right\}$$

The function *expand* uses the definition of *permute* to construct permutations and then limits them to those that are LP acceptable. The input is a multiset of feature graphs, the output a set of LP acceptable sequences of the feature graphs from the input multiset.

**19 Definition (expand)**

*Let* $G = \langle T, NT, TH, AS, IDR, LPR, SFS\rangle$ *be a typed unification ID/LP grammar and* $\alpha$ *a multiset of feature graphs or terminals,* $\beta$ *a sequence of feature graphs or terminals, then the function* $expand : \wp_m(NT \cup T) \rightarrow \wp((NT \cup T)^*)$ *is defined as*

$$\forall \alpha \in \wp_m(NT \cup T)$$
$$expand(\alpha) = \left\{ \beta \;\middle|\; \begin{array}{l} \beta \in permute(\alpha) \text{ and} \\ \beta \text{ is LP acceptable} \end{array} \right\}$$

Note however that this does not rule out the cases where no LP rule applies at the moment but may apply later because there has been some nonlocal feature passing. This leads to a definition of derivation that does not exclude all cases of non LP acceptable sequences, i.e. local trees. So this is the point where the problem can be seen most clearly. There will be no try to incorporate the proposed treatment of it here, since it would confuse the definitions and it is nevertheless possible to give a definition for the language generated by a TU ID/LP grammar, even if it contains some redundancy.

Now all necessary auxiliary terms have been defined, so we can proceed with the definition of derivation parallel to the one in the previous section.

Note that the right hand side of the rule is not directly substituted for the nonterminal, but rather an LP acceptable sequence generated by it. The ambiguity introduced by having different LP acceptable sequences from a multiset is a real one introduced by the grammar.

**20 Definition (derives directly ($\Rightarrow$))**

*Let $G = \langle T, NT, TH, AS, IDR, LPR, SFS \rangle$ be a typed unification ID/LP grammar and all $FS_k, FR_j$ are feature graphs or terminals, $\alpha$ a multiset of feature graphs or terminals then*

$$\begin{array}{llll} & FS_1 \ldots FS_{i-1} & FS_i & FS_{i+1} \ldots FS_n \\ \Rightarrow & FS'_1 \ldots FS'_{i-1} & FR'_1 \ldots FR'_m & FS'_{i+1} \ldots FS'_n \end{array}$$

*iff* $FS_i \sqcup FR_0$ *and*
$FR_0 \rightarrow \alpha$ *is in* $IDR$ *and*
$FR_1 \ldots FR_m \in expand(\alpha)$.

After all those preliminary definitions the next is the same as in the previous section.

**21 Definition (derives ($\Rightarrow^*$))**

*Let $G = \langle T, NT, TH, AS, IDR, LPR, SFS \rangle$ be a typed unification ID/LP grammar, $B$ is a feature graph or terminal, and $\alpha, \beta, \gamma, \alpha', \gamma', \mu_i$ are (possibly empty) sequences of feature graphs or terminals, then*

$$\alpha B \gamma \Rightarrow^* \alpha' \beta \gamma'$$

*iff* $\alpha B \gamma = \alpha' \beta \gamma'$ *or*
$\alpha B \gamma \Rightarrow \mu_1 \Rightarrow \ldots \Rightarrow \mu_n \Rightarrow \alpha' \beta \gamma'$.

Since the definitions so far do not suffice to conclude that all derivations are LP acceptable, it becomes necessary to fall back on the concept of a parse tree to enable the definition of the language generated by a TU ID/LP grammar. It has already been given informally in the last section. A parse tree represents an equivalence class of derivations, namely those that differ only in the order of the rule applications, but do not assign a different structure to the input string. Naturally, leaves have to be terminals and the order of the daughters is crucial.

It is easier to introduce the more general notion of a tree first and to extend it afterwards to the notion of a parse tree. The definitions are modified versions from those given in Partee et al. (1990) and therefore not constructive ones, but rather admissibility conditions.

The concepts developed until now allow rules of a format such that only a single nonterminal is rewritten[15], therefore it is possible to straightforwardly define trees generated by a grammar, i.e. parse trees. The following paragraphs contain somewhat trivial concepts, nevertheless they are stated here so that any possible confusion is avoided.

The information displayed graphically in a tree consists of dominance, precedence and labeling information. Dominance is a relation between two nodes. Two nodes stand in this relation if they are connected by a sequence of branches. Since branches are directional, a node $a$ dominates a node $b$ in case there exist branches from $a$ to $b$. Whenever two nodes are directly con-nected by a branch they stand in the relation of immediate dominance. Clearly dominance is a transitive relation, since whenever the nodes $a$ and $b$ and $b$ and $c$ are in it, then $a$ and $c$ are also connected by some branches. Usually it is assumed that every node dominates itself, i.e. the relation is reflexive. Finally, dominance is antisymmetric, since two nodes have to be the same if they dominate each other. Otherwise cycles would be allowed. This constitutes a weak partial order. A node which is not dominated by any other node is called a root. A node that does not dominate any other node is called a leaf. Nodes that are immediately dominated by the same node are called sisters. And finally, a node immediately dominated by another one is called a daughter of that node.

[15] In contrast to for example context sensitive rules.

Precedence is a relation that exists between every pair of nodes that does not stand in the dominance relation,i.e. dominance and precedence partition $N \times N$. Intuitively speaking it constitutes the order of the tree. Precedence is as well transitive, but irreflexive which follows directly from the property of being reflexive of the dominance relation and the condition that precedence exists only between elements that do not stand in the dominance relation. If $x$ precedes $y$, then $y$ can not precede $x$, i.e. precedence is asymmetric. This constitutes a strict partial order.[16]

Since distinct nodes may have identical labels, it is obvious that we can not treat the nodes as the labels themselves. Therefore a labeling function is introduced which has as the domain the nodes and as range a set of labels.

**22 Definition (tree)**

*A tree is a quintuple* $\langle N, Q, D, P, L\rangle$ *where*

> $N$ *is a finite set, the set of nodes*
>
> $Q$ *is a set, the set of labels*
>
> $D$ *is a weak partial order in* $N \times N$*, the dominance relation*
>
> $P$ *is a strict partial order in* $N \times N$*, the precedence relation*
>
> $L$ *is a function from* $N$ *into* $Q$*, the labeling function*

*such that*

> $\exists x \in N\ \forall y \in N\ \langle x, y\rangle \in D$ *and*
>
> $\forall x, y \in N\ (\langle x, y\rangle \in P \vee \langle y, x\rangle \in P) \leftrightarrow (\langle x, y\rangle \notin D \wedge \langle y, x\rangle \notin D)$ *and*
>
> $\forall w, x, y, z \in N\ (\langle w, x\rangle \in P \wedge \langle w, y\rangle \in D \wedge \langle x, z\rangle \in D) \rightarrow (\langle y, z\rangle \in P)$.

The first condition ensures that only a single root is permitted. Generally this is not necessary, but in linguistics it seems a compulsory assumption. The second condition ensures what we informally already introduced above. Only elements that do not stand in the dominance relation are allowed to stand in the precedence relation. The third condition excludes cases where we have crossing branches and those cases where we have nodes with more than one branch entering it, i.e. where $\langle x, y\rangle$ and $\langle z, y\rangle$ are both in $D$. Since all elements that are connected by branches have to appear in $D$, this excludes crossing branches between different local trees as well. If the acceptability of a tree is solely defined via local trees as is the case in Gazdar et al. (1985), this is easily overlooked. The same is true for nodes with multiple branches entering. Thus our definition limits the general mathematical notion of a tree to the relevant cases of trees.

[16] A strict partial order is a binary relation which is transitive, irreflexive and asymmetric.

Now the notion of a tree and a TU ID/LP grammar are connected to yield the notion of a parse tree, again in terms of admissibility conditions. Recall from the beginning of section 5.1 that the rules were not presented entirely correct. To enable a sound definition of structure sharing, they would have to be contained in a single feature graph or multiply rooted feature graph respectively. The same applies here. The applied rules now represent local trees in the parse tree, but the structure sharing between them is still valid. Therefore one would have to encode the parse tree in a feature graph. Again Gerdemann (1991) by using a single feature graph or Sikkel (1993) by using a multiply rooted feature graph show how it is done properly for a slightly different class of grammars. The presentation here abstracts away from it as done previously.

**23 Definition (parse tree)**

*A grammar* $G = \langle T, NT, TH, AS, IDR, LPR, SFS \rangle$ *generates a tree – called a parse tree – iff*

> *the root is labeled with a feature graph* $M$ *which is subsumed by the initial symbol, i.e.* $SFS \sqsubseteq M$ *and*
>
> *the sequence* $\sigma = \sigma_1 \ldots \sigma_n$ *of the leaves of the tree ordered according to the precedence relation is a sequence of terminals, i.e.* $\forall i \; 1 \leq i \leq n \;\; \sigma_i \in T$ *and*
>
> *for all subtrees* $\tau = \begin{array}{c} FS_0{}' \\ \diagup \quad \diagdown \\ FS_1{}' \;\; \ldots \;\; FS_n{}' \end{array}$ *in the tree, where* $FS_0{}'$ *immediately dominates* $FS_1{}' \ldots FS_n{}'$*, there is a rule* $FS_0 \rightarrow \alpha$ *in* $IDR$ *such that*
>
> > $FS_1 \ldots FS_n \in permute(\alpha)$ *and*
> > $\forall i \; 1 \leq i \leq n \;\; FS_i \sqsubseteq FS_i{}'$ *and*
> > *the sequence generated from* $\tau$ *is LP acceptable.*

Note that the precedence relation in the second condition is the one from the previous tree definition and is not directly connected with the relation of linear precedence defined in $LPR$. In the third condition, the right hand side of the rule does not need to be in the exact order as the sequence generated from $\tau$. Nevertheless there has to be a licensing rule. This is achieved by having the sequence generated from $\tau$ being member of the possible permutations of the right hand side of a rule whose left hand side category subsumes the mother of $\tau$. LP acceptability is enforced on all local trees in the last condition. The formulation of the second condition is sufficient because all the leaves of a tree are totally ordered according to the precedence relation. A formal proof of this statement can be found in Partee et al. (1990).

The definition of a local tree corresponds to the one for subtrees implicit in the above definition.

This last definition is nearly the same as in the previous section – and would be the same if the problem with nonlocal feature passing did not exist – except for an additional clause which reflects the need to exclude all the local trees that are not longer LP acceptable, but could not be ruled out at the time the direct derivations took place. This is done by demanding that all sentences have a valid parse tree under the above definition which ensures the LP acceptability of all the local trees.

**24 Definition (Language of the Grammar ($L(G)$))**

*Let $G = \langle T, NT, TH, AS, IDR, LPR, SFS \rangle$ be a typed unification ID/LP grammar, then the language generated by G is defined as*

$$L(G) = \left\{ w = w_1 \ldots w_n \;\middle|\; \begin{array}{l} SFS \Rightarrow^* w \text{ and} \\ \text{there exists a parse tree } \tau = \\ \langle N, Q, D, P, L \rangle \text{ generated by } G \\ \text{such that } w_1 \ldots w_n \text{ are the leaves} \\ \text{of } \tau \text{ ordered according to } P \end{array} \right\}.$$

This definition would have been the same if derives directly would use *permute* instead of *expand*, i.e. all permutations of the feature graphs on the right hand side would be accepted. It would lead to all possible derivations without limitations through LP acceptability. The second clause above would nevertheless ensure LP acceptability for the sentences of the language by excluding those derivations that contain a non LP acceptable local tree. But the exposition here mirrors more closely the way the algorithm will work and defines the problem to some extend more formally. Note that this definition does not rule out those cases where a derivation and a parse tree exist, but a more specific instance of the parse tree would contain a not LP acceptable local tree. We return to this question in section 6. So the aim for the algorithm will be to get rid of the second clause of definition 24 by obliterating the need for a separate checking of the local trees of the resulting parse tree by incorporating it into the derivation steps. The definition of the language of a TU ID/LP grammar is correct though.

A simple example for such a TU ID/LP grammar with the language generated is given in figure 13. It is nearly the same as the one given in the previous section in figure 11. The changes appear in the split of the unification rules into immediate dominance and linear precedence rules. Note that there is now a comma between the elements on the right hand side of a rule.

The language of this TU ID/LP grammar is the same as the one given for the TUG in the previous section and we have the identical parse tree. But note however that this time the string **walks she* is excluded because of the given LP rule and not because there is no unification rule allowing it, as in the

Let $G = \langle T, NT, TH, AS, IDR, LPR, SFS \rangle$ be TU ID/LP G with

$$T = \{I, she, walks\}$$

$TH + AS =$

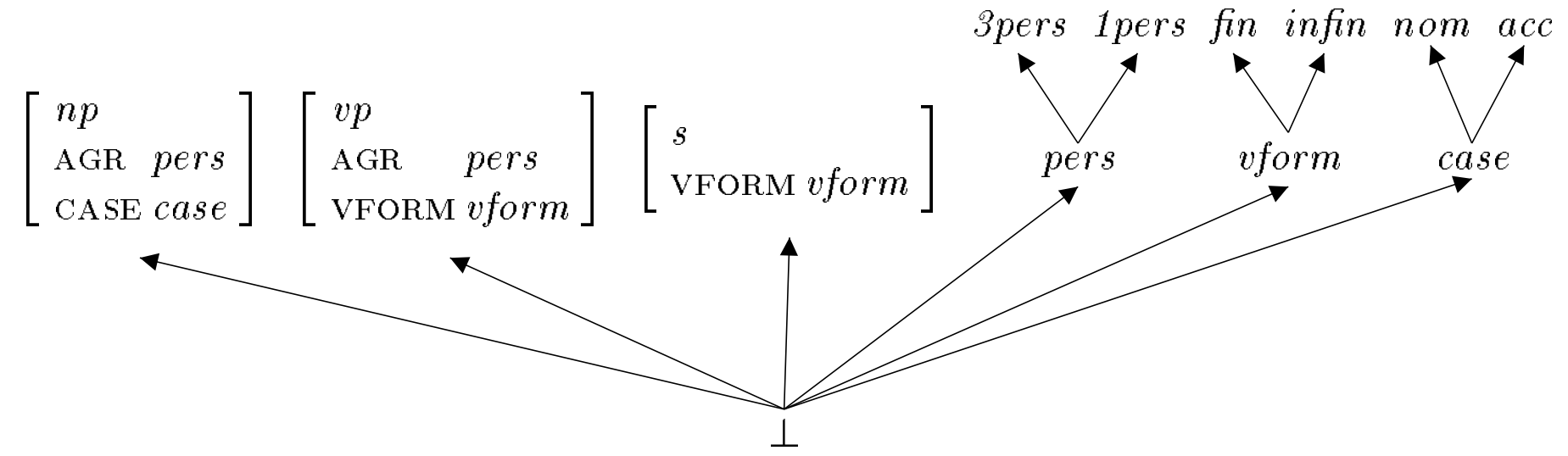

$$IDR = \left\{ \begin{array}{lll} \begin{bmatrix} s \\ \text{VFORM} \boxed{1}\ vform \end{bmatrix} & \rightarrow & \begin{bmatrix} np \\ \text{AGR} \boxed{2}\ pers \\ \text{CASE}\ nom \end{bmatrix}, \begin{bmatrix} vp \\ \text{AGR} \boxed{2}\ pers \\ \text{VFORM} \boxed{1}\ vform \end{bmatrix} \\ \begin{bmatrix} np \\ \text{AGR}\ 1pers \\ \text{CASE}\ nom \end{bmatrix} & \rightarrow & \text{I} \\ \begin{bmatrix} np \\ \text{AGR}\ 3pers \\ \text{CASE}\ nom \end{bmatrix} & \rightarrow & \text{she} \\ \begin{bmatrix} vp \\ \text{AGR}\ 3pers \\ \text{VFORM}\ fin \end{bmatrix} & \rightarrow & \text{walks} \end{array} \right\}$$

$$LPR = \left\{ \begin{bmatrix} np \\ \text{CASE}\ nom \end{bmatrix} \prec \begin{bmatrix} vp \\ \text{VFORM}\ fin \end{bmatrix} \right\}$$

$$SFS = \begin{bmatrix} s \\ \text{VFORM}\ vform \end{bmatrix}$$

$$L(G) = \{she\ walks\}$$

Figure 13: An example for a TU ID/LP G

previous section. If we eliminate the LP rule, the resulting language of $G$ is the set containing the sentences *she walks* and *walks she* with the parse trees given in figure 14.

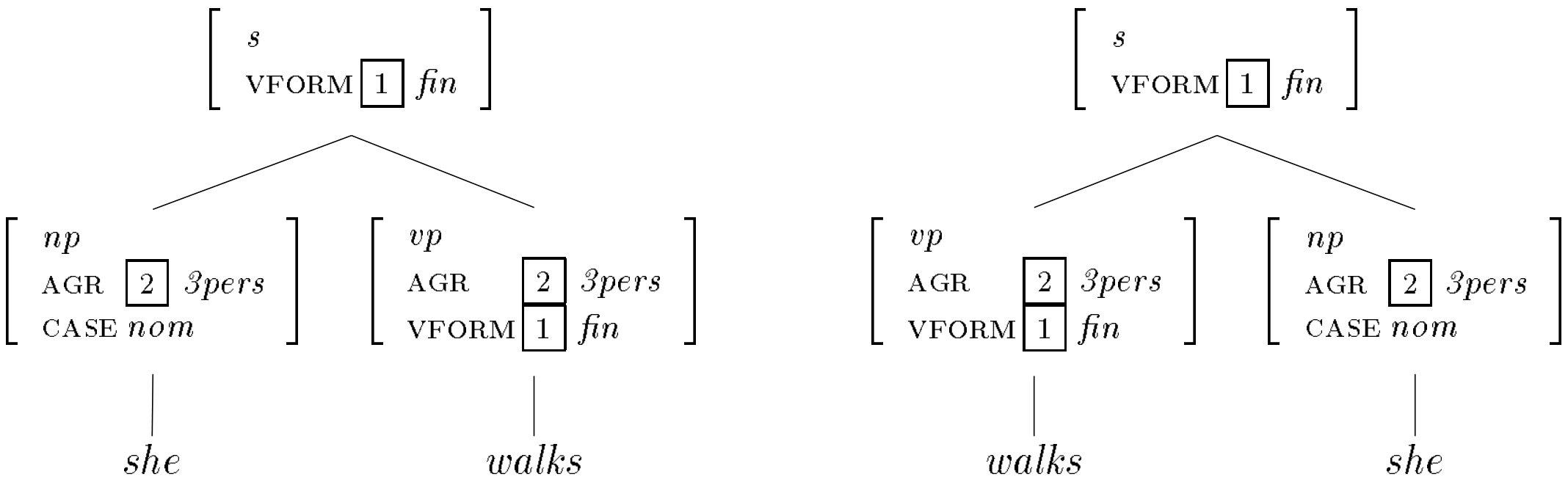

Figure 14: Two parse trees for the TU ID/LP grammar in figure 13

# 6 An Algorithm for Parsing TU ID/LP Grammars

After having described the problem of nonlocal feature passing and having formalized the class of typed unification ID/LP grammars, we now present in this section of the thesis the actual algorithm to decide whether a given string is a member of a defined language. The algorithm is explained with some examples and its complexity is discussed briefly. But before the exact formalization of the algorithm, the idea for solving the problem of the nonlocal flow of information is presented informally.

## 6.1 The Idea

If one would follow the definition of the language of a TU ID/LP grammar, one could just take a usual parser that allows all the permutations of the right hand side of the rules and later filter those parse trees which contain not LP acceptable local trees. In the approaches discussed in section 2.2 on direct parsing of ID/LP grammars, it can be seen that it is preferable to interleave the generator, i.e. *permute*, and the tester, i.e. LP acceptability, to achieve better performance. But as could be seen in my analysis of Seiffert's approach in the same section, his algorithm still has two steps to ensure LP acceptability of the parses since the first step allows some invalid ones due to nonlocal feature passing. Therefore he has to filter them in a second step. He gains on the brute force method described above though, because he rules out as many wrong analyses as possible as early as possible. In the following algorithm, we will interleave these two steps completely so that acceptance can be determined in just one traversal of the input. To achieve this, we mimic to some extent techniques employed in the implementation of freeze or delay primitives for programming languages such as for example SICStus Prolog (Carlsson and Widén 1988) or Prolog II (Giannesini, Kanoui, Pasero and Caneghem 1985). When during parsing the situation arises that the parser cannot decide whether the encountered structure is LP acceptable or not, it stores the relevant environment and resumes action as if the structure would have been LP acceptable. If later any new information is added to the frozen environment, the parser resumes action at the point where it left off and decides LP acceptability. Since this can only happen in case some new information is added from another edge[17], this addition is not allowed, thereby preventing the parser from constructing invalid structures. The old structure is not discarded since it may be used in some other way which does not lead to an LP violation.

Augmenting Seiffert's (Seiffert 1991) and Shieber's (Shieber 1984) definitions in the following section, we will define a relation that tells when an LP rule might *weakly apply* to a sequence. Whenever this occurs, we add a pair of sequences of feature graphs to the store. This store may be thought of as information how further instantiation of the rule in question may be restricted.

[17]This is an edge constructed from a chart parser and not an edge of a feature graph.

The restriction is not immediately obvious, but as soon as the rule is further instantiated a test for LP acceptability is performed and the hypothesis discarded if the structure contained an LP violation. The first element of this kind of pair is structure shared with the sequence the parser assumes to be LP acceptable to be able to continue computation. The second element is a copy of this sequence. This means that it is exactly the same, but without any connection to existing structures, especially without any structure sharings apart from those local to the copy. Since any changes that are made to the sequence via structure sharing are eminent in the first element of this pair, it is easy to tell whether anything has changed by comparing it with the copy. This is done by the function $dif$. The parser tests for an occurrence of such a difference and when it occurs, checks LP acceptability on the new sequence, i.e. the one that has changed. If it is not LP acceptable, the edge is not constructed. If it is LP acceptable or possibly LP violated, the parser continues the processing. Instead of using this construction with $dif$, the algorithm could just store only the structure shared sequence and test it always blindly for LP acceptability. But since the test for LP acceptability is very costly, it seems preferable to do this double storage although it takes up more space. Additionally, the algorithm has to percolate the LP stores. This percolation has to ensure that all the edges do contain the pending LP information of their subcomponents. Since they are sets, the LP stores are unioned when two edges are combined.

## 6.2 Preliminaries

In this section the presentation of the actual algorithm is further prepared by giving some additional definitions, explaining some necessary preliminary steps and providing some information about the notation used.

### 6.2.1 Additional Definitions

The definitions in this section are motivated to cope with the problem of nonlocal feature passing. In this sense, it will extend the definitions given in section 5 to deal with LP acceptability directly in the derivation process, even if there is a nonlocal feature passing, though no new complete set of definitions is given. The way they are used can be seen directly in the algorithm.

The definition of possible violation is given to detect the cases of nonlocal feature passing. This may occur in those cases where the elements of a sequence are (not yet) specific enough to allow to determine LP acceptability. Recall from definition 16 on application of an LP rule that this was done via subsumption. What we need to know is whether at some later stage of processing the information could possibly be there. To achieve this, a test unification with the categories and the LP elements takes place. It suffices if one of the two categories of the proposed sequence matches only by unification even if the other one matches by the stronger condition of subsumption. If a feature graph is subsumed by an element of an LP rule, it unifies with it as well. So we can not exclude the cases of application of an LP rule from our definition. This leads to a complication of the following definition of possible LP violation.

**25 Definition (weakly applies)**

*An LP rule $FS_1 \prec FS_2$ weakly applies to a sequence $\sigma = \sigma_1 \ldots \sigma_n$ iff there exist $i, j$, $i \neq j$, $1 \leq i \leq n$, $1 \leq j \leq n$ such that*

$FS_1 \sqcup \sigma_i$ *and*

$FS_2 \sqcup \sigma_j$.

Here the usage of well–typed feature structures gains efficiency since the unifications are dependent on the types and the appropriateness restrictions so that only a limited number of them succeeds. In a system which employs nontyped feature structures, this approach leads to a large number of unnecessarily stored pending LP information because a lot of the unifications succeed without there being a possibility for the structure to be extended in that way by the grammar. A prototype of this algorithm has been implemented for UNICORN (Gerdemann and Hinrichs 1988) and even for a very simple example grammar, namely the grammar Seiffert illustrates nonlocal feature passing with, it turned out that a number of features and attributes had to be introduced to avoid this spurious application of the LP rules to almost all proposed sequences. But this is clearly not desired and no option for the implementation of a reasonably large fragment.

Since we can tell when there might be enough information to determine LP acceptability at some later stage of the processing by weak application, it is straightforward to define possible LP violation. It presupposes that the structure is LP acceptable under definition 17, i.e. that no LP rule applies to it and is violated, because in that case we do not need to check for further (possible) violations since the sequence is ruled out anyway. And additionally, the weak application has to result in a violation if it is to be important at all.

**26 Definition (possibly LP violated)**

*A sequence $\sigma$ is possibly LP violated iff it is LP acceptable, but an LP rule weakly applies to it with $j < i$.*

These two definitions are used in the algorithm to ensure that at places where we may run into the discussed problem, we construct such a pair as explained in section 6.1 to monitor them and come back to them if necessary.

As discussed above, the pairs are not really essential to the algorithm, but improve somewhat on efficiency. The algorithm could blindly try to find whether an element of the store is now specific enough by checking LP acceptability on each of the elements in each step and, whenever one is found, do the appropriate steps (or rather not do anything, as will become clear from the algorithm). Clearly this would need less storage space. But to make this more efficient (though there may be a tradeoff between time and space) and to stay closer to the implementation of delay primitives, the concept of difference is introduced. To tell when actually more information on the objects in the store is available, the store consists of pairs of sequences where the first one is structure shared with the sequence that is used in the processing, whereas the other one is a copy of the original one at the moment of the construction of the pair. So just by comparing them, one can tell when a new check for LP acceptability is appropriate. The comparison is defined as follows, the input is a pair of sequences, the output a truth value.

**27 Definition (dif)**

*Let* $G = \langle T, NT, TH, AS, IDR, LPR, SFS \rangle$ *be a typed unification ID/LP grammar,* $(\sigma, \tau)$ *be a pair of sequences of feature graphs,* $t$ *and* $f$ *standing for true and false respectively, then the function*
$dif : \wp((NT \cup T)^*) \times \wp((NT \cup T)^*) \rightarrow \{t, f\}$ *is defined as*

$$dif(\sigma,\tau) = \left\{ \begin{array}{ll} t & \text{if } \sigma \not\doteq \tau \\ f & \text{if } \sigma \doteq \tau \end{array} \right\}$$

The '$\doteq$' sign in the definition above is meant to indicate that the two structures are identical. Since we are not dealing with the semantics, we can not simply say that the two structures have to be semantically equivalent. Instead, this difference is implemented neither as literal identity nor as term unification, but rather as mutual subsumption. This abstracts from the problems involved with the internal representations and ensures equality on a purely syntactic level. It is the closest we can get to the identity of the two structures without refering explicitly to the semantics.

### 6.2.2 Some Comments

Before the algorithm is presented, it seems necessary to provide the reader with some more comments, namely on the transitivity condition of the LP rules and some changes in the predictor compared to Shieber's algorithm.

In definition 15 on TU ID/LP grammars, the LP rules are defined in such a way that the underlying set of the occuring feature graphs and the operator constitute a transitive, asymmetric and irreflexive relation. To care for the transitivity is somewhat problematic. One might consider situations where we are faced with two feature graphs to be ordered and there are no LP rules that apply directly. But if an intermediate feature graph would be present, two LP rules would apply in such a way as to enforce an ordering on the first two. To give an example (not with feature graphs, but it just serves to illustrate the point): the task is to order $A$ and $B$, no LP rule applies. But if $A$, $B$, and $C$ would have to be ordered, the LP rules $A \prec C$ and $C \prec B$ would apply. The algorithm would allow $A$ and $B$ to occur in both orders. If one wishes $A$ and $B$ only to occur in the fixed order $AB$ the transitive closure of the LP rules would have to be computed beforehand. This can be done by modifications of standard closure algorithms as for example Warshall's algorithm given in Prolog in O'Keefe (1990). But this does indeed depend on how one wants to treat transitivity, i.e. whether transitivity is just supposed to apply if indeed all components are present or if transitivity has to apply in all cases. More on this can be found in Meurers and Morawietz (ms).

The check for LP acceptability has been omitted from the predictor. The inclusion of the test would rule out some edges. But as Seiffert notes, the test is costly at this place since there may not be enough information present. The approach pursued here has another reason for omitting the check. As explained above, it utilizes the notion of structure sharing to keep track of changes of the elements on the LP store. To include the test for possible LP violation in the predictor would mean to construct a sequence instead of merely keeping track of it, since one would have to check on a proposed sequence. And nothing would be gained by the storing of this proposed sequence, because only the completer does indeed test for changes of the LP store. Naturally, the predictor could do this as well, but since the rules may be rather underspecified, not much could be gained. This construction could be done by structure sharing both parts of the proposed sequence and checking on that, but it would become more complicated than in the completer without being really necessary, because the information and the structure sharing would no longer be local to an edge, but rather involve two edges. On a quick look this may not be fatal, but if one reconsiders structure sharing and realizes that it is local to one feature graph the problem becomes clear immediately. Furthermore the exclusion of the edges does not lead to an overall performance improvement. As discussed in Morawietz (in preparation) the inclusion of the test interacts with other mechanisms of the parser in such a way that neither the number of edges nor the parsing time is lower compared to the version without the test for LP acceptability in the predictor.

### 6.2.3 Notation

To help the reader with the algorithm, we give an overview on the notation used.

Format for the edges:
$[\ Start\ ,\ End\ ,\ Lhs \rightarrow BeforeDot \bullet AfterDot\ ,\ LPStore\ ]$

$Start$, $End$, $h$, $i$, $j$, $k$, $l$, $m$, $n \in \mathbb{N}$.

$Lhs$, $A$, $B$, $C$, $D$ are feature graphs. $***$ is a distinct symbol for a feature graph.

$w_h$ is the terminal at the position $h$ in the input.

$BeforeDot$, $\alpha$, $\beta$, $\gamma$, $\sigma$, $\tau$ are (possibly empty) sequences of feature graphs.[18]
$[]$ is the empty sequence.

$\alpha A$ is the concatenation of the sequence $\alpha$ with the feature graph $A$.

$AfterDot$, $\eta$, $\zeta$ are (possibly empty) multisets of feature graphs.

$LPStore$, $S$, $S1$, $S2$ are (possibly empty) sets of pairs $(\sigma, \tau)$ of sequences of feature graphs.

[18] Either a single feature graph or a multiply rooted feature graph (see section 5).

All dashed versions of feature graphs are supposed to stand for the same structure, but possibly modified because of a unification, i.e. $A'$ is the same as $A$, but after some unification has taken place, possibly resulting in an alteration of $A$; the same holds for sequences. It is always the case that $A \sqsubseteq A'$.

Boxed integers represent structure sharing introduced by the grammar, boxed lowercase letters structure sharing introduced by the algorithm. Note that whole sequences of feature graphs are shared with those reentrancies introduced by the algorithm.

## 6.3 The Algorithm

The parsing algorithm presented here is a modification of Earley's algorithm (Earley 1970) along the lines of Shieber (1984) and Seiffert (1991). The paper by Earley describes a parser and a recognizer. Since there exist algorithms for extracting all parse trees from the parse lists with some encoding (for example Graham, Harrison and Ruzzo (1980)), we can abstract away from the distinction and can view the recognizer in some sense as a parser. Strictly speaking, we just present a recognizer, but this distinction is somewhat irrelevant with unification based systems. Since the concept of a context free grammar is augmented to a TU ID/LP grammar, there is another possibility available for generating a parse structure. It is convenient, if not necessary (cf. Meurers (1994)), to encode the syntactic and semantic structure in the feature graphs themselves in such a way that the grammar constructs the parse and all necessary information is contained in the maximal left hand side category, i.e. the root of the parse tree.

Input to Earley's algorithm is a context free grammar and a string. The grammar is a quadruple consisting of the terminals, nonterminals, the production system (the rules) and the start symbol. The algorithm consists of three major parts. A predictor, a completer and a scanner. The predictor proposes hypotheses in a top down fashion according to the rules. The algorithm works from left to right, predicting always on the leftmost nonterminal. The completer combines already existing information to yield (partially) verified hypotheses. The scanner matches parts of the hypotheses against the input and therefore provides a bottom up component. More exactly, the algorithm constructs $n$ parse lists for an input of length $n$. Each list can contain several items, so called dotted productions. Each of those items reflects a part of the input which has already been recognized and which part is still to be found to yield a certain nonterminal, provided a rule exists in the production system which allows this constellation. To explain this more closely, consider the following situation. On input $\sigma = \sigma_1 \ldots \sigma_n$, the parse lists $L_0$ to $L_{i-1}$ are already constructed, i.e. currently $L_i$ is being build. Then, if the rule $A \rightarrow \alpha\beta$ exists, $A$ a nonterminal, $\alpha$, $\beta$ sequences of terminals and nonterminals, the item $[j, A \rightarrow \alpha \bullet \beta]$ constitutes the fact that $\alpha$ yields the sequence $\sigma_{j+1} \ldots \sigma_i$ and that there exist sequences $\tau = \sigma_1 \ldots \sigma_j$ and $\phi$ such that there is a deriva-tion from the start symbol to the sequence $\tau A\phi$. The output of the algorithm is *yes* or *no* – depending on whether $\sigma$ is in the language of the grammar or not – and the parse lists containing all parses. The input $\sigma$ is in the language if an item covering the whole string is in the last parse list, i.e. an item $[0, SS \rightarrow \omega\bullet]$ is in $L_n$. As is known, the algorithm accepts the input if and only if there exists a parse for it, i.e. the algorithm is correct.

As could be seen in section 2.2, Shieber's modifications stay fairly close to Earley's original algorithm. Naturally the change in the grammar formalism, namely the split of the information on linear precedence and immediate dominance, affects both the completer and the predictor. They only accept and predict LP acceptable permutations, i.e. every sequence found to the left of the dot in an item is LP acceptable, whereas the part of the right hand side of the rule which has still to be found is treated as a multiset, i.e. every element can be taken from the multiset to the right of the dot if the resulting sequence is still LP acceptable.

Seiffert's extension goes a bit further. Firstly he incorporates all the necessary changes to cope with unification based grammars. He is not longer storing the information on (partially) aquired parse results in parse lists, but rather in a chart, thereby adding the information what part of the input string is looked at to the edges. And finally, the problems he encounters require the extension of the algorithm with a second part (see section 2.2).

There are obvious connections between an Earley style recognizer and chart parsers. In fact they are almost identical. For a more detailed discussion of this see for example Kay (1980), Thompson (1983), or Kilbury (1985).

The algorithm presented here is still based on the same ideas, but overcomes the discussed problem of nonlocal feature passing. As in Seiffert's approach, a chart is used instead of parse lists and a distinct symbol is used to initialize the chart. This makes the formulation of the algorithm slightly easier. An initialization with the start symbol, instead of the one with a new distinct symbol as is done here, would be possible, but in this case it would be necessary to do the initialization on all rules which have a feature graph on the left hand side which is subsumed by the start feature graph. This is more complicated and the algorithm can do the work with the existing mechanisms if initialized as indicated here.

Altough the grammar formalism is changed compared to Seiffert's definitions, these changes are not reflected in the algorithm itself, but rather in the operations used by the algorithm.[19] The scanner is not much different – with the exception that Seiffert's scanner only introduces passive edges since his grammar formalism allows the terminals only to occur in the special *lex* relation and not in the grammar rules themselves. Since this algorithm is supposed to be as general as possible, the change to allow terminals and nonterminals to mix on the right hand sides of rules seems appropriate, but nothing hinges on it. In particular, if the part to the left of the dot, i.e. $\alpha$ in the algorithm, and the remainder of the multiset without the terminal, i.e. $\eta$, are empty, there is no difference any more. Following Barton et al. (1987), one could produce fewer edges by having a multiset to the left of the dot instead of having an ordered list. But as discussed above, this would reduce the parser to a recognizer, since it would no longer be possible to extract the parse from the chart. If the parse is constructed by the feature graphs themselves, it might be worthwhile to consider this proposal as an optimization for other algorithms. But the proposed algorithm can not do this for the reason that it needs the recognized sequences to construct the pairs on the LP store. And they have to be kept to ensure that the appropriate changes appear on the LP store via structure sharing.

[19] Naturally only the first part of Seiffert's algorithm is under consideration here, since the second part is no longer necessary.

Compared to Shieber's algorithm, the check for LP acceptability has been omitted from the predictor. As discussed in section 6.2.2, firstly for the reason Seiffert mentions, namely that it is costly to test, and secondly because it would require an even more complicated way of treating possible violations of LP rules than in the completer. Clearly this would bring fewer edges, but would be inefficient to perform.

To avoid the problem of nontermination discussed in Shieber (1985) the use of a restrictor has been included in the predictor. Whether this is really necessary depends on the grammar. In his paper Shieber gives a somewhat artificial example of a counting grammar to illustrate the problem, but there are cases of linguistic examples where a restrictor seems necessary. In parsing, the problems occur with very simple HPSG style grammars which handle subcategorization requirements of VPs and in Earley generation with gap threading (see Gerdemann (1991)). A restrictor may be thought of as a set of paths that partition the infinite domain of the nonterminals into a finite number of equivalence classes by limiting the feature graphs to those paths given in the restrictor. If the grammar does not induce this problem of nontermination of the predictor, the restrictor is not necessary and could be eliminated. This will not be the case for the implementation because, as explained in section 7.4, the restrictor is used for optimization techniques as well. The restrictor has to be specified by the grammar writer and influences the behaviour of the algorithm; even correctness may depend on an appropriately choosen restrictor (Seiffert 1987). In the case where the LP relevant information was discarded by restriction in the prediction and not available at the completion of a particular local tree, the first step of his algorithm would accept some invalid structures concerning linear precedence since the information would naturally be added nonlocally at some later stage of processing. The algorithm of direct parsing would be incorrect. If one proceeds along the lines explained in Sikkel (1993), a default restrictor may be derived automatically from the grammar by including all the paths which appear on the right hand side of rules. This may not be the best possible restrictor in terms of efficiency, but it ensures the appropriate termination. For an approach towards ID/LP parsing, it becomes necessary to include all the paths which appear in the LP rules as well so that the necessary information to determine LP acceptability is not discarded. The usage of a restricted feature graphs is indicated by writing the feature graph *modulo* the restrictor.

The completer contains the solution to the problem of nonlocal feature passing. As in Seiffert's approach, two edges are taken from the chart in such a way that the appropriate categories unify. But instead of simply adding the item to the chart if the resulting structure is LP acceptable, the test for possible LP violation is performed. If this test is not successful, we proceed like Seiffert's algorithm usually works, i.e. it results in failure if the structure was not LP acceptable and in acceptance if the structure is LP acceptable only. If the test does succeed, i.e. the structure is possibly LP violated, a pair as described previously is added to the LP store by union. The completer unions the LP stores of the two edges involved if a new edge is constructed at all. In between the unification and this test for LP acceptability, the test concerning the pending LP information – which has been explained above – is performed on the LP store. If there exists a pair on the store, such that the elements of the pair are different, the structure shared one is tested for LP acceptability. No edge is added, if the sequence is not LP acceptable. If the test yields LP acceptability, the algorithm proceeds as discribed directly above. So, as soon as the information that could not be determined before is available, the algorithm discardes the hypothethis. But the algorithm does not remove the edges used to produce this invalid hypothethis from the chart, i.e. it does not backtrack in any way since there may be another way to use them that does not lead to failure.

Naturally the algorithm adds only edges to the chart if they are not already there. This is done – as in Seiffert's algorithm – by checking if there is an edge in the chart that subsumes the newly created one. An edge subsumes another just in case the feature graphs involved subsume the ones in the other edge and everything else is identical.

Note that the control structure of the algorithm is left vague. Specific implementations of the algorithm will have to care for the realization of the closure operation on scanner, completer and predictor. This takes the given definition closer to an algorithm schema as defined in Kay (1980) as opposed to an actual algorithm. Since Sikkel (1993) takes this distinction between a parser and a parsing schema even further, we are not going to use this terminology to avoid confusion.

After all this preliminary discussions, the code for the algorithm looks as follows:

**28 Algorithm**

Input: a TU ID/LP grammar $G = \langle T, NT, TH, AS, IDR, LPR, SFS \rangle$
a sentence $w = w_1 \ldots w_n$
a restrictor $R$

Output: a chart containing all parses

1. Initialisation:
   Add $[\ 0\ ,\ 0\ ,\ *** \rightarrow []\ \bullet\ \{SFS\}\ ,\ \{\}\ ]$ to the Chart.

2. Compute the closure $\forall j\ 0 \leq j \leq n$ under (3), (4) and (5)

3. Scanner:
   For all edges $[\ i\ ,\ j-1\ ,\ A \rightarrow \alpha\ \bullet\ \{w_j\} \cup \eta\ ,\ S\ ]\ \in$ Chart
   such that $w_j \in T$
   add an edge $[\ i\ ,\ j\ ,\ A \rightarrow \alpha w_j\ \bullet\ \eta\ ,\ S\ ]$ to the Chart
   if $j > 0$ and the edge is not subsumed by an edge already in the Chart.

4. Completer:
   For all edges $[\ i\ ,\ j\ ,\ A \rightarrow \alpha\ \bullet\ \{\}\ ,\ S1\ ]\ \in$ Chart and
   for all edges $[\ k\ ,\ i\ ,\ B \rightarrow \beta\ \bullet\ \{C\} \cup \eta\ ,\ S2\ ]\ \in$ Chart such that
   $C' = C \sqcup A$ and
   no $(\sigma, \tau) \in S1 \cup S2$ such that

   $dif(\sigma, \tau)$ and $\sigma$ is not LP acceptable

   add an edge $[\ k\ ,\ j\ ,\ B' \rightarrow \boxed{\text{x}}\beta'C'\ \bullet\ \eta'\ ,\ \{(\boxed{\text{x}}\beta'C', \beta'C')\} \cup S1 \cup S2\ ]$ to the Chart
   if $\beta'C'$ is possibly LP violated and
   the edge is not subsumed by an edge already in the Chart
   else add an edge $[\ k\ ,\ j\ ,\ B' \rightarrow \beta'C'\ \bullet\ \eta'\ ,\ S1 \cup S2\ ]$ to the Chart
   if $\beta'C'$ is LP acceptable and
   the edge is not subsumed by an edge already in the Chart.

5. Predictor:
   For all edges $[\ i\ ,\ j\ ,\ A \rightarrow \alpha\ \bullet\ \{C\} \cup \zeta\ ,\ S\ ]\ \in$ Chart
   such that $B \rightarrow \eta\ \in IDR$ and
   $B' = B \sqcup (C\ mod\ R)$
   add an edge $[\ j\ ,\ j\ ,\ B' \rightarrow []\ \bullet\ \eta'\ ,\ \{\}\ ]$ to the Chart
   if the edge is not subsumed by an edge already in the Chart.

6. $w$ is recognized iff $[\ 0\ ,\ n\ ,\ *** \rightarrow M\ \bullet\ \{\}\ ,\ S\ ]\ \in$ Chart and $SFS \sqsubseteq M$.

As can be seen in the completor, the concept of structure sharing seems to have been augmented. The reentrancies introduced by the algorithm do not indicate reentrant paths between feature graphs, but rather forces the sharing of whole sequences of feature graphs. This is artificial which is due to the fact that we abstracted away from rules being a single feature graph. If one recalls from a previous discussion in section 5.1 that one has to take the view that a rule is a single feature graph or a mutiply rooted feature graph, the structures to be shared do actually exist. This structure sharing is used to construct the stored elements used to emulate the techniques normally employed with the design of delay primitives for programming languages. In those implementations, a whole environment concerning for example all variable bindings has to be stored to be able to resume computation at exactly the point where the delay was forced.[20] If the algorithm would perform exactly the same operations, it would be more efficient, since no work would have to be redone. At the moment, the algorithm has to redo the computation of LP acceptability every time a change in the LP store is discovered since the information on the sequences is all that is available. Naturally one could store the relevant LP rules together with the pending LP information so that not the whole test for LP acceptability would have to be redone.Maybe even the node in the feature graphs responsible for the posssible applications could be stored. But this would need a lot of space since if one is faced with extremely underspecified categories – which might very well appear due to problems with the implementation of HPSG like theories (Meurers 1994) – a large number of LP rules might weakly apply. And it seems more important to clarify what is going on instead of confusing the issue by incorporating all possible optimizations. So this approach does not follow the example of delay primitives to the last, but compromises on efficiency – which is anyway not the major issue in the presentation of an algorithm – and clarity.

The definition of the language of the TU ID/LP grammars allowed sentences which have valid derivations and parse trees, but whose parse trees may have more specific instances which are not LP acceptable. This is reflected in the algorithm as well. According to the presentation of the algorithm, a parse may be completed and still contain pending LP information. Since feature graphs are entirely syntactic constructs, they do not represent descriptions of objects. Therefore there is nothing that forces the algorithm to rule out those cases. There are several possibilities what to do with this information. The simplest one is just to ignore it. But clearly this would mean to discard that some extensions of this parse are not valid whereas others are. Another choice which seems adequate for a grammar developing environment is to report this result to the grammar writer and to enable some kind of choice whether the structure should be accepted or not. Naturally the grammar writer would have to change the grammar accordingly to prevent those occurences if they are not wanted. A further possibility would be to construct all possible extensions by unifying the pending information in. This would mean that all permutations of the valid unifications would have to be constructed. This would result in the return of several parses instead of just one. Maybe the combination of the last two possibilities is the way to go, because it leaves the grammar writer the greatest opportunities to decide what to do.

## 6.4 Some Examples

To explain more closely how the algorithm works, some example traces are given. Firstly a trace for the simple grammar in figure 13 is given and secondly the grammar presented by Seiffert to illustrate the problem of nonlocal feature passing is recast in TU ID/LP format and a trace for an input that demonstrates the rejection of a not acceptable structure is given. Note that none of those examples are supposed to be more than toy grammars to illustrate a point.

[20] Discussions of the design, problems and solutions of coroutining can for example be found in Carlsson (1986) or Carlsson (1987).

In figure 15 the chart for the input grammer in figure 13 and the sentence *she walks* is given.[21] The restrictor has been omitted since it plays no role here.

| | | | |
|---|---|---|---|
| | 1. | $[\,0\,,\,0\,,\,{*}{*}{*} \to [\,]\bullet \left\{\begin{bmatrix} s \\ \text{VFORM}\ \mathit{vform}\end{bmatrix}\right\},\{\}\,]$ | by Ini. |
| $j=0$ | 2. | $[\,0\,,\,0\,,\begin{bmatrix} s \\ \text{VFORM}\ \boxed{1}\ \mathit{vform}\end{bmatrix} \to [\,]\bullet \left\{\begin{bmatrix} \mathit{np} \\ \text{AGR}\ \boxed{2}\ \mathit{pers} \\ \text{CASE}\ \mathit{nom}\end{bmatrix},\begin{bmatrix} \mathit{vp} \\ \text{AGR}\ \boxed{2}\ \mathit{pers} \\ \text{VFORM}\ \boxed{1}\ \mathit{vform}\end{bmatrix}\right\},\{\}\,]$ | by Pred. & 1. |
| | 3. | $[\,0\,,\,0\,,\begin{bmatrix} \mathit{np} \\ \text{AGR}\ \mathit{3pers} \\ \text{CASE}\ \mathit{nom}\end{bmatrix} \to [\,]\bullet \{\mathit{she}\},\{\}\,]$ | by Pred. & 2. |
| | 4. | $[\,0\,,\,0\,,\begin{bmatrix} \mathit{np} \\ \text{AGR}\ \mathit{1pers} \\ \text{CASE}\ \mathit{nom}\end{bmatrix} \to [\,]\bullet \{\mathit{I}\},\{\}\,]$ | by Pred. & 2. |
| | 5. | $[\,0\,,\,0\,,\begin{bmatrix} \mathit{vp} \\ \text{AGR}\ \mathit{3pers} \\ \text{VFORM}\ \mathit{fin}\end{bmatrix} \to [\,]\bullet \{\mathit{walks}\},\{\}\,]$ | by Pred. & 2. |
| $j=1$ | 6. | $[\,0\,,\,1\,,\begin{bmatrix} \mathit{np} \\ \text{AGR}\ \mathit{3pers} \\ \text{CASE}\ \mathit{nom}\end{bmatrix} \to [\mathit{she}]\bullet \{\},\{\}\,]$ | by Scan. & 3. |
| | 7. | $[\,0\,,\,1\,,\begin{bmatrix} s \\ \text{VFORM}\ \boxed{1}\ \mathit{vform}\end{bmatrix} \to \left[\begin{bmatrix} \mathit{np} \\ \text{AGR}\ \boxed{2}\ \mathit{3pers} \\ \text{CASE}\ \mathit{nom}\end{bmatrix}\right]\bullet \left\{\begin{bmatrix} \mathit{vp} \\ \text{AGR}\ \boxed{2}\ \mathit{3pers} \\ \text{VFORM}\ \boxed{1}\ \mathit{vform}\end{bmatrix}\right\},\{\}\,]$ | by Comp. & 2. & 6. |
| | 8. | $[\,1\,,\,1\,,\begin{bmatrix} \mathit{vp} \\ \text{AGR}\ \mathit{3pers} \\ \text{VFORM}\ \mathit{fin}\end{bmatrix} \to [\,]\bullet \{\mathit{walks}\},\{\}\,]$ | by Pred. & 7. |
| $j=2$ | 9. | $[\,1\,,\,2\,,\begin{bmatrix} \mathit{vp} \\ \text{AGR}\ \mathit{3pers} \\ \text{VFORM}\ \mathit{fin}\end{bmatrix} \to [\mathit{walks}]\bullet \{\},\{\}\,]$ | by Scan. & 8. |
| | 10. | $[\,0\,,\,2\,,\begin{bmatrix} s \\ \text{VFORM}\ \boxed{1}\ \mathit{fin}\end{bmatrix} \to \left[\begin{bmatrix} \mathit{np} \\ \text{AGR}\ \boxed{2}\ \mathit{3pers} \\ \text{CASE}\ \mathit{nom}\end{bmatrix}\begin{bmatrix} \mathit{vp} \\ \text{AGR}\ \boxed{2}\ \mathit{3pers} \\ \text{VFORM}\ \boxed{1}\ \mathit{fin}\end{bmatrix}\right]\bullet \{\},\{\}\,]$ | by Comp. & 7. & 9. |
| | 11. | $[\,0\,,\,2\,,\,{*}{*}{*} \to \left[\begin{bmatrix} s \\ \text{VFORM}\ \mathit{fin}\end{bmatrix}\right]\bullet \{\},\{\}\,]$ | by Comp. & 1. & 10. |

Figure 15: The chart for input *she walks* and the grammar given in figure 13

A rough way of describing how the algorithm works in this case is as follows

[21] The chart is represented in such a way that in the leftmost column the current $j$ the algorithm is operating on is given, then there is a numbering for the edges, followed by the edge itself and then by some information how the edge was created. Again, structure sharing is indicated with boxed integers and lowercase letters. The actual value for the reentrancy is written behind all occurrences of the tag though it exists only once. Tags that indicate structures which are no longer reentrant with anything are removed.

(see figure 15). The algorithm starts – after the initialization – with the prediction of the NP and the VP since both are elements of the multiset to the right of the dot, and continues with the scan of the NP. The prediction of all categories from the multiset is clearly inefficient. To improve on this, one could imploy the $First^+$ relation defined by Kilbury (1984a) and Dörre and Momma (1985). This was not done for reasons of the simplicity of the presentation. Alternativly, one might be tempted to try to generalize the NP and the VP and do the generalization only on this generalized feature graph. But this would lead to less top down guidance, since information contained in the categories would be discarded. And more importantly, at least as many rules would be predicted since the generalized feature graph would surly unify with the left hand side of as many rules as the single original categories involved. As the next step, the completer takes the sentence rule and moves the NP to the list to the left of the dot, i.e. to the already recognized part. Now only the VP is predicted since this is the only feature graph left in the multiset. After scanning of the VP, the completer finishes the computation by moving the VP to the list to the left of the dot thereby checking LP acceptability. Now there are no new edges that could be added to the chart. And the chart contains a passive edge that covers the whole input and has a feature graph as mother which is subsumed by the start symbol. The input is recognized.

The grammar given in figure 16 is designed to emulate the behaviour of Seiffert's example grammar (see figure 3). The CAT feature has been omitted and the values of these features are now used as types with certain features. But since there would not be any difference between some of them any more, a new feature, namely SP and some other types, i.e. *hip*, *hop* and *sp* have been introduced to create the difference.[22] To keep the LP rules simple, the varieties $d$ and $f$ are both of type $x$ and likewise $e$ and $g$ of type $y$. All these changes can be seen in the type hierarchy. Naturally the grammar rules had to be changed accordingly. Seiffert's *lex* relation is now part of the unification ID rules and we have separate sets for the terminals and nonterminals.

In figure 17 the chart for the input grammar from figure 16 and input word *ihjk* is given. Again the restrictor has been omitted. This input triggers the mechanism that treats the problem of nonlocal feature passing. Ultimately the input word is not recognized because there is a violation of an LP rule which can only be discovered after some computation of other local trees has been completed.

In comparison to the other example, there will be no explanation in similar detail as before. It is assumed that the basic principles involved are known. Emphasis will be on the working of the mechanism that copes with nonlocal feature passing.

The first twelve edges are created by applications of the standard definitions of the predictor and scanner. The edge 13. deserves special attention. Here the completer encounters a case where an LP rule waekly applies. A new pair

[22] Since this is artificial, SP and *sp* stand for spurious.

Let $G = \langle T, NT, TH, AS, IDR, LPR, SFS \rangle$ be TU ID/LP G such that

$T \; = \{h, i, j, k\}$

$TH + AS =$

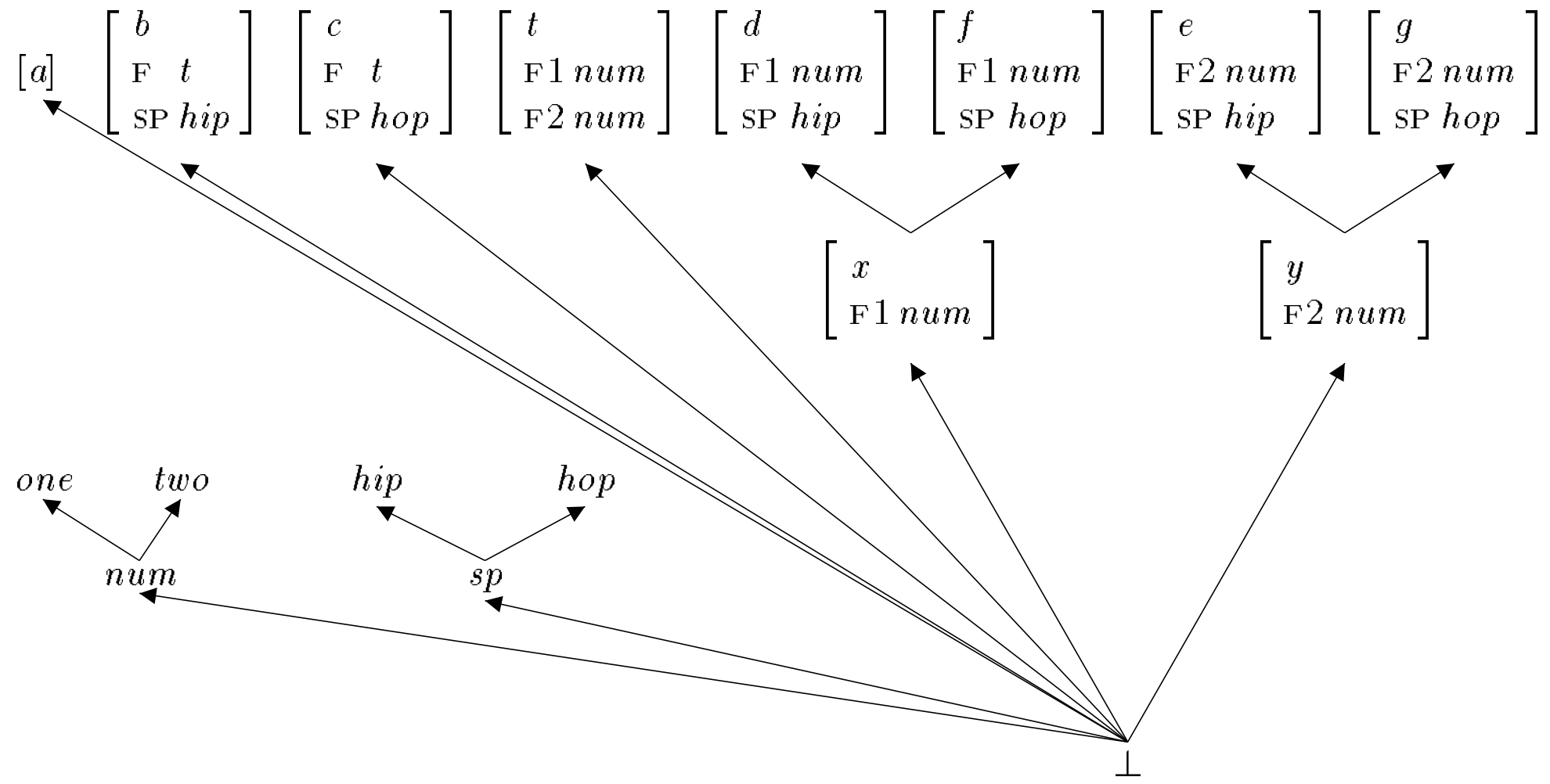

$$IDR = \left\{ \begin{array}{lll}
[a] & \rightarrow & \begin{bmatrix} b \\ \text{F} \; \boxed{1} \; t \end{bmatrix}, \begin{bmatrix} c \\ \text{F} \; \boxed{1} \; t \end{bmatrix} \\
\begin{bmatrix} b \\ \text{F} \begin{bmatrix} t \\ \text{F1} \; \boxed{1} \; num \\ \text{F2} \; \boxed{2} \; num \end{bmatrix} \end{bmatrix} & \rightarrow & \begin{bmatrix} d \\ \text{F1} \; \boxed{1} \; num \end{bmatrix}, \begin{bmatrix} e \\ \text{F2} \; \boxed{2} \; num \end{bmatrix} \\
\begin{bmatrix} c \\ \text{F} \begin{bmatrix} t \\ \text{F1} \; \boxed{1} \; num \\ \text{F2} \; \boxed{2} \; num \end{bmatrix} \end{bmatrix} & \rightarrow & \begin{bmatrix} f \\ \text{F1} \; \boxed{1} \; num \end{bmatrix}, \begin{bmatrix} g \\ \text{F2} \; \boxed{2} \; num \end{bmatrix} \\
[d] & \rightarrow & h \\
[e] & \rightarrow & i \\
\begin{bmatrix} f \\ \text{F1} \; one \end{bmatrix} & \rightarrow & j \\
\begin{bmatrix} g \\ \text{F2} \; two \end{bmatrix} & \rightarrow & k
\end{array} \right\}$$

$$LPR = \left\{ [b] \prec [c], \begin{bmatrix} x \\ \text{F1} \; one \end{bmatrix} \prec \begin{bmatrix} y \\ \text{F2} \; two \end{bmatrix} \right\}$$

$$SFS = [a]$$

$$L(G) = \{hijk\}$$

Figure 16: Seiffert's grammar in TU ID/LP G format

1. $[\,0\,,0\,,*\!*\!*\rightarrow[\,]\bullet\{[a]\}\,,\{\}\,]$ by Ini.

$j=0$ 2. $$[\,0\,,0\,,[a]\rightarrow[\,]\bullet\left\{\begin{bmatrix} b \\ \text{F}\,\boxed{1}\;t \end{bmatrix},\begin{bmatrix} c \\ \text{F}\,\boxed{1}\;t \end{bmatrix}\right\},\{\}\,]$$ by Pred. & 1.

3. $$[\,0\,,0\,,\begin{bmatrix} b \\ \text{F}\begin{bmatrix} t \\ \text{F1}\,\boxed{2}\;num \\ \text{F2}\,\boxed{3}\;num \end{bmatrix}\end{bmatrix}\rightarrow[\,]\bullet\left\{\begin{bmatrix} d \\ \text{F1}\,\boxed{2}\;num \end{bmatrix},\begin{bmatrix} e \\ \text{F2}\,\boxed{3}\;num \end{bmatrix}\right\},\{\}\,]$$ by Pred. & 2.

4. $$[\,0\,,0\,,\begin{bmatrix} c \\ \text{F}\begin{bmatrix} t \\ \text{F1}\,\boxed{4}\;num \\ \text{F2}\,\boxed{5}\;num \end{bmatrix}\end{bmatrix}\rightarrow[\,]\bullet\left\{\begin{bmatrix} f \\ \text{F1}\,\boxed{4}\;num \end{bmatrix},\begin{bmatrix} g \\ \text{F2}\,\boxed{5}\;num \end{bmatrix}\right\},\{\}\,]$$ by Pred. & 2.

5. $$[\,0\,,0\,,\begin{bmatrix} d \\ \text{F1}\;num \end{bmatrix}\rightarrow[\,]\bullet\{h\}\,,\{\}\,]$$ by Pred. & 3.

6. $$[\,0\,,0\,,\begin{bmatrix} e \\ \text{F2}\;num \end{bmatrix}\rightarrow[\,]\bullet\{i\}\,,\{\}\,]$$ by Pred. & 3.

7. $$[\,0\,,0\,,\begin{bmatrix} f \\ \text{F1}\;one \end{bmatrix}\rightarrow[\,]\bullet\{j\}\,,\{\}\,]$$ by Pred. & 4.

8. $$[\,0\,,0\,,\begin{bmatrix} g \\ \text{F2}\;two \end{bmatrix}\rightarrow[\,]\bullet\{k\}\,,\{\}\,]$$ by Pred. & 4.

$j=1$ 9. $$[\,0\,,1\,,\begin{bmatrix} e \\ \text{F2}\;num \end{bmatrix}\rightarrow[i]\bullet\{\}\,,\{\}\,]$$ by Scan. & 6.

10. $$[\,0\,,1\,,\begin{bmatrix} b \\ \text{F}\begin{bmatrix} t \\ \text{F1}\,\boxed{2}\;num \\ \text{F2}\,\boxed{3}\;num \end{bmatrix}\end{bmatrix}\rightarrow\left[\begin{bmatrix} e \\ \text{F2}\,\boxed{3}\;num \end{bmatrix}\right]\bullet\left\{\begin{bmatrix} d \\ \text{F1}\,\boxed{2}\;num \end{bmatrix}\right\},\{\}\,]$$ by Comp. & 3. & 9.

11. $$[\,1\,,1\,,\begin{bmatrix} d \\ \text{F1}\;num \end{bmatrix}\rightarrow[\,]\bullet\{h\}\,,\{\}\,]$$ by Pred. & 10.

$j=2$ 12. $$[\,1\,,2\,,\begin{bmatrix} d \\ \text{F1}\;num \end{bmatrix}\rightarrow[h]\bullet\{\}\,,\{\}\,]$$ by Scan. & 11.

13. $$[\,0\,,2\,,\begin{bmatrix} b \\ \text{F}\begin{bmatrix} t \\ \text{F1}\,\boxed{2}\;num \\ \text{F2}\,\boxed{3}\;num \end{bmatrix}\end{bmatrix}\rightarrow\left[\boxed{\text{x}}\begin{bmatrix} e \\ \text{F2}\,\boxed{3}\;num \end{bmatrix}\begin{bmatrix} d \\ \text{F1}\,\boxed{2}\;num \end{bmatrix}\right]\bullet\{\}\,,$$
$$\left\{\left(\boxed{\text{x}}\begin{bmatrix} e \\ \text{F2}\,\boxed{3}\;num \end{bmatrix}\begin{bmatrix} d \\ \text{F1}\,\boxed{2}\;num \end{bmatrix},\begin{bmatrix} e \\ \text{F2}\;num \end{bmatrix}\begin{bmatrix} d \\ \text{F1}\;num \end{bmatrix}\right)\right\}\,]$$ by Comp. & 10. & 12.

14. $$[\,0\,,2\,,[a]\rightarrow\left[\begin{bmatrix} b \\ \text{F}\,\boxed{1}\begin{bmatrix} t \\ \text{F1}\,\boxed{2}\;num \\ \text{F2}\,\boxed{3}\;num \end{bmatrix}\end{bmatrix}\right]\bullet\left\{\begin{bmatrix} c \\ \text{F}\,\boxed{1}\begin{bmatrix} t \\ \text{F1}\,\boxed{2}\;num \\ \text{F2}\,\boxed{3}\;num \end{bmatrix}\end{bmatrix}\right\},$$
$$\left\{\left(\begin{bmatrix} e \\ \text{F2}\,\boxed{3}\;num \end{bmatrix}\begin{bmatrix} d \\ \text{F1}\,\boxed{2}\;num \end{bmatrix},\begin{bmatrix} e \\ \text{F2}\;num \end{bmatrix}\begin{bmatrix} d \\ \text{F1}\;num \end{bmatrix}\right)\right\}\,]$$ by Comp. & 3. & 13.

Figure 17: The chart for input word *ihjk* and Seiffert's grammar (see figure 16), part I

15. $$[\,2\,,2\,,\left[\begin{array}{l} c \\ \mathrm{F}\left[\begin{array}{l} t \\ \mathrm{F1}\,\boxed{6}\ \mathit{num} \\ \mathrm{F2}\,\boxed{7}\ \mathit{num}\end{array}\right]\end{array}\right] \rightarrow [\,]\bullet\left\{\left[\begin{array}{l} f \\ \mathrm{F1}\,\boxed{6}\ \mathit{num}\end{array}\right],\left[\begin{array}{l} g \\ \mathrm{F2}\,\boxed{7}\ \mathit{num}\end{array}\right]\right\},\{\}\,]$$ by Pred. & 14.

16. $$[\,2\,,2\,,\left[\begin{array}{l} f \\ \mathrm{F1}\,\mathit{one}\end{array}\right] \rightarrow [\,]\bullet\{j\},\{\}\,]$$ by Pred. & 15.

17. $$[\,2\,,2\,,\left[\begin{array}{l} g \\ \mathrm{F2}\,\mathit{two}\end{array}\right] \rightarrow [\,]\bullet\{k\},\{\}\,]$$ by Pred. & 15.

$j=3$ 18. $$[\,2\,,3\,,\left[\begin{array}{l} f \\ \mathrm{F1}\,\mathit{one}\end{array}\right] \rightarrow [j]\bullet\{\},\{\}\,]$$ by Scan. & 16.

19. $$[\,2\,,3\,,\left[\begin{array}{l} c \\ \mathrm{F}\left[\begin{array}{l} t \\ \mathrm{F1}\,\boxed{6}\ \mathit{one} \\ \mathrm{F2}\,\boxed{7}\ \mathit{num}\end{array}\right]\end{array}\right] \rightarrow \left[\left[\begin{array}{l} f \\ \mathrm{F1}\,\boxed{6}\ \mathit{one}\end{array}\right]\right]\bullet\left\{\left[\begin{array}{l} g \\ \mathrm{F2}\,\boxed{7}\ \mathit{num}\end{array}\right]\right\},\{\}\,]$$ by Comp. & 15. & 18.

20. $$[\,3\,,3\,,\left[\begin{array}{l} g \\ \mathrm{F2}\,\mathit{two}\end{array}\right] \rightarrow [\,]\bullet\{k\},\{\}\,]$$ by Pred. & 19.

$j=4$ 21. $$[\,3\,,4\,,\left[\begin{array}{l} g \\ \mathrm{F2}\,\mathit{two}\end{array}\right] \rightarrow [k]\bullet\{\},\{\}\,]$$ by Scan. & 20.

22. $$[\,2\,,4\,,\left[\begin{array}{l} c \\ \mathrm{F}\left[\begin{array}{l} t \\ \mathrm{F1}\,\boxed{6}\ \mathit{one} \\ \mathrm{F2}\,\boxed{7}\ \mathit{two}\end{array}\right]\end{array}\right] \rightarrow \left[\left[\begin{array}{l} f \\ \mathrm{F1}\,\boxed{6}\ \mathit{one}\end{array}\right]\left[\begin{array}{l} g \\ \mathrm{F2}\,\boxed{7}\ \mathit{two}\end{array}\right]\right]\bullet\{\},\{\}\,]$$ by Comp. & 19. & 21.

*23. $$[\,0\,,4\,,[a]\rightarrow\left[\left[\begin{array}{l} b \\ \mathrm{F}\,\boxed{1}\left[\begin{array}{l} t \\ \mathrm{F1}\,\boxed{2}\ \mathit{one} \\ \mathrm{F2}\,\boxed{3}\ \mathit{two}\end{array}\right]\end{array}\right]\left[\begin{array}{l} c \\ \mathrm{F}\,\boxed{1}\left[\begin{array}{l} t \\ \mathrm{F1}\,\boxed{2}\ \mathit{one} \\ \mathrm{F2}\,\boxed{3}\ \mathit{two}\end{array}\right]\end{array}\right]\right]\bullet\{\},$$ by Comp. & 14. & 22.

$$\left\{\left(\left[\begin{array}{l} e \\ \mathrm{F2}\,\boxed{3}\ \mathit{two}\end{array}\right]\left[\begin{array}{l} d \\ \mathrm{F1}\,\boxed{2}\ \mathit{one}\end{array}\right],\left[\begin{array}{l} e \\ \mathrm{F2}\,\mathit{num}\end{array}\right]\left[\begin{array}{l} d \\ \mathrm{F1}\,\mathit{num}\end{array}\right]\right)\right\}\,]$$

Figure 17: The chart for input word *ihjk* and Seiffert's grammar (see figure 16), part II

is added to the LP store containing the relevant information. The reentrancies contained in the sequence, i.e. feature graphs, in the second place of the new pair do not longer convey any meaning and are thus eliminated. The structure sharings in the first element of the pair are linked to the edge 13. and therefore do convey information. They are altered as soon as these feature graphs are changed during the computation process by unification in the completer. The store is passed on to all further constructions which relay in some way on this particular edge. The edges between 14. and 22. are again created following the standard procedures of Earley's algorithm. The edge 23. is included in this chart only for the reason to illustrate the point. It would *not* be constructed by the algorithm. The reason for this is as follows. The completer would try to build the edge from the edges 14. and 22. by moving the recognized feature graph with the type $c$ to the list to the left of the dot. This feature graph is the result of the unification between two feature graphs, namely those two from the edges 14. and 22. which have the type *c*. This unification forces identity between the reentrancies $\boxed{2}$ and $\boxed{6}$ and $\boxed{3}$ and $\boxed{7}$ respectivly. This forces the values *one* and *two* on the elements on the LP store. Now the function *dif* notes the difference between the two components of the pair on the LP store. The first of those two is tested for LP acceptability and the violation is detected. The hypothesis is discarded as desired. No new edges can be added, i.e. the closure is computed. Since no edge that spans the whole string and conforms to the start feature graph is contained in the chart, the input is not recognized.

### 6.5 Complexity

This section is mainly influenced by Barton's work (Barton 1985) on the complexity of ID/LP parsing and of the recognition problem for GPSG (Barton et al. 1987), but it does not present a formal account of the complexity. It just points out some properties of the algorithm presented previously. Some knowledge on complexity is assumed.

In chapter 7 of their book, Barton, Berwick and Ristad proved that ID/LP recognition is inherently difficult, i.e. NP–complete. Consequently it follows that there can not be a general recognition algorithm with a polynomial runtime bound, unless the complexity class $\mathcal{P}$ equals $\mathcal{NP}$. It gets only worse if the other principles of GPSG are taken into account. Consequently, the universal recognition problem for GPSG is EXP–POLY hard. There is no reason to expect the typed unification based ID/LP grammars to behave better with respect to the ID/LP component, although there are no metarules and instantiation principles any more.

Since the algorithm follows Shieber's algorithm very closely, most of the statements made by Barton, Berwick and Ristad about his algorithm are true of the one presented here as well.

For Earley's algorithm the runtime bound of $O(|G|^2 \cdot n^3)$ is derived by the following reasoning. There are at most $k+1$ possible dotted rules for each context free rule with length $k$ of the right hand side. The number of possible dotted rules is bounded by $|G|$. Since the input length is $n$, no state set can contain more than $O(|G| \cdot n)$ items. The scan operation is assumed to be constant, there can not be more than $O(|G| \cdot n)$ steps. The same bound is evident for the prediction step since there can be at most as many predictions as there are grammar rules. The completion operation is bounded by $O(|G|^2 \cdot n^2)$ since each completion requires at most the quadratic length of the rule. At the end, there can be at most $O(|G|^2 \cdot n^2)$ steps to process each state set. To process all state sets, the parser needs at most $O(|G|^2 \cdot n^3)$ steps.

For Shieber's direct ID/LP parsing algorithm[23], the worst case is the one

[23] Barton, Berwick and Ristad improve on Shieber's algorithm by changing the way the recognized categories are memorized. Shieber keeps them in a list whereas Barton, Berwick and Ristad keep them as a (multi)set which saves items although it is not longer possible to

where there is no order imposed on the rules. The argument runs as follows. For an ID rule of length $k$, there are $2^k$ possible dotted rules as opposed to $k+1$ rules in the context free case. Hence the number of possible dotted rules is not bounded by $|G|$, but rather by $2^{|G|}$. The direct ID/LP parsing approach suffers from this combinatorial explosion because there are exponetially many ways to progress through an unordered right hand side. If no information concerning the linear precedence is available, the parser has to keep track of all of them.

Since the reasononing concerning the complexity is done via the number of the possible states introduced, it becomes even worse for the algorithm presented here. Since we can not keep the recognized categories as a set due to the structure sharing with the LP store and since a state does contain an LP store now, there are further sources for a gain in complexity. The definition of possible LP violation allows the parser to construct more distinct items than before. Since possible LP violation is checked on pairs of feature graphs in relation to those that have already been recognized, the number of items increases with the length of the part of the structure which has been recognized. For example, each dotted rule with two recognized elements has another one with the elements in the LP store, each dotted rule with three recognized categories three additonal ones .... There is no easy regularity to be observed since the number of items is influenced by a set, namely the right hand side, and a list, namely the already recognized categories. A further complication arises since in case there is a possibly LP violated structure, there can not be an LP acceptable identical structure. But in case the structures subsume – apart from the LP store – this leads to different items in the chart.

Summing up, it seems that the algorithm does not improve in any way on Shieber's algorithm, rather the contrary is the case. In their discussion of Shieber's algorithm, Barton, Berwick and Ristad argue strongly that the algorithm behaves better than Earley's algorithm on a grammar with the same coverage and that one should not discard the algorithm because of its worst case behaviour since the average behaviour is sufficiently efficient. The same can be said about the algorithm presented here. For the average application it may be a valid option to use.

extract the parse from those states. They are only dealing with a recognizer, not a parser.

# 7 Issues concerning the Implementation

The implementation of the algorithm was done for the typed feature system TROLL developed at the Universität Tübingen in the Teilprojekt B4 "Constraints on Grammar for Efficient Generation" of the Sonderforschungsbereich 340 "Sprachtheoretische Grundlagen der Computerlinguistik" of the Deutsche Forschungsgemeinschaft by Dale Gerdemann and Thilo Götz (Gerdemann et al. forthcoming) based on a logic developed by Paul King (King (1989) and King (1994a)) designed to handle HPSGs. In the following section, TROLL is presented to provide the reader with some background to the sections on the actual implementation. Furthermore, the changes from the algorithm to the implementation are explained, some important procedures are presented in some detail, a short discussion on optimizations and some test experiences conclude the section. TROLL is implemented in Quintus Prolog (Quintus Corporation 1988) and so is the algorithm.

The implementation follows the example for a top–down parser along the lines given in Gazdar and Mellish (1989). This is done because the parsing algorithm and the procedures involved should be well known so that the changes to treat the nonlocal flow of information can be understood more easily. Therefore this implementation follows closely the presentations in the literature on direct parsing of ID/LP grammars instead of trying to be maximally efficient.

## 7.1 TROLL

The TROLL system is designed to efficiently implement the logic by Paul King using typed feature structures as normal form descriptions. To ensure efficiency, the system was provided with a phrase structure backbone so that specialized parsing algorithms as the one described in this thesis are applicable, instead of relying on for example pure type constraint resolution like systems as TFS (see Emele and Zajac (1990), Zajac (1991) and Aït-Kaci (1984)). The main features of TROLL are *type resolution* and *unfilling*. *Type resolution* is an operation that bumps all the types of a feature structure $F$ to a disjunction of the most specific types that partition the type in question. This is one of the sources of the necessary changes to the definitions of type hierarchy and feature graphs hinted at in section 4. Naturally this assumes a closed world assumption. The result of *type resolution* is a set of feature structures $F'$ which satisfy the appropriateness conditions. This is implemented as having massive disjunctions of feature structures which are compacted down to a single feature structure with distributed disjunctions of types. No new definition is given for feature graphs compared to those in section 4 although these are not the normal form feature structures used in TROLL because this is not possible without presenting most of Götz (1994). Since King (1994b) shows that a feature structure is satisfiable if and only if it has a type resolvant, it becomes possible to use resolution of disjunctions instead of more complicated operations which might not even maintain full satisfiability like Carpenter's *type inferencing* (Carpenter 1992). The other main feature of TROLL is the operation of *unfilling*. As Götz (1994) shows, it is possible to remove redundant arcs while maintaining the semantic equivalence of feature structures. This is important for efficiency since it keeps the feature structures as small as possible.

TROLL provides definite clause attachments so that in case there are constraints the grammar writer can not or wishes not to express in the type hierarchy, these can be encoded nevertheless. A caveat seems necessary. Those definite clauses are in general not efficient and may not be decidable so that the way to write efficient TROLL grammars is to put as much information in the type hierarchy as possible although may prove difficult, if not impossible to put all the in information into the types (Meurers 1994).

In TROLL, unification is a purely syntactic, nondestructive operation that takes normal form feature structures and results in a normal form feature structure. On the semantic side this represents the intersection of the denotations of the descriptions which have to be conjoined. It is important to note that at the same time the unification algorithm maintains satisfiability. TROLL's feature structures are closed under unification with respect to type resolving, but not concerning unfilling. Therefore this has to be explicitly included in the unification algorithm. The subsumption algorithm is as well purely syntactic, with the exception that subsumption is not a part of the logic devised by King (1989). On the semantic side, TROLL's subsumption is sound, but not complete, i.e. if two feature structures subsume each other, the denotations stand in the subset relation, but not vice versa. Since we are dealing with the syntactic side only, this does not affect the presented implementation.

A TROLL grammar consists of a type hierarchy with appropriateness specifications, a lexicon, phrase structure rules and (possibly) some definite clauses.

## 7.2 The Changes from the Algorithm to the Implementation

The implementation differs from the definitions given for the algorithm in several aspects.

The parser utilizes an agenda to influence the control strategy instead of using Prolog backtracking. Therefore it is no longer necessary to rely on the closure operation which had to be done for all the possible vertices, i.e. string positions. This splits the completer into two operations since now we have to deal either with an active or a passive edge from the agenda and, depending on that, slightly different procedures. One takes a passive edge and searches the chart for a fitting active edge, creating the appropriate new edges; the other one takes an active edge and searches the chart for a fitting passive edge. Fitting in this sense means unifiable in the relevant categories as presented in the algorithm, i.e. the left hand side of the passive edge and one element of the right hand side multiset of the active edge observing possible LP violation. The procedures dealing with the LP information stay the same as in the definitions on the algorithm. The predictor is unchanged as well since it deals only with active edges anyway.

Some changes from the algorithm to the actual implementation are induced by Troll. The input grammar has to conform to the given standard. Troll has a lexicon which specifies the relation between lexical entries and the feature graphs describing them. No sets of terminals and nonterminals are specified, but rather a special lexicon. This allows for a change in the initialization step of the algorithm. Since it is known beforehand that no terminals may occur in the rules (see below) and the processing order is determined by an agenda instead of by the vertex numbers, it is now possible to scan the input before the equivalent to the closure operation for the parsing process is started. We loose the top down guidance on the scan operation, but it seems a small price to pay for the initialization of the chart and since we now can determine whether all input words are indeed valid before the parse starts without having to do extra work. In this way, the scanner is included in the initialization step and the created edges are used to initialize the chart. If the grammar contains empty categories, they have to be initialized by adding vertices for them spanning no part of the string, but allowing the consumption of categories in the rules. This is achieved by adding passive edges to the chart for each empty category contained in the lexicon as left hand side of the rule for each position between the words of the input string, and before and after the string. The Troll grammar has to be augmented by a restrictor. As discussed, this restrictor has to specify the relevant paths for the predictions so that nontermination is no longer a problem.[24] This specification has to be done by hand. An initial category and some LP rules have to be added as well. LP rules are of a format that they take two feature graphs, meaning that the first one has to precede the second one.

Probably the biggest change is required by Troll's view that a rule is a single feature graph. Note that although a rule is a single feature graph, an LP rule is not. In effect, this prevents the possibility of having structure sharing between the LP elements. But since this appears not to be desired by linguists, it does not seem to be a limitation. If it would be needed, it could be implemented easily. So far the PS rules necessitated the "system" type *ps_rule* with the appropriate features LHS, RHS and GOALS. LHS and RHS are typed with *list_sign* and GOALS with *list_goal*. The typing of RHS ensures that no terminals may occur here, if they are not subtypes of *sign*. Usually this is not the case because the terminals are treated solely in the lexicon. GOALS contains the definite clauses attached to this particular rule. Since the algorithm employs structure sharing between recognized sequences of feature graphs and an LP store, both have to be contained in the feature graphs denoting the rules. This makes it necessary to augment the type hierarchy with respect to *ps_rule*. The features REC, typed as being *list_sign*, and LP_STORE, being typed as *set_pair*, are introduced. And the types *pair* and *set_pair* become necessary for the LP store. The type *pair* consists, as described in section 6.1, of sequences of feature graphs and *set_pair* is more or less self evident. These changes to the type hierarchy are graphically displayed in figure 18.[25]

[24] The procedures of reading in a restrictor and restricting a feature graph have been implemented by Thilo Götz for usage with an Earley style generator, but could be employed here without changes.

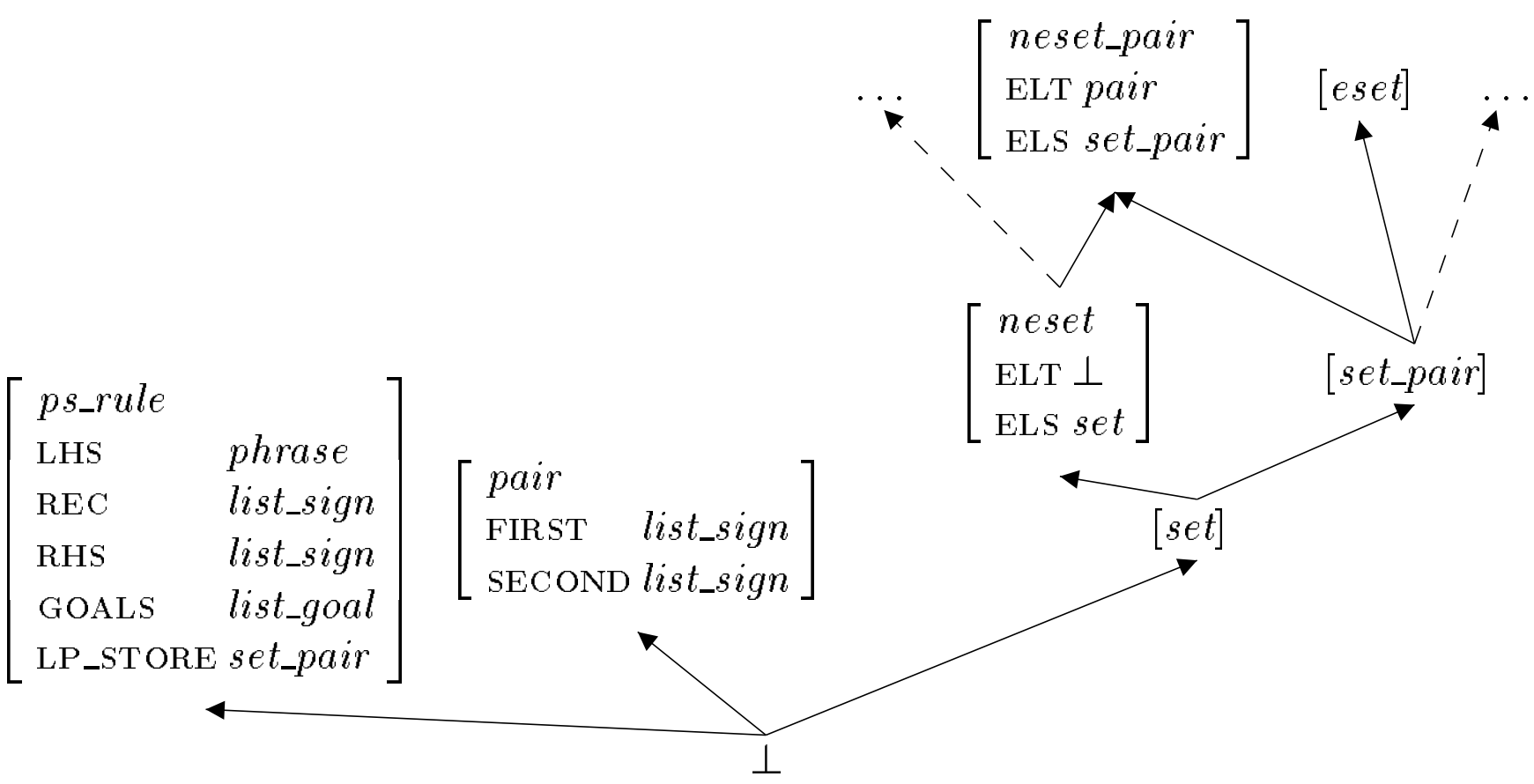

Figure 18: The necessary changes to the TROLL type hierarchy

The features REC and LP_STORE are initialized as being empty (list and set respectively) in the definition of the PS rules since in the standard case no rule definition contains already recognized categories or any pending LP information. If one does not want the grammar writer to specify those by hand, there are no difficulties adding them automatically to each phrase structure rule in a precompilation step. In TROLL, this could be done when the grammar is processed, i.e. when the format the grammar writer specifies the grammar in is transformed into the needed Prolog format.

Another addition consists in TROLL's definite clauses. They have to be executed at some point during parsing since they contain information on the acceptability of the proposed rules. In the implementation, this execution is done whenever a passive edge might be constructed, i.e. the right hand side of the rule is empty. If the definite clauses fail, no edge is constructed. The definite clauses are executed at this point because after all categories have been recognized the maximum amount of information is available so that success of the definite clauses can be determined. If there were a way to interleave the execution of the definite clauses and the recognizing of categories, this might rule out some edges earlier as they can be ruled out now. But this presupposes a very detailed analysis when there is enough information to execute a particular definite clause. For a slightly different setup the methods to do this are described in Minnen (in preparation).

[25]The feature RHS is still typed with *list_sign* although this should now rather be a set. In fact, the algorithm treats this list as a set, but since sets are implemented in TROLL as lists there was no real reason to change. By leaving it the way it is, greater compatibility is achieved.

The definite clauses allow for a specification of delay patterns on them. This could provide a mechanism to interleave execution and recognition. It does not seem worthwhile to interleave the definite clauses and the test for LP acceptability by always trying to execute the frozen goals first and executing them if enough information is available because the test whether a feature graph is specific enough to be unfrozen is very costly. To test for this for all definite clauses for each edge is a great effort which would exclude only few edges.

The definite clauses create a problem concerning the ID/LP component since they may provide the information necessary to determine LP acceptability. But this has to be decided for some edges before the definite clauses are executed. There are two choices. Either the check for LP acceptability has to be repeated or postponed after the execution of the definite clauses; or nothing is done at this point, but rather the mechanism that deals with the nonlocal flow of information has to deal with it in the following manner. Since there would have to be a possibility for an application of an LP rule before the definite clauses are executed, this information would be preserved in the LP store. When the edge is used later on, the difference of the elements of this particular pair on the LP store is noted, the test for LP acceptability performed and the edge ruled out if necessary. The first approach would have the advantage that the edge could be ruled out rather early and would not be tried for any further completion. The drawback is that the test for LP acceptability is costly and has to be performed whenever a passive edge is constructed. The advantage of the second approach is that the test is only performed in case it is needed. The drawback is that the chart contains some edges that are invalid concerning the LP statements. The implementation takes this lazy second approach because any additional LP testing requires a lot of time and complicates the algorithm whereas it is rather nice that the mechanism designed to handle the nonlocal flow of information already provides everything to treat the problem correctly.

No mechanism is provided at the moment to deal with cases where after completion of the entire parse some LP information is still pending. This is a problem more concerned with the user interface which is not the aim of this thesis.

### 7.3 Some specific Procedures

Some particular procedures deserve special attention to show how exactly the ideas presented in the algorithm are implemented.

Firstly it can be said that no extra mechanism was necessary to enable the parser to structure share the sequences of feature graphs on the LP store and the list of the already recognized categories. Since the features REC and FIRST are typed as being *list_sign*, both provide the possibility to access part of the list by creating a reentrancy. The REC list is kept in reverse order which enables us just to share the desired part of the list. If any new categories are recognized and added to the list, these do not appear on the LP store since they are not reentrant. So the structure sharing the algorithm needs is implemented using just the already existing mechanisms TROLL provides.

The check for the validity of the LP stores consists of the extraction and the union of the LP stores of the edges involved and the actual test whether anything has changed that necessitates a new check of LP acceptability on an element on the resulting LP store. Therefore the LP stores are extracted after the unification has taken place and – since they are sets – unioned. This union is implemented by comparing the second elements of the pairs. Whenever they are identical, the elements in question are identical. If they differ, there is a new element. Building on that, the union procedure is easy. The identity check for the elements of the pairs is implemented using mutual subsumption to avoid problems concerned with different internal representations.

The test for possible LP violation is implemented in such a way that the outcome of the test, i.e. possible violation or overall LP acceptance is returned by one traversal, but if the structure is not LP acceptable, it results in failure. This is achieved by interleaving both tests. Recall from definition 26 that possible LP violation presupposes LP acceptance. The LP rules are kept as a list and the proposed sequence is tested by recursing through this list.

The test for LP acceptability is part of a complex procedure which tests LP acceptability and performs the necessary actions on the list of the recognized structures and the LP store. The appropriate feature graph is extracted from the right hand side of the PS rule, then tested for possible LP violation against the already recognized categories and LP acceptability against the remaining elements in the right hand side multiset. If the flag marks no possible violation of LP rules, the feature graph is appended to the already recognized list and the LP store remains unchanged. If there is a possible violation of an LP rule, a new entry for the LP store is created by producing a copy of the sequence in question for the second element and a reentrancy is introduced that links the first element of the pair to the recognized sequence of feature graphs. This new element is unioned into the LP store which is the result of the union of the two LP stores from the edges involved in the construction.

This should suffice to provide the reader with a close knowledge of the processes involved. For more detailed information (the actual code), see Appendix A.

## 7.4 Optimization Techniques

Not very many optimization techniques were employed since the parser should only serve as a prototype to demonstrate the solution of the problem and not as a real system parser. There had to be a compromise between clarity and comparability to those approaches described in the literature on the one hand and efficiency on the other hand. Since it seemed clear beforehand that the efficiency would be low, not too much effort went into optimizations. This low efficiency is on the one hand due to the inherent complexity of the problem of ID/LP parsing and on the other hand due to the grammars usually used with Troll as explained in section 7.5. Nevertheless, some optimizations were included in the implementation.

The main optimization involves a technique described in Gerdemann (1989) which uses restriction to limit the number of edges resulting from prediction. Every time a prediction is made, it is checked whether there has already been a prediction on the restricted version of this category at this particular vertex. This requires storage of already predicted categories with the respective vertex number. If there has been one, the prediction is not repeated. If not, the algorithm proceeds as usual, i.e. it predicts on the edges. Especially with the rather underspecified categories of the Troll grammars tested and ID/LP parsing which allows for prediction from the whole right hand side multiset, this saved a lot of edges.

A further optimization is closely connected to that. To save edges on the chart, we store only those active edges that could contribute to the parse, i.e. if for example no predictions could be done from an active edge, this edge can not contribute to a successful parse and does not need to be stored in the chart. This requires the addition of a marker for successful prediction to the storage of the already predicted edges.

Since the implementation utilizes an agenda, the optimization techniques concerned with intelligent agenda management as described in Erbach (1991) could be employed. But this would require extensive additional work which does not seem necessary at this point, since usually one is interested in all results, not just one. The parser has to exhaust the search space and so the search strategy is secondary. If one is interested in just one parse, it may be a valid option of optimization.

Another point to apply optimization is to avoid the recomputation of the LP acceptability after a change on the LP store has been discovered. As discussed briefly before, the algorithm has to redo the test for LP acceptability of the whole structure with all rules. One choice to optimize this might be to store the relevant LP rules together with the pair so that the computation does not involve all LP rules. This may improve on the speed, but certainly will take up a lot of space and has therefore not been implemented. Another choice might be to go even further and not to store the whole sequence, but rather to find a way to store just those parts of the feature graphs which are responsible for the possible violation of an LP rule. This would require a fairly sophisticated algorithm to determine the relevant substructures of the sequences. In the end, the goal might be to achieve the exact behavior of delay patterns where computation is resumed at exactly the point it was left of. It seems not possible to do this with the mechanisms employed here, i.e. the existing Troll procedures. So this is left for further research.

Naturally, the code was kept as deterministic as possible. To save on storage space, failure driven loops were used in the completion and prediction predicates instead of using built in findall predicates, though those might be more efficiently implemented internally. Standard Prolog optimization techniques such as lagging and first argument indexing were employed as well. For a description of the techniques, see O'Keefe (1990).

## 7.5 Some Test Experiences

All the discussions in this section relate to tests with ID/LP versions of Seiffert's example grammar, a very simple HPSG grammar covering *she walks*, and the implementation of the handling of partial verb phrases along the lines of Hinrichs and Nakazawa (1993) as described in Meurers (1994).

As it turned out, the parser is not particularly well suited to the efficient parsing of HPSG like grammars. The problems are rather general. HPSG being a grammar formalism where most of the information is not present in the rules/schemata, but rather in the lexical entries, the top down guidance the parser could provide can not be utilized to its fullest. In the worst case, the parser works best without any top down guidance at all (see Morawietz (in preparation)). Furthermore, the guidance can not be used to achieve the greatest advantage since the unlimited prediction from the right hand side multiset may lead to too many predictions – in the worst case all rules are predicted – and so the guidance is lost. What complicates the top down guidance as well is that in the implementation a lot of the information on the rules is contained in the attached definite clauses (Meurers 1994) and is thereby not available in the early stages of processing so that a lot of work is done which does not contribute to the resulting parse. This is due to the fact that the definite clauses are only executed in case all categories of the right hand side have been recognized. But as discussed in section 7.2, this is the only way available so far to ensure that all information necessary to execute the definite clauses has been found. To overcome this problem – at least in parts – the deterministic definite clauses where partially executed automatically. This lead to an improvement of about 30%. But since nothing prevents the account from being used on bottom up chart parsers which exhibit better performance for this kind of grammars, a real efficient ID/LP parser for HPSG like grammars is still to be implemented. The top down presentation was chosen to stay close to the approaches discussed in the literature so far. And we are contend to present a prototype for the solution.

To go into more detail, one test experience resulted in the change of some parts of the implementation. At first, both the chart and the agenda were kept as a list, but some test runs quickly showed that this resulted in very large structures and thereby caused severe problems with Quintus Prolog's memory management. The idea of keeping chart and agenda as lists or heaps, and to pass them on as arguments had to be abandoned. Both chart and agenda are now asserted (and retracted) in the database. This may be seen as bad programming style and as being inefficient. But experiences with bottom up parsers differing only in this data storage showed little or no difference in timings. So this (forced) choice does not seem to induce any extra cost.

Apart from correctness, the test whether an edge is subsumed by an edge already in the chart does not rule out many hypotheses. This is partly due to the fact that the predictions are already limited as much as possible so that a lot of edges which could be ruled out by subsumption are never stored in the agenda. Since the test is costly to perform, especially in case one is dealing with a lot of dependent disjunctions, it seems in the end more efficient to switch the test off in case no termination problems occur. That handling of dependent disjunctions is expensive is due to the fact that all possible permutations have to be calculated and stored in case two dependency groups are compared. The storage seems necessary because one has to keep track of the interactions that occurred previously to ensure that the subsumption between the groups is still possible. It might be the case that particular possibilities do not contribute and must not contribute later on.

The prediction step was tested both with and without the usage of LP rules. As discussed in Morawietz (in preparation), the usage of LP rules may lead to more edges on the chart and agenda and due to this to a lengthening of the parsing times. The reasons for this are that some very general edges are discarded which saved via subsumption and the test for predictions already done a lot of effort on specific instances of those discarded edges. So the test for LP acceptability does indeed not gain anything in the predictor, even apart from the problems discussed in previous sections.

To get any comparison between the ID/LP and the phrase structure version of the partial verb phrase grammar at all, an Earley style chart parser with the same optimizations as the ID/LP parser was implemented. Needless to say, both are slower than the usual bottom up parser. The phrase structure grammar contains 25 rules, the ID/LP version 13; lexicon, definite clauses, restrictor and initial category stay the same for both; the ID/LP version employs 7 LP rules. For NPs, the ID/LP parser needs between 140% and 300% of the time the Earley parser needs. The number of edges considered, i.e. those on the agenda, is not that much different although there are more in case of the ID/LP parser; but the number of edges that do appear on the chart differ considerably (up to 40% more). For VPs and sentences, the time used by the ID/LP parser lies between 300% and 500% of the times used by the Earley parser and increases with the length of the sentences; the number of edges on the agenda increases only by 20%, but the number of edges on the chart nearly doubles. Overall, it can be said that the ID/LP parser needs much more time than the phrase structure version. This may be due to the fact that the considered fragment was too small, so that the difference in the number of rules is not big enough, i.e. not enough rules could be made redundant because they were only differing in word order. But the nature of HPSG is such that it tries to work with only a few ID schemata, and if this carries over to the phrase structure implementations of HPSG grammars, i.e.

that there are not that many rules, and the ID/LP versions of those do not gain enough in the saving of the number of rules, then it may not be worth to use the ID/LP format computationally. Naturally, the grammar writer still has the discussed benefits of the formalism.

Summing up, it can be said that the proposed treatment of the nonlocal flow of information works as desired, but that it remains to be seen whether there is a way to implement it in a fashion which does not cost too much computationally.

# 8 Conclusions

In this thesis, we defined typed unification ID/LP grammars and gave an algorithm for parsing them, thereby solving the problem of nonlocal feature passing concerning linear precedence phenomena. This enables linguists to write ID/LP grammars and gives the computational linguists a clear formalization and the means to directly parse such grammars in one pass. We also hinted at the possibility to view HPSG grammars under certain circumstances as typed unification ID/LP grammars thereby proposing a method for processing of HPSG grammars by phrase structure based natural language systems like TROLL.

Nevertheless, some open questions remain. Some are induced by the difference between HPSG grammars and typed unification ID/LP grammars, some by linguistic needs and some by formal considerations.

Although we are not really concerned with it, from a linguistic point of view there are some demands that have not (yet) been met by the proposed approach. In some linguistic analyses, only binary branching structures are considered, for example Uszkoreit (1984), which would lead to problems with direct parsing from a linguistic point of view because in our approach only local trees can be ordered by LP rules. If one is dealing with these binary structures, the domain of application of LP rules has to be enlarged, for example to something along the lines of a head domain, as in Engelkamp et al. (1992). In our approach, this would have to be done by some extra mechanism that collects the categories in the list under the recognized feature until the projection becomes itself the argument. This would cost extra time and effort, but there is no reason why it would not be possible to do so. Another linguistic problem which has not been addressed at all, is the one of discontinuous constituents. So far, no conclusive linguistic analyses of those phenomena are found in the literature and therefore it is not at all clear which kind of technical problems may be involved. So, the typed unification ID/LP grammars do not directly support discontinuous constituents, unless the linguist finds a way to incorporate them in the present formalization. The same is true for conflicting LP statements in the sense of Uszkoreit (1984). If the grammar writer does not find a way to incorporate a treatment of the problem in the present formalism – which is highly unlikely considering the definitions concerning feature graphs – the structures are not acceptable under the present setup of the formalization.

The difference between HPSG and the class of typed ID/LP grammars is considerable and not all differences can be discussed here. We believe the major problem for our purposes to be the ID schemata used in HPSG, as opposed to the ID rules employed in our formalization. It would be quite interesting for further research whether there is a way to automatically derive rules from the given schemata. For some, it is very easy since they do not constitute real schemata since their number of daughters is already fixed. It is more complicated for the other schemata, but this could be done for example by trying to match the schemata against all the instances of subcategorization requirements of the lexical entries, thereby gaining rules. Later on these rules could be compacted by generalizing them so that only a limited and tractable number of them remains. Naturally, this is only a very tentative thought and whether the problem can be resolved by this is a topic of further research. The problems involved are complex. Just to give an example, it is not clear how to handle constructions like argument raising, as explained in Hinrichs and Nakazawa (1989), where the complements of one verb are raised to a higher one to be satisfied on a different level. Though maybe even this could be solved by a match against the subcategorization information of the elements in the lexicon. The major problems that have not been addressed are naturally the semantic properties underlying the formalism of HPSG and our approach of handling HPSG procedurally. For these reasons, we did not formalize the connection between HPSG and the typed unification ID/LP grammars and therefore the impact of the approach proposed here in relation to HPSG is open to further research.

This leads to the last open question to be addressed here, namely that we do not provide any semantics for LP rules. The formalization does not contribute in any way to a better understanding what LP rules mean for the present formalism and neither for HPSG. It just shows a way how to deal with the ID/LP distinction in a technical manner without trying to incorporate it into a logic which formalizes HPSG. This is clearly beyond the scope of this thesis, since it would require an entirely different approach.

Apart from those discussions, our approach presents a local solution to a non-local phenomenon. Although it does not rule out the violations by something as simple as unification, it nevertheless uses the existing mechanism of the test for LP acceptability to detect the violations as early as they can be discovered. No complex backtracking into the chart or of the whole process is necessary since not the local tree where the LP violation occurs is seen as being invalid, but rather the combination of the information of this local tree with another local tree is not acceptable since it leads to a sequence of categories which does not confirm to the LP rules. It is perfectly acceptable to combine the local tree with another structure in a way that does not lead to failure.

Furthermore, the solution is neutral toward processing direction, i.e. bottom up versus top down, and processing mode, i.e. parsing versus generation. The way the solution is presented here is for a top down parser, but nothing in the way the problem is handled requires any of the special mechanisms involved with top down parsing. In fact, the first prototype to be implemented was a bottom up parser as mentioned before in section 6.2.1. More exactly, we believe it possible to define an ID/LP parsing schema along the lines done for other parsing problems in Sikkel (1993). This would allow for a better classification of the approach. This definition is left to further research for reasons of time. And since Earley's algorithm may be used for generation as well as for parsing (Gerdemann 1991), it seems straightforward to incorporate our treatment of the ID/LP formalism for generation.

One last point to discuss seems to be the question of dynamic and static approaches to the problem. Clearly, our approach is a dynamic one. LP rules are applied during runtime. In principle, it is still a generate and test solution although generating and testing has been interleaved to the point of being done simultaneously. If one could find a static approach which solves the problem at compile time, this could be more efficient. It is a question of how and when which approach is going to be used. The dynamic approach of direct parsing allows for a flexible change of ID and LP rules and very exact testing of parts of the grammar. A static approach would allow for better parsing times, but might not have the flexibility since the compilation process would be very complex. This seems to indicate that a static approach may be better for a system and a grammar that is to be used, whereas the dynamic approach seems to be well suited to a grammar developing environment where the grammar is to be tested and changed accordingly.

# A The Code

This appendix contains the code for the described implementation.[26] Since it seems easier to understand the code, if the internal format is known, TROLL's way of representing feature graphs internally is described here. A feature graph is a triple, consisting of the feature tree, the reentrancy list and the dependency list, i.e. `fs(FT,RL,DL)`. The feature tree is encoded by having nodes with the types and features, i.e. `node(Type,Features)`. The features are a prolog list with arguments consisting of a feature and a node, i.e. `[feat(Feat1,Node1),feat(Feat2,Node2),...]`. The features are ordered alphabetically on the feature list. Reentrancies in the feature tree are wrapped, so that they are easily recognizable, i.e. `rnum(Num)`. The reentrancy list consists of entries of a variable and the appropriate node, i.e. `[re(Num1,Node1),re(Num2,Node2),...]`. The reentrancy list may contain chains. The dependency list entries are of the format `group(Group,Template,Cases)`. Group is a variable which identifies the group. Template is a list of variables which enable to identify the possible values belonging to a particular variable, i.e. which values occur at a particular node. Cases is a list of lists which contains the columns which belong in some sense together, i.e. the single elements of this set of feature graphs. The format for the LP rules is `lp(FS1,FS2)`. Lexical entries are `lex_entry(Word,FS)` and rules are `rule(FS)`.

Some predicates do occur in the code, but are not defined since they were not written by me or are self explanatory. The most prominent of those are `unify_fs/3` and `subsumes/2`. The unification is non destructive which means that the result is in the third argument whereas the two input feature graphs remain unchanged in the first and second argument. Subsumption has to have the more general feature graph in the first argument, the specific one in the second. Auxiliary predicates are

`make_restrictor/2` with input a list of paths and output the needed format for `restrict/3`,

`restrict/3` with input restrictor and a feature graph and output the restricted version of the input feature graph,

`reentrancy_member/2` is just as the standard `member/2` predicate only for reentrancy lists,

`remove_reentrancy/3` removes an entry from the reentrancy list and returns the rest of the list,

`clean_up/2` cleans the reentrancy and dependency lists from chains of entries and entries which are no longer necessary,

`cpu_time/2` takes a goal and returns the time which was necessary to execute it.

[26]The code given in this section is somewhat pretty printed so that the predicate names correspond to the intuitions. In the real implementation this was not done because of compatibility reasons with other parsers.

`write_sentence/2` prints the sentence to the screen with the edge numbering and returns the length of the input string.

`get_chart_length/1` counts the edges on the chart.

`get_lp_rules/1` collects all the LP rules in a list.

`get_restrictor/1` returns the restrictor as a list of paths.

`get_lp_store/2` extracts the LP store from a PS rule.

`extract_lhs/2` extracts the left hand side from a PS rule.

`embed_lhs/2` embeds a feature structure so that it becomes the left hand side of a PS rule.

`add_lp_store/3` adds an LP store to a PS rule.

`get_length_rhs/2` calculates the length of the right hand side of a PS rule.

`empty_rhs/1` checks whether the right hand side of a PS rule is empty.

`append_fs/3` appends a FS at the beginning of a list of signs.

`copy_fs/4` copies a feature graph thereby removing all but the internal reentrancies.

`report/5` prints the cpu time used, the resulting number of edges on chart and agenda and the number of solutions or the solutions themselves.

## A.1 The Parser

```
%%%%%%%%%%%%%%%%%%%%%%%%%%%%%%%%%%%%%%%%%%%%%%%%%%%%%%%%%%%%%%%
%                                                             %
%    Top-down ID/LP Earley parser for Troll that solves       %
%    the problem of non-local backward feature passing        %
%                                                             %
%%%%%%%%%%%%%%%%%%%%%%%%%%%%%%%%%%%%%%%%%%%%%%%%%%%%%%%%%%%%%%%

 % In principle this is a modified version of the top-down parser found
 % in Gazdar & Mellish (1989) augmented to handle feature structures and
 % ID/LP grammars. It utilizes a chart and an agenda. Several
 % improvements have been added to make it more efficient. The
 % improvements involve the use of a restrictor to ensure termination
 % and several techniques to save on the prediction and the addition of
 % edges to the chart (Gerdemann 1991).

:- ensure_loaded(lp_acceptable).

 % parse(+String)
 % top level goal of the parser, resets the dynamic predicates in the
 % database and initializes the Restrictor, the StartRule, the LPRules,
```

```
 % the Chart and the Agenda. Additionally some output is given, namely
 % the input sentence with the respective vertex numbers and the
 % resulting output of the parser, i.e. timings and the counted edges.
 % Note that both Chart and Agenda are asserted in the database because
 % of Quintus memory problems.

parse(String):-
        reset_database,
        write_sentence(String,End),
        get_restrictor(Restrictor),
        make_start_rule(StartRule),
        get_lp_rules(LPRules),
        predict(StartRule,0,Restrictor,_Flag),
        scan(String),
        cpu_time(extend_edges(1,Restrictor,End,LPRules,Num_Edges),Time),
        report(parse,Time,End,Num_Edges,_Chart).


 % scan(+String)
 % takes the string and calls start_chart/2, starting with vertex 0.

scan(String):-
        start_chart(String,0).


 % start_chart(+String,+Vertex)
 % initializes the chart by asserting edges for all lexical entries
 % found for each word, and for all positions the empty categories are
 % added in the form of passive edges as well. This is done in a findall
 % loop to be able to discover easily whether there is a lexical entry
 % for a word at all. The edges get a 'passive' wrapper as to find them
 % easier in the database. Each lexical entry is turned into a ps_rule
 % with empty goals, rhs, lp store and rec features.

start_chart([],V0):-
        findall(edge(V0,V0,FT),
                (lex_entry([],fs(FT,RL,DL)),
                assertz(passive(V0,V0,fs(node(ps_rule,
                                 [feat(goals,node(e_list,[])),
                                 feat(lhs,FT),
                                 feat(lp_store,node(eset,[])),
                                 feat(rec,node(e_list,[])),
                                 feat(rhs,node(e_list,[]))])
                           ,RL,DL)))).
               _Chart).
```

```
start_chart([Word|Words],V0):-
        findall(edge(V0,V0,FT),
                (lex_entry([],fs(FT,RL,DL)),
                assertz(passive(V0,V0,fs(node(ps_rule,
                                 [feat(goals,node(e_list,[])),
                                 feat(lhs,FT),
                                 feat(lp_store,node(eset,[])),
                                 feat(rec,node(e_list,[])),
                                 feat(rhs,node(e_list,[]))])
                            ,RL,DL)))),
                _Chart),
        V1 is V0 +1,
        findall(edge(V0,V1,FT),
                (lex_entry(Word,fs(FT,RL,DL)),
                assertz(passive(V0,V1,fs(node(ps_rule,
                                 [feat(goals,node(e_list,[])),
                                 feat(lhs,FT),
                                 feat(lp_store,node(eset,[])),
                                 feat(rec,node(e_list,[])),
                                 feat(rhs,node(e_list,[]))])
                            ,RL,DL)))),
                Chart1),
        ( Chart1 = [] ->
            write('No lexical entry found for word >>'),
            write(Word),
            write(' <<'),nl,nl
        ;   true),
        start_chart(Words,V1).


 % extend_edges(+Counter,+Restrictor,+StringLength,+LPRules,-NumOfEdges)
 % takes an edge from the agenda, checks whether it is already subsumed
 % by an edge in the chart and starts the appropriate further operations
 % with the edge (predicate new_edges/5). The edge is asserted to
 % the chart just in case it was either a passive edge or an active edge
 % that was used in a successful prediction. It also counts the edges
 % taken from the agenda. The Restrictor, the StringLength and the
 % LPRules are just passed on.

extend_edges(N,Restrictor,End,LPRules,No):-
        retract(agenda(Edge)),!,
        (subsumes_edge(Edge) ->
            true
        ;   (new_edges(Edge,Restrictor,End,LPRules,Flag),
             (Flag == good ->
                 assertz(Edge)
```

```
              ;   true))),
        N1 is N+1,
        extend_edges(N1,Restrictor,End,LPRules,No).
extend_edges(N,_Restrictor,_End,_LPRules,NumOfEdges):-
        NumOfEdges is N-1.


 % new_edges(+Edge,+Restrictor,+StringLength,+LPRules,-Flag)
 % takes the edge and calls the appropriate complete and prediction
 % steps. The Restrictor, the StringLength and the LPRules are just
 % passed on. The Flag markes whether the edge has to be asserted to the
 % chart.

new_edges(passive(V1,V2,Rule),_Restrictor,End,LPRules,good) :-
        complete_passive(V1,V2,Rule,End,LPRules).
new_edges(active(V1,V2,Rule),Restrictor,End,LPRules,Good) :-
        predict(Rule,V2,Restrictor,Good),
        complete_active(V1,V2,Rule,End,LPRules).


 % complete_passive(+From,+To,+Rule,+StringLength,+LPRules)
 % does the actual completion step. Since the parser deals with
 % non-local backward feature passing, the process is a bit more
 % complicated than usual though the principle stays the same.
 % The Left Hand Side of the input Rule is extracted to check whether
 % there exists an active edge in the chart where one of its elements on
 % the right hand side unifies with this LHS. This is done in a fail
 % loop. To be able to unify the structures properly, the length of the
 % right hand side is computed and LHS is embedded nondeterministically
 % on all possible positions on the righthandside. If this unification
 % was successful, the LP stores are extracted, unified, checked for
 % acceptance and added to the resulting new rule. Only if this does not
 % lead to a violation the new edge is constructed. This construction
 % (change_fs/4) contains the check for LP acceptability - including the
 % construction of a new LP store - and the moving of the recognized
 % category from the right hand side of the rule to the recognized
 % sequence of FSs. If the righthandside is empty, the goals are
 % executed and a cleaned version of the edge is asserted. Otherwise,
 % the resulting cleaned edge is asserted.

complete_passive(V1,V2,Rule,End,LPRules):-
        extract_lhs(Rule,LHS),
        active(V0,V1,Rule1),
        get_length_rhs(Rule1,Length),
        embed_fs(LHS,Length,RuleLHS,Pos),
        unify_fs(RuleLHS,Rule1,Rule2),
```

```
        get_lp_store(Rule,LPStore),
        get_lp_store(Rule2,LPStore2),
        union(LPStore,LPStore2,OutLP_Store1),
        check_lp_store(OutLP_Store1,LPRules,OutLP_Store),
        add_lp_store(Rule2,OutLP_Store,Rule3),
        change_fs(Rule3,Pos,LPRules,Rule4),
        ( empty_rhs(Rule4) ->
            execute_goals(Rule4,OutRule4),
            clean_up(OutRule4,OutRule),
            assertz(agenda(passive(V0,V2,OutRule)))
        ;   clean_up(Rule4,OutRule),
            assertz(agenda(active(V0,V2,OutRule)))),
        fail.
complete_passive(_,_,_,_,_).


 % complete_active(+From,+To,+Rule,+StringLength,+LPRules)
 % does the actual completion step. An appropriate passive edge is
 % taken from the chart, its left hand side taken and embedded
 % nondeterministically according to the computed length of the right
 % hand side of the active rule to be able to test unifiability. If this
 % unification was successful, the LP stores are extracted, unified,
 % checked for acceptance and added to the resulting new rule. Only if
 % this does not lead to a violation the new edge is constructed. This
 % construction contains the check for LP acceptability - including the
 % construction of a new LP store - and the moving of the recognized
 % category from the right hand side of the rule to the recognized
 % sequence of FSs. If the right hand side is empty, the goals are
 % executed and a cleaned version of the edge is asserted. Otherwise,
 % the resulting cleaned edge is asserted.

complete_active(V1,V2,Rule,End,LPRules):-
        get_length_rhs(Rule,Length),
        passive(V2,V3,Rule1),
        extract_lhs(Rule1,LHS1),
        embed_fs(LHS1,Length,RuleLHS1,Pos),
        unify_fs(RuleLHS1,Rule,Rule2),
        get_lp_store(Rule1,LPStore1),
        get_lp_store(Rule2,LPStore2),
        union(LPStore1,LPStore2,OutLP_Store1),
        check_lp_store(OutLP_Store1,LPRules,OutLP_Store),
        add_lp_store(Rule2,OutLP_Store,Rule3),
        change_fs(Rule3,Pos,LPRules,Rule4),
        ( empty_rhs(Rule4) ->
            execute_goals(Rule4,OutRule4),
            clean_up(OutRule4,OutRule),
```

```
            assertz(agenda(passive(V1,V3,OutRule)))
        ;   clean_up(Rule4,OutRule),
            assertz(agenda(active(V1,V3,OutRule)))),
        fail.
complete_active(_,_,_,_,_).


 % predict(+Rule,+Vertex,+Restrictor,-Flag)
 % constitutes the prediction step of the algorithm. The Rule is taken
 % and via fail loop all FSs on the right hand side are extracted and
 % tried for prediction. To ensure termination only the restricted
 % versions of the FSs are used. For greater efficiency only those
 % predictions are actually made that haven't been done before. A Flag
 % markes if there have been any predictions at all. If there are none
 % or have been none the FS does not need to be added to the chart.
 % This is the version that does not check lp acceptability.

predict(InRule,Num,Restrictor,_Flag):-
        get_next_fs(InRule,FS),
        restrict(Restrictor,FS,OutFS),
        (already_predicted(Num,OutFS,Flag) ->
            (Flag = good ->
                assert(flag(good))
            ;   true)
        ;   predict_aux(Num,OutFS),
            (flag(good) ->
                assertz(predicted(Num,OutFS,good))
            ;   assertz(predicted(Num,OutFS,bad)))),
        fail.
predict(_,_,_,good):-
        flag(good),
        retractall(flag(_)).
predict(_,_,_,bad):-
        \+flag(good).


 % predict_aux(+Vertex,+FS)
 % does the actual prediction. The FS is embedded under the left hand
 % side and then tried to unify with any rule. An active edge is
 % constructed and asserted to the agenda.

predict_aux(Num,OutFS):-
        embed_lhs(OutFS,RuleFS),
        rule(NewRule),
        unify_fs(NewRule,RuleFS,Rule),
        clean_up(Rule,OutRule),
```

```
        assertz(agenda(active(Num,Num,OutRule))),
        assertz(flag(good)),
        fail.
predict_aux(_,_).
```

## A.2 LP Acceptability

```
%%%%%%%%%%%%%%%%%%%%%%%%%%%%%%%%%%%%%%%%%%%%%%%%%%%%%%%%%%%%%%
%                    LP acceptability                        %
%%%%%%%%%%%%%%%%%%%%%%%%%%%%%%%%%%%%%%%%%%%%%%%%%%%%%%%%%%%%%%

 % poss_lp_accept(+SequenceOfFS,+RL,+DL,+LPRules,+InFlag,-OutFlag)
 % fails if the SequenceOfFS is not LP acceptable, unpacks the first
 % level of the sequence, recurses down the rest, returns yes in OutFlag
 % if the sequence was possibly LP violated.

poss_lp_accept(rnum(Num),RL,DL,LPRules,InFlag,OutFlag):-
        clean_up(fs(rnum(Num),RL,DL),fs(NewFS,RL0,DL0)),
        poss_lp_accept(NewFS,RL0,DL0,LPRules,InFlag,OutFlag).
poss_lp_accept(node(ne_list_sign,[feat(hd,_),
                                  feat(tl,node(e_list,[]))]),_,_,
             _LPRules,Flag,Flag):- !.
poss_lp_accept(node(ne_list_sign,[feat(hd,HD),
                                  feat(tl,TL)]),RL,DL,
              LPRules,InFlag,OutFlag):-
        poss_lp_acceptable(LPRules,fs(HD,RL,DL),fs(TL,RL,DL),no,Flag),
        (Flag == no ->
            InFlag1 = InFlag
        ;   InFlag1 = yes),
        poss_lp_accept(TL,RL,DL,LPRules,InFlag1,OutFlag).


 % poss_lp_acceptable(+LP_Rules,+FirstOfRec,+RestOfRec,+InFlag,-OutFlag)
 % unpacks the FSs from the lp wrapper, fails if First is not allowed
 % after Rest, descends recursively through all LP rules. This is the
 % version that cares for the cases of possible lp violation as well as
 % lp acceptance, the resulting lp situation is marked in OutFlag.

poss_lp_acceptable([],_First,_Rest,Flag,Flag).
poss_lp_acceptable([lp(FS1,FS2)|T],First,fs(Rest,RL,DL),InFlag,OutFlag):-
        poss_lp_aux_1(Rest,RL,DL,First,FS1,FS2,Flag),
        ( Flag == yes ->
            InFlag1 = yes
        ;   InFlag1 = InFlag),
        poss_lp_acceptable(T,First,fs(Rest,RL,DL),InFlag1,OutFlag).
```

```
 % poss_lp_aux_1(+RestList,+RL,+DL,+First,+FS1,+FS2,-Flag)
 % checks whether First is subsumed by the second FS1 of the lp rule, if
 % this is not the case, a test unification is performed to determine
 % whether an lp rule might apply, if that is not the case as well, it
 % is not necessary to check lp acceptability for this constellation;
 % if any of this is the case, it remains to be checked whether one of
 % the elements from the RestList is subsumed by FS2 -> violation of the
 % lp rule if we had a subsumption for FS1 as well, otherwise the test
 % is for possible lp violation. The result is stored in Flag.

poss_lp_aux_1(rnum(Num),RL,DL,First,FS1,FS2,Flag):-
        clean_up(fs(rnum(Num),RL,DL),fs(NewFS,NewRL,NewDL)),
        poss_lp_aux_1(NewFS,NewRL,NewDL,First,FS1,FS2,Flag).
poss_lp_aux_1(node(e_list,[]),_RL,_DL,_First,_FS1,_FS2,_Flag):- !.
poss_lp_aux_1(node(ne_list_sign,List),RL,DL,First,FS1,FS2,Flag):-
        (subsumes(FS1,First) ->
            poss_lp_aux(node(ne_list_sign,List),RL,DL,FS2,no,Flag)
        ;   (unify_fs(FS1,First,_) ->
                poss_lp_aux(node(ne_list_sign,List),RL,DL,FS2,yes,Flag)
            ;   true)).


 % poss_lp_aux(+SubcatList,+FS,+RL,+DL,+InFlag,-OutFlag)
 % checks whether an element of List is subsumed by FS -> fails in
 % the case that the InFlag signals that there was already a subsumption
 % with the other lp element, a violation of an LP rule has occurred. In
 % the other cases a test for possible lp violation is done. if none of
 % these occur, the structure is lp acceptable.

poss_lp_aux(node(e_list,[]),_,_,_FS,Flag,Flag):- !.
poss_lp_aux(node(ne_list_sign,[feat(hd,Next),feat(tl,T)]),RL,DL,
                FS,InFlag,OutFlag):-
        ( subsumes(FS,fs(Next,RL,DL)) ->
            ( InFlag == no ->
                fail
            ;   InFlag1 = yes)
        ;   (unify_fs(FS,fs(Next,RL,DL),_) ->
                InFlag1 = yes
            ;   InFlag1 = InFlag)),
        poss_lp_aux(T,RL,DL,FS,InFlag1,OutFlag).


 % lp_acceptable_rest(+LP_Rules,+FirstOfRec,+RestOfRec)
 % unpacks the FSs from the lp wrapper, fails if First is not allowed
```

```
% before Rest, descends recursively through all LP rules.

lp_acceptable_rest([],_First,_Rest).
lp_acceptable_rest([lp(FS1,FS2)|T],First,fs(Rest,RL,DL)):-
        lp_aux_rest_1(Rest,RL,DL,First,FS2,FS1),
        lp_acceptable_rest(T,First,fs(Rest,RL,DL)).


% lp_aux_rest_1(+RestList,+RL,+DL,+First,+FS2,+FS1)
% checks whether First is subsumed by the second FS1 of the lp rule, if
% this is not the case, it is not necessary to check lp acceptability
% for this constellation; if it is the case, it remains to be checked
% whether one of the elements from the RestList is subsumed by FS2 ->
% violation of the lp rule.

lp_aux_rest_1(rnum(Num),RL,DL,First,FS2,FS1):-
        clean_up(fs(rnum(Num),RL,DL),NewFS),
        lp_aux_rest_1(NewFS,RL,DL,First,FS2,FS1).
lp_aux_rest_1(node(e_list,[]),_,_,_,_,_):- !.
lp_aux_rest_1(node(ne_list_sign,List),RL,DL,First,FS2,FS1):-
        (user:subsumes(FS2,First) ->
            lp_aux_rest(node(ne_list_sign,List),RL,DL,FS1)
        ;   true).

% lp_aux_rest(+List,+RL,+DL,+FS)
% checks whether an element of List is subsumed by FS -> fails in
% this case since a violation of an LP rule has occurred.

lp_aux_rest(node(e_list,[]),_,_,_FS):- !.
lp_aux_rest(node(ne_list_sign,[feat(hd,Next),feat(tl,T)]),RL,DL,FS):-
        \+ user:subsumes(FS,fs(Next,RL,DL)),
        lp_aux_rest(T,RL,DL,FS).
```

### A.3 Auxiliary Procedures

```
%%%%%%%%%%%%%%%%%%%%%%%%%%%%%%%%%%%%%%%%%%%%%%%%%%%%%%%%%%%%%%%
%                       Edge Subsumption routines             %
%%%%%%%%%%%%%%%%%%%%%%%%%%%%%%%%%%%%%%%%%%%%%%%%%%%%%%%%%%%%%%%

% subsumes_edge(+Edge)
% tries to find an edge in the database that is identical to Edge with
% respect to th vertices and where the rule subsumes the rule of the
% input edge.
```

```
subsumes_edge(active(V0,V1,Rule)):-
        active(V0,V1,Rule1),
        subsumes(Rule1,Rule),
        !.
subsumes_edge(passive(V0,V1,Rule)):-
        passive(V0,V1,Rule1),
        subsumes(Rule1,Rule),
        !.


%%%%%%%%%%%%%%%%%%%%%%%%%%%%%%%%%%%%%%%%%%%%%%%%%%%%%%%%%%%%%%%%
%                       Initialization routines                %
%%%%%%%%%%%%%%%%%%%%%%%%%%%%%%%%%%%%%%%%%%%%%%%%%%%%%%%%%%%%%%%%

 % make_start_rule(-StartRule)
 % constructs a ps rule with the appropriately instantiated features and
 % values from the initial FS.

make_start_rule(fs(node(ps_rule,[feat(goals,node(e_list,[])),
                                 feat(lhs,node(dummy,[])),
                                 feat(lp_store,node(eset,[])),
                                 feat(rec,node(e_list,[])),
                                 feat(rhs,node(ne_list_sign,
                                           [feat(hd,Symbol),
                                            feat(tl,node(e_list,[]))]))]))
                        ,RL,DL)):-
        initial(fs(Symbol,RL,DL)).


%%%%%%%%%%%%%%%%%%%%%%%%%%%%%%%%%%%%%%%%%%%%%%%%%%%%%%%%%%%%%%%
%              extracting of substructures from a FS          %
%%%%%%%%%%%%%%%%%%%%%%%%%%%%%%%%%%%%%%%%%%%%%%%%%%%%%%%%%%%%%%%

 % get_next_fs(+PSRule,-FS)
 % takes the input ps rule and returns a member of the right hand side
 % by calling member_rhs/4

get_next_fs(fs(node(ps_rule,[_Goals,_LHS,_LPStore,_Rec,
                        feat(rhs,RHS)]),RL,DL),fs(FS,RL,DL)):-
        member_rhs(RHS,RL,FS).


 % member_rhs(+RHS,+RL,-FS)
 % returns a FS from RHS.
```

```
member_rhs(rnum(Num),RL,FS):-
        reentrancy_member(re(Num,FT),RL),
        member_rhs(FT,RL,FS).
member_rhs(node(ne_list_sign,[feat(hd,FS),_]),_RL,FS).
member_rhs(node(ne_list_sign,[_,feat(tl,RestFS)]),RL,FS):-
        member_rhs(RestFS,RL,FS).


 % pop_rhs(+RHS,+RL,-RHS,-RL,+Acc,+Pos,-FS)
 % pops the FS at position Pos from the right hand side, returns the FS,
 % the resulting right hand side and the resulting reentrancy list.

pop_rhs(rnum(RNum),RL0,RNum,RL,Acc,Pos,FS):-
        remove_reentrancy(re(RNum,FT),RL0,RL1),
        pop_rhs(FT,RL1,PoppedFT,RL2,Acc,Pos,FS),
        RL = [re(RNum,PoppedFT)|RL2].
pop_rhs(node(ne_list_sign,
             [feat(hd,FS),
              feat(tl,Rest)]),RL,Rest,RL,Pos,Pos,FS):- !.
pop_rhs(node(ne_list_sign,
             [feat(hd,HD),
              feat(tl,Rest)]),RLIn,
        node(ne_list_sign,
             [feat(hd,HD),
              feat(tl,Out)]),RLOut,Acc,Pos,FS):-
        Acc1 is Acc+1,
        pop_rhs(Rest,RLIn,Out,RLOut,Acc1,Pos,FS).


%%%%%%%%%%%%%%%%%%%%%%%%%%%%%%%%%%%%%%%%%%%%%%%%%%%%%%%%%%%%%%%%
%              changing of feature structures                  %
%%%%%%%%%%%%%%%%%%%%%%%%%%%%%%%%%%%%%%%%%%%%%%%%%%%%%%%%%%%%%%%%

 % change_fs(+PSRule,+Pos,+LPRules,-PSRule)
 % The FS appropriate according to Pos is extracted from the right hand
 % side of the PSRule, a test is performed whether the feature graph in
 % question may precede all the remaining elements in the right hand
 % side (multi) set, then the test for possible lp violation is
 % performed. If the Flag markes no possible violation of LP rules,
 % the result is the feature structure as before.
 % If there is a possible violation of an LP rule, a new entry for the
 % LP store is created by producing a cleaned copy of the sequence in
 % question for the second element and a reentrancy is introduced that
 % links the first element of the pair to the recognized feature
 % structure. This new element is unified by union_aux2/6 into the
 % existing LP store.
```

```
change_fs(fs(node(ps_rule,
               [feat(Goal,Goals),
                feat(lhs,LHS),
                feat(lp_store,LPStore),
                feat(rec,Rec),
                feat(rhs,RHS)]),RL0,DL),
           Pos,
           LPRules,
           fs(node(ps_rule,
               [feat(Goal,Goals),
                feat(lhs,LHS),
                feat(lp_store,NewLPStore),
                feat(rec,NewRec),
                feat(rhs,RestRHS)]),RL,DL)):-
        pop_rhs(RHS,RL0,RestRHS,RL1,1,Pos,FS),
        lp_acceptable_rest(LPRules,fs(FS,RL0,DL),fs(RestRHS,RL1,DL)),
        poss_lp_acceptable(LPRules,fs(FS,RL0,DL),fs(Rec,RL0,DL),no,Flag),
        append_fs(Rec,FS,NewRec0),
        ( Flag == no ->
            NewLPStore = LPStore,
            RL = RL1,
            NewRec = NewRec0
        ;   copy_fs(NewRec0,RL0,DL,fs(CopyFS,RL4,_DL1)),
            union_aux2(LPStore,RL0,DL,
                  node(pair,[feat(first,rnum(A)),
                             feat(second,CopyFS)]),[re(A,NewRec0)|RL4],
                  NewLPStore,_RL3,DL3),
            NewRec = rnum(A),
            RL = [re(A,NewRec0)|RL0]).


%%%%%%%%%%%%%%%%%%%%%%%%%%%%%%%%%%%%%%%%%%%%%%%%%%%%%%%%%%%%%%%%
%              adding of information to Feature structures     %
%%%%%%%%%%%%%%%%%%%%%%%%%%%%%%%%%%%%%%%%%%%%%%%%%%%%%%%%%%%%%%%%

 % embed_fs(+FS,+MaxEmbedding,-PSRule,-Pos)
 % The intention is to embed a FS on the right hand side of a rule. Here
 % it has to be possible to embed on every depth up to MaxEmbedding
 % since we are dealing only with ID rules. The depth on which the
 % embedding took place is returned in Pos.

embed_fs(fs(FT,RL,DL),Max,fs(node(ps_rule,[feat(rhs,RHS)]),RL,DL),Pos):-
        embed_aux(FT,Max,RHS,1,Pos).
```

```
 % embed_aux(+FS,+MaxEmbedding,-RHS,+Acc,-Acc)
 % takes FS and embeds it in a hpsg type of list as long as the
 % accumulator is smaller or equal to the MaxEmbedding.

embed_aux(FS,_,node(ne_list_sign,[feat(hd,FS)]),Pos,Pos).
embed_aux(FS,Max,node(ne_list_sign,[feat(hd,node(bot,[])),feat(tl,TL)]),
              Acc,Pos):-
        Max > Acc,
        Acc1 is Acc +1,
        embed_aux(FS,Max,TL,Acc1,Pos).


%%%%%%%%%%%%%%%%%%%%%%%%%%%%%%%%%%%%%%%%%%%%%%%%%%%%%%%%%%%%%%%%
%             Comparison of feature structures                 %
%%%%%%%%%%%%%%%%%%%%%%%%%%%%%%%%%%%%%%%%%%%%%%%%%%%%%%%%%%%%%%%%

 % diff(+FS1,+FS2)
 % succeeds if FS1 and FS2 do not subsume each other.

diff(FS1,FS2):-
        \+identical(FS1,FS2).


 % identical(+FS1,+FS2)
 % succeeds if FS1 and FS2 do subsume each other.

identical(FS1,FS2):-
        subsumes(FS1,FS2),
        subsumes(FS2,FS1).


%%%%%%%%%%%%%%%%%%%%%%%%%%%%%%%%%%%%%%%%%%%%%%%%%%%%%%%%%%%%%%%
%             Operations on lp stores                         %
%%%%%%%%%%%%%%%%%%%%%%%%%%%%%%%%%%%%%%%%%%%%%%%%%%%%%%%%%%%%%%%

 % check_lp_store(+LPStore,+LPRules,-LPStore)
 % extracts the pairs from the LP store, where the elements differ and
 % checks whether they are lp acceptable. Returns an LP store that
 % contains only the elements which are still in doubt.

check_lp_store(fs(node(eset,[]),RL,DL),_,fs(node(eset,[]),RL,DL)):- !.
check_lp_store(fs(InLP_Store,RL,DL),LPRules,fs(OutLP_Store,RL0,DL0)):-
        dif(InLP_Store,RL,DL,LPRules,OutLP_Store,RL0,DL0).
```

```
 % dif(+LPStore,+RL,+DL,+LPRules,-LPStore,-RL,-DL)
 % finds the elements in LPStore where the cleaned first and second
 % elements of a pair are different and tests possible LP violation
 % on them. If they are LP acceptable, they are removed from the LP
 % store.

dif(node(eset,[]),RL,DL,_,node(eset,[]),RL,DL):- !.
dif(node(neset_pair,[feat(elt,node(pair,[feat(first,FS1),
                                         feat(second,FS2)])),
                     feat(els,Rest)]),RL,DL,
    LPRules,
    OutFS,RL0,DL0):-
        clean_up(fs(FS1,RL,DL),FS1Out),
        diff(FS1Out,fs(FS2,RL,DL)),
        !,
        poss_lp_accept(FS1,RL,DL,LPRules,no,Flag),
        ( Flag == yes ->
            OutFS = node(neset_pair,
                          [feat(elt,node(pair,
                               [feat(first,FS1),
                                feat(second,FS2)])),
                           feat(els,RestFS)])
        ;   OutFS = RestFS),
        dif(Rest,RL,DL,LPRules,RestFS,RL0,DL0).
dif(node(neset_pair,[feat(elt,node(pair,Pair)),
                     feat(els,Rest)]),RL,DL,
    LPRules,
    node(neset_pair,[feat(elt,node(pair,Pair)),
                     feat(els,OutFS)]),RL0,DL0):-
        dif(Rest,RL,DL,LPRules,OutFS,RL0,DL0).


 % union(+LPStore1,+LPStore2,-LPStore)
 % It is necessary to unify the LP stores in the parser but this can't
 % be done with the ordinary unification algorithm since set unification
 % is essentially the same as list unification in Troll. union/3 is the
 % top level predicate of this process, it catches the easy cases and
 % calls union_aux1/6 to do the work. The proper set unification can be
 % achieved here only for this special case because of the format of the
 % data. If you want this in general for Troll, you need another logic.

union(fs(node(eset,[]),_,_),LPStore,LPStore):- !.
union(LPStore,fs(node(eset,[]),_,_),LPStore):- !.
union(fs(LPStore1,RL1,DL1),fs(LPStore2,RL2,DL2),fs(OutLPStore,RL,DL)):-
        union_aux1(LPStore1,RL1,DL1,LPStore2,RL2,DL2,OutLPStore,RL,DL).
```

```
 % union_aux1(+LPStore1,+RL1,+DL1,+LPStore2,+RL2,+DL2,-LPStore,-RL,-DL)
 % recurses through the first LP store and extracts the relevant pairs
 % and calls union_aux2/6 to unify this pair into the second LP store.

union_aux1(node(eset,[]),_,_,LPStore,RL,DL,LPStore,RL,DL):- !.
union_aux1(node(neset_pair,[feat(elt,Pair),feat(els,Rest)]),RL1,DL1,
      LPStore,RL2,DL2,
      OutLPStore,RL,DL):-
        union_aux2(LPStore,RL2,DL2,Pair,RL1,DL1,NewLPStore,RL0,DL0),
        union_aux1(Rest,RL1,DL1,NewLPStore,RL0,DL0,OutLPStore,RL,DL).


 % union_aux2(+LPStore,+RL,+DL,+Pair,+RL,+DL,-LPStore,-RL,-DL)
 % the pair can be unified into the input LP store just in case it is
 % not already there, or we find another pair whose second element is
 % literally identical to the second element of the input pair. In this
 % case the first elements of those pairs have to be unified. The
 % predicate recurses through the LP store.

union_aux2(node(eset,[]),RL1,DL1,
          Pair,RL,DL,
          node(neset_pair,[feat(elt,Pair),
                           feat(els,node(eset,[]))]),OutRL,OutDL):-
        !,
        append(RL,RL1,OutRL),
        append(DL,DL1,OutDL).
union_aux2(node(neset_pair,[feat(elt,node(pair,[feat(first,First1),
                                                 feat(second,Second1)])),
                           feat(els,Rest)]),RL1,DL1,
          node(pair,[feat(first,First2),
                     feat(second,Second2)]),RL2,DL2,
          node(neset_pair,[feat(elt,node(pair,[feat(first,First),
                                                 feat(second,Second)])),
                           feat(els,Rest)]),RL,DL):-
        identical(fs(Second1,RL1,DL1),fs(Second2,RL2,DL2)),
        !,
        user:unify_fs(fs(First1,RL1,DL1),fs(First2,RL2,DL2),fs(First,RL,DL)),
        clean_up(fs(First,RL,DL),fs(Second,_RL3,_DL3)).
union_aux2(node(neset_pair,[feat(elt,Pair1),feat(els,InRest)]),RL1,DL1,
          Pair,RL2,DL2,
          node(neset_pair,[feat(elt,Pair1),feat(els,OutRest)]),RL,DL):-
        union_aux2(InRest,RL1,DL1,Pair,RL2,DL2,OutRest,RL,DL).
```

```
%%%%%%%%%%%%%%%%%%%%%%%%%%%%%%%%%%%%%%%%%%%%%%%%%%%%%%%%%%%%%%%
%                     Auxiliary predicates                    %
%%%%%%%%%%%%%%%%%%%%%%%%%%%%%%%%%%%%%%%%%%%%%%%%%%%%%%%%%%%%%%%

 % already_predicted(+Num,+FS,-Flag)
 % checks if the prediction to be made has already been done with a FS
 % that subsumes the input FS. If this is successful the Flag indicates
 % whether any predictions were made from it.

already_predicted(Num,FS,Flag):-
        predicted(Num,FS1,Flag),
        subsumes(FS1,FS),
        !.
```